%% file: main_document.tex
\documentclass[twoside]{article}

\usepackage[accepted]{aistats2024}

\usepackage[utf8]{inputenc} % allow utf-8 input
\usepackage{hyperref}
\usepackage{cleveref}
\usepackage{url}
\usepackage{wrapfig}
\usepackage{floatflt} % Wrapfig alternative
\usepackage{amsthm}
\usepackage{thmtools, thm-restate}
\usepackage{wasysym}
\usepackage{standalone}
\usepackage{array,multirow,booktabs}
\usepackage{natbib}
\usepackage[inline]{enumitem}
\usepackage{xcolor}

\RequirePackage{algorithm}
\RequirePackage{algorithmic}

\usepackage{tikz}
\usetikzlibrary{arrows.meta,positioning,calc}
\newlength{\hor}
\newlength{\ver}
\newcommand{\ap}{\mathrm{Pr}}
\newcommand{\morg}{\widehat{M}}
\newtheorem{thm}{Theorem}[section]
\newtheorem{lem}[thm]{Lemma}
\newtheorem{expl}[thm]{Example}
\newtheorem{dfn}[thm]{Definition}
\newtheorem{ass}[thm]{Assumption}

\Crefformat{thm}{#2Theorem #1#3}
\Crefformat{lem}{#2Lemma #1#3}
\Crefformat{dfn}{#2Definition #1#3}
\Crefformat{ass}{#2Assumption #1#3}
\Crefformat{expl}{#2Example #1#3}
\Crefformat{reb}{#2Remark #1#3}

\input{abbrv}

\begin{document}

\runningauthor{Wäldchen, Sharma, Turan, Zimmer, Pokutta}

\twocolumn[

\aistatstitle{Interpretability Guarantees with Merlin-Arthur Classifiers}

\aistatsauthor{Stephan Wäldchen$^1$ \And Kartikey Sharma$^1$ \And Berkant Turan$^{1,2}$}
\vspace{5pt}
\aistatsauthor{Max Zimmer$^{1,2}$ \And Sebastian Pokutta$^{1,2}$}
 \vspace{8pt}
  \aistatsaddress{$^{1}$Zuse Institute Berlin \\ $^{2}$Technische Universität Berlin} 
]

\begin{abstract}
We propose an interactive multi-agent classifier that provides provable interpretability guarantees even for complex agents such as neural networks.
These guarantees consist of lower bounds on the mutual information between selected features and the classification decision.
Our results are inspired by the Merlin-Arthur protocol from Interactive Proof Systems and express these bounds in terms of measurable metrics such as soundness and completeness.
Compared to existing interactive setups, we rely neither on optimal agents nor on the assumption that features are distributed independently. Instead, we use the relative strength of the agents as well as the new concept of Asymmetric Feature Correlation which 
captures the precise kind of correlations that make interpretability guarantees difficult.
We evaluate our results on two small-scale datasets where high mutual information can be verified explicitly. 
\end{abstract}

%%%%%% MAIN PAPER

\input{main_part}

\bibliographystyle{abbrvnat}
\bibliography{references,interpretability,theory,rebuttal}

\newpage
\appendix
\onecolumn
\aistatstitle{Supplementary Material: Interpretability Guarantees with Merlin-Arthur Classifiers}
{\vskip 0pt plus  -1fil}
\vspace{0cm}

%%%%%% Supplement
\input{online_supplement}

\end{document}

%% file: main_part.tex
\section{Introduction}
\label{sec:introduction}
\begin{figure}
    \centering
    \includegraphics[width=0.4\textwidth]{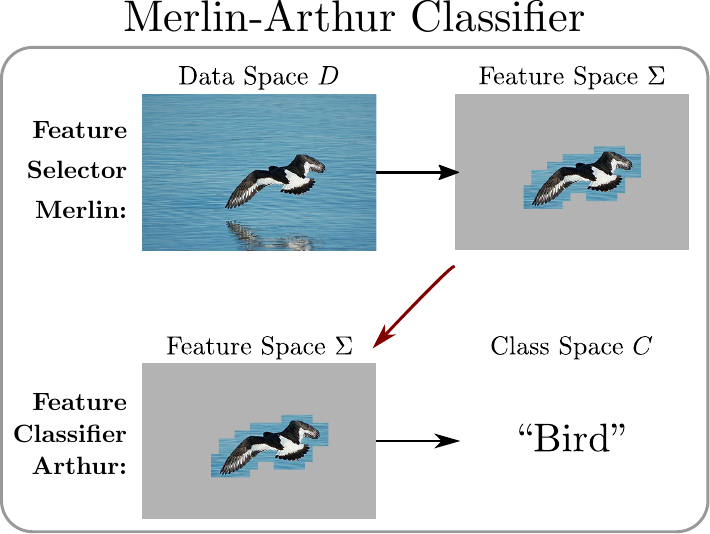}
    \vspace{-0.2cm}
    \caption{\small{The Merlin-Arthur classifier consists of two interactive agents that communicate over an exchanged feature. This feature serves as an interpretation of the classification.}}
    \label{fig:overview}
    \vspace{-0.3cm}
\end{figure}

Safe deployment of \emph{Neural Network} (NN) based AI systems in high-stakes applications requires that their reasoning be subject to human scrutiny.
The field of \emph{Explainable AI} (XAI) has thus put forth a number of interpretability approaches, among them saliency maps~\citep{mohseni2021multidisciplinary}, mechanistic interpretability~\citep{olah2018building} and self-explaining networks~\citep{alvarez2018towards}.
These have had some successes, such as detecting biases in established datasets~\citep{lapuschkin2019unmasking}. However, these approaches are motivated primarily by heuristics and come without any theoretical guarantees. 
Thus, their success cannot be verified. It has also been demonstrated for numerous XAI-methods that they can be manipulated by a clever design of the NNs~\citep{slack2021counterfactual, slack2020fooling, anders2020fairwashing, dimanov2020you}.
On the other hand, formal approaches run into complexity barriers when applied to NNs and require an exponential amount of time to guarantee useful properties~\citep{macdonald2020explaining,ignatiev2019abduction}. This makes any ``right to explanation,'' as in the EU's \emph{GDPR}~\citep{goodman2017european}, unenforceable.

In this work, we design a classifier that guarantees feature-based interpretability under reasonable assumptions. For this, we connect classification to the \emph{Merlin-Arthur protocol}~\citep{arora2009computational} from \emph{Interactive Proof Systems} (IPS), see \Cref{fig:overview}. Our setup consists of a \emph{feature classifier} called Arthur (acting as a verifier) and two \emph{feature selectors} referred to as Merlin and Morgana (acting as provers). Merlin and Morgana choose features from the input and send them to Arthur. 
Merlin aims to send features that cause Arthur to correctly classify the underlying data point. Morgana instead selects features to convince Arthur of the wrong class.
Arthur does not know who sent the feature and is allowed to say ``Don't know!'' if he cannot discern the class.
In this context, we can then translate the concepts of \emph{completeness} and \emph{soundness} from IPS to our setting. Completeness describes the probability that Arthur classifies correctly based on features from Merlin. Soundness is the probability that Arthur does not get fooled by Morgana, thus either giving the correct class or answering ``Don't know!''.
These two quantities can be measured on a test dataset and are used to lower bound the information contained in features selected by Merlin.

\begin{figure}
     \centering \textcolor{black}{
        \includegraphics[width=0.44\textwidth]{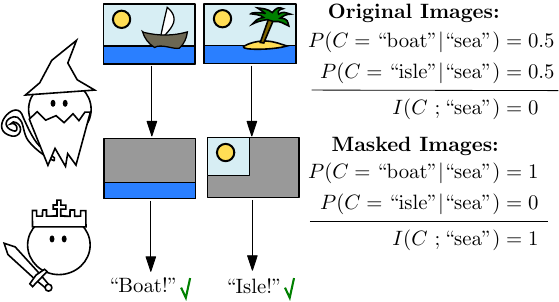}
        \vspace{-0.2cm}
    \caption{ \label{fig:cheating}\small{Illustration of ``cheating'' behaviour. In the original dataset, the features ``sea'' and ``sky'' appear equally in both classes ``boat'' and ``island''. In the partial images created by Merlin, the ``sea'' feature appears only in ``boat'' images and the ``sky'' feature only for ``islands''. Thus, these features now strongly indicate the class of the image. This allows Merlin to communicate the correct class with uninformative features --- in contrast to our concept of an interpretable classifier. }}}
    % \vspace{-3mm}
\end{figure}

\subsection{Related Work}

Formal approaches to interpretability, such as mutual information~\citep{chen2018learning} or Shapley values~\cite{frye2020shapley}, generally make use of partial inputs to the classifier. These partial inputs are realised by considering distributions over inputs conditioned on the given information. However, modelling these distributions is difficult for non-synthetic data. This has been pursued practically by training a generative model as in~\cite{chattopadhyay2022interpretable}.
But as of yet there is no approach that provides a bound on the quality of these models. We discuss these approaches and their challenges in greater detail in~\Cref{apx:modelling}.

\begin{figure*}[t]
\centering
\begin{tabular}{cccc}
     \includegraphics[height=0.22\textwidth]{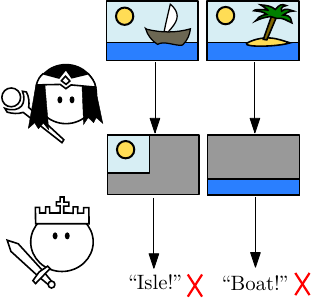}
     &
      \includegraphics[height=0.22\textwidth]{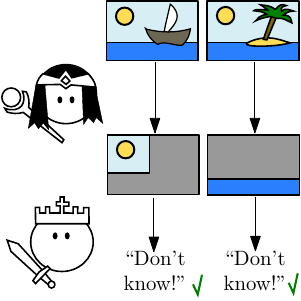}
      &
       \includegraphics[height=0.22\textwidth]{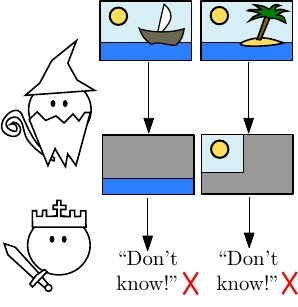}
       &
       \includegraphics[height=0.22\textwidth]{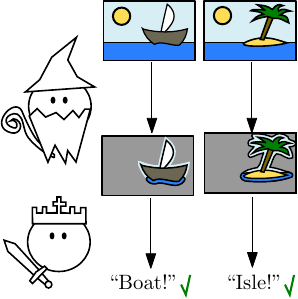}
       \\
     a) & b) & c) & d) 
\end{tabular}
\vspace{-0.2cm}
\caption{\small{Strategy evolution with Morgana. a) Due to the ``cheating'' strategy from \Cref{fig:cheating}, Arthur expects the ``sea'' feature for boats and the ``sky'' for islands. Morgana can exploit this and send the ``sky'' feature to trick Arthur into classifying a ``boat'' image as an ``island'' (and vice versa with ``sea'').
b) To not be fooled into the wrong class when represented with an ambiguous feature, Arthur refrains from giving a concrete classification.
c) Since Arthur does not know who sends the features, he now cannot leverage the uninformative features sent by Merlin.
d) Merlin adapts his strategy to only send unambiguous features that cannot be used by Morgana to fool Arthur.}
\label{fig:evolution}
}
\end{figure*}

Interactive classification in form of a prover-verifier setting has emerged as a way to design inherently interpretable classifiers~\citep{lei2016rationalizing, bastings2019interpretable}. In this setup, the feature selector chooses a feature from a data point and presents it to the classifier who decides the class, see~\Cref{fig:cheating}. 
The classification accuracy is meant to guarantee the informativeness of the exchanged features. However, it was noted by \citeauthor{yu2019rethinking} that the selector and the classifier can cooperate to achieve high accuracy while communicating over uninformative features, see~\Cref{fig:cheating} for an illustration of this ``cheating''. 
Thus, one cannot hope to bound the information content of features via accuracy alone. 
\citeauthor{chang2019game} include an adversarial selector to prevent the cheating. The reasoning is that any ``cheating'' strategy can be exploited by the adversary to fool the classifier into stating the wrong class, see~\Cref{fig:evolution} for an illustration. \citeauthor{anil2021learning} investigate scenarios in which the three-player setup converges to an equilibrium of perfect completeness and soundness. However, this work assumes that a perfect strategy exists and can be reached through training. For many classification problems, such as the ones we explore in our experimental section, no strategies are perfectly sound and complete when the size of the certificate is limited.

Alternative adversarial setups have been proposed in \cite{yu2019rethinking} and \cite{irving2018ai}, but without information bounds. We discuss these ideas in detail in~\Cref{apx:alternative} and show via counterexamples why these formulations cannot straightforwardly yield bounds similar to ours.

An additional theoretical focus has been the learnability of interpretations~\cite{goldwasser2021interactive,yadav2022learning,poulis2017learning}. In this work, we do not focus on the question of learnability. We instead propose to evaluate soundness and completeness directly on the test dataset, as state-of-the-art models are too complex to guarantee generalisation from a realistic number of training samples.

\citeauthor{chang2019game} introduced prover-verifier games for interpretable classification. The authors show that the best strategy for the provers is to select features with high mutual information with respect to the class, and demonstrate that this setup can be stably trained for large-scale text data. However, these results have three restriction that we resolve in this work:
\textbf{(i)} The features are assumed to be independently distributed. This is an unrealistic assumption for most dataset, including the ones used by \citeauthor{chang2019game}, since features are generally correlated. In this regime, simply modelling the data distribution directly is possible. \textbf{(ii)} The provers can only select one feature at a time without context. This strategy is unlikely to yield useful rationalizations for most types of data where the importance of a feature strongly depends on the features surrounding it, like images and text. \citeauthor{chang2019game} do not keep this restriction for their numerical investigation.
\textbf{(iii)} The result is not quantitative. Since we cannot expect the agents to play optimally on complex data, we need measures of their performance and how this relates to the mutual information of the features.

\subsection{Contribution}
\label{ssc:contribution}

We provide what we believe to be the first quantitative lower bound on the information content of the features in an interpretive setup, eliminating the need to trust a model of the data distribution. Additionally, we improve existing analyses in the following ways:
\begin{enumerate}
 \item We do not assume our agents to be optimal. In~\Cref{thm:minmax}, Merlin is allowed to have an arbitrary strategy and in~\Cref{thm:main}, all three players can play suboptimally. We rather rely on the relative strength of Merlin and Morgana for our bound. We also allow our provers to select the features with the context of the full data point.
 \item We do not make the assumption that features are independently distributed. Instead, we introduce the concept of Asymmetric Feature Correlation (AFC), which captures the correlations that complicate establishing an information bound. In~\Cref{thm:minmax} we circumvent the issue by reducing the dataset, and in \Cref{thm:main} we incorporate the AFC explicitly. In~\Cref{sec:discussion} we discuss why the AFC also matters for other interactive settings.
\end{enumerate}
We numerically demonstrate how the interactive setup prevents a major manipulation that has been demonstrated for other XAI-methods.
Finally, we evaluate our theoretical bounds on the MNIST dataset for provers based on Frank-Wolfe optimisers and U-Nets.

\section{Theoretical Framework}\label{sec:theory}

In this section we develop the theoretical framework for the Merlin-Arthur classifier.
What reasonably constitutes a feature strongly depends on the context and prior work often considered subsets of the input as features. W.l.o.g we will stay with this convention for ease of notation. But nothing in our framework relies on these specifics and our theoretical results can be extended to more abstract features as in \cite{chen2018learning} or \cite{ribeiro2018anchors}.

We consider abstract datasets $D\subset [0,1]^d$, where $D$ is possibly infinite, e.g., the set of all images of hand-written digits. $\CD$ is a distribution on this set. The finite training and test sets, e.g., MNIST, are assumed to be faithful samples from this distribution.
Given a vector $\bfx\in D$, we use $\bfx_S$ to represent a vector made of the components of $\bfx$ indexed by the set $S \subseteq \{1,\dots,d\}$.
\begin{dfn}
 Given a dataset $D\subset [0,1]^d$, we define the corresponding \emph{partial} dataset $D_p$ as
 \[
  D_p = \bigcup_{\bfx \in D} \bigcup_{S \subset [d]} \bfx_S.
 \]

\end{dfn}

Every vector $\bfx \in D \subset [0,1]^d$ can be uniquely represented as a set $\skl{(1,x_1), (2,x_2), \dots, (d,x_d)}$.
A partial vector $\bfz \in D_p$ can then be a subset of $\bfx$. 
Thus, $\bfz \subseteq \bfx$ indicates that $\bfx$ contains the feature $\bfz$. 
The set $D_p$ might be further restricted to include only connected sets (for image or text data) or only sets of a certain size as in our numerical investigation.

In our theoretical investigation, we restrict ourselves to two classes and assume the existence of a unique class for every data point. These are restrictions that we hope to relax in further research.

\begin{dfn}[Two-class Data Space]
We consider the tuple $\mathfrak{D} = (D,\CD, c)$ a \emph{two-class data space} consisting of the dataset $D \subseteq [0,1]^d$, a probability distribution $\CD$ along with the ground truth class map $c: D \rightarrow \skl{-1,1}$.
The \emph{class imbalance} $B$ of a two-class data space is
$
  \max_{l\in \skl{-1,1}}{{\P_{\bfx\sim \CD}\ekl{c(\bfx)=l}}/{\P_{\bfx\sim \CD}\ekl{c(\bfx)=-l}}}.
$
\end{dfn}

We will oftentimes make use of restrictions of the set $D$ and measure $\CD$ to a certain class, e.g., $D_l = \skl{\bfx \in D \,|\, c(\bfx) = l}$ and
$\CD_l = \CD|_{D_l}$.

\begin{dfn}[Feature Selector]
 For a given dataset $D$, we define a \emph{feature selector} as a map $M:D \rightarrow D_p$ such that for all $\bfx \in D$ we have $M(\bfx) \subseteq \bfx$.
This means that for every data point $\bfx \in D$ the feature selector $M$ chooses a feature that is present in $\bfx$.
We call $\CM(D)$ the space of all feature selectors for a dataset $D$.
\end{dfn}

\begin{dfn}[Feature Classifier]
We define a \emph{feature classifier} for a dataset $D$ as a function $A: D_p \rightarrow \skl{-1,0,1}$. Here, $0$ corresponds to the situation where the classifier is unable to identify a correct class. We call the space of all feature classifiers $\CA$.
\end{dfn}

\subsection{Mutual Information, Entropy and Precision}

We consider a feature to carry class information if it has high mutual information with the class. For a given feature $\bfz\in D_p$ and data points $\bfx\sim \CD$ the mutual information is
\[
 I_{\bfx\sim\CD}(c(\bfx); \bfz \subseteq \bfx) := H_{\bfx\sim\CD}(c(\bfx)) - H_{\bfx\sim\CD}(c(\bfx)\,|\,\bfz \subseteq \bfx).
\]
When the conditional entropy $H_{\bfx\sim\CD}(c(\bfx)\,|\,\bfz \subseteq \bfx)$ goes to zero, the mutual information becomes maximal and reaches the pure class entropy $H_{\bfx\sim\CD}(c(\bfx))$ which measures how uncertain we are about the class a priori.
A closely related concept is \emph{precision}. Given another data point $\bfy$ with feature $\bfz$, precision is defined as $\text{Pr}(\bfz; \bfy):=\P_{\bfx\sim\CD}\ekl{c(\bfx) = c(\bfy) \,|\, \bfz \subseteq \bfx}$ and was introduced in the context of interpretability by~\cite{ribeiro2018anchors} and~\cite{narodytska2019assessing}. We extend this definition to a feature selector.
\begin{dfn}[Average Precision]
  For a given two-class data space $\mathfrak{D}$ and a feature selector $M\in \CM(D)$, we define the \emph{average precision} of $M$ wrt. $\CD$ as
  \[
  \ap_{\CD}(M) := \E_{\bfy\sim \CD} \ekl{\P_{\bfx\sim\CD}\ekl{c(\bfx) = c(\bfy) \,|\, M(\bfy) \subseteq \bfx}}.
  \]
 \end{dfn}
 The average precision $\ap_{\CD}(M)$ can be used to bound the \emph{average} conditional entropy of Merlin's features, defined as
 \begin{multline}
H_{\bfx,\bfy\sim\CD}(c(\bfx)\,|\, M(\bfy) \subseteq \bfx) := \\ \E_{\bfy\sim \CD}\ekl{H_{\bfx \sim \CD}(c(\bfx)\,|\, M(\bfy) \subseteq \bfx)},
\end{multline}
and accordingly the average mutual information, see~\Cref{apx:entropy}. Using this, we can lower-bound the mutual information as follows: 
\begin{multline}\label{eq:mutual_information_bound}
 \E_{\bfy \sim \CD} [I_{\bfx\sim\CD}(c(\bfx); M(\bfy) \subseteq \bfx)] \\ \geq H_{\bfx\sim\CD}(c(\bfx)) - H_b(\ap_{\CD}(M)).
\end{multline}
When the precision goes to 1, the binary entropy $H_b(p)=-p \log(p) - (1-p)\log(1-p)$ goes to 0 and the mutual information becomes maximal. Our results are easier to state in terms of $\ap_\CD(M)$, because of the infinite slope of the binary entropy.

We can connect $\ap_\CD(M)$ back to the precision of any feature selected by $M$ in the following way.
\begin{restatable}{lem}{featureprob}\label{lem:feature_prob}
Given $\mathfrak{D}=(D,\CD, c)$,  $M\in \CM(D)$ and $\delta \in [0,1]$. Let $\bfx, \bfy\sim\CD$, then with probability $1- \delta^{-1}\kl{1-\ap_{\CD}(M)}$, $M(\bfy)$ is a feature s.t.
\[
  \P_{\bfx\sim\CD}\ekl{c(\bfx) = c(\bfy) \,|\, M(\bfy)\subseteq \bfx} \geq 1- \delta.
\]
\end{restatable}
The proof follows directly from Markov's inequality, see~\Cref{apx:entropy}.
We will now introduce a new framework that will allow us to prove bounds on $\ap_{\CD}(M)$ and thus assure feature quality. For $I$ and $H$, we will leave the dependence on the distribution implicit when it is clear from context.

\begin{figure*}[t]
    \centering
    \includegraphics[width=\textwidth]{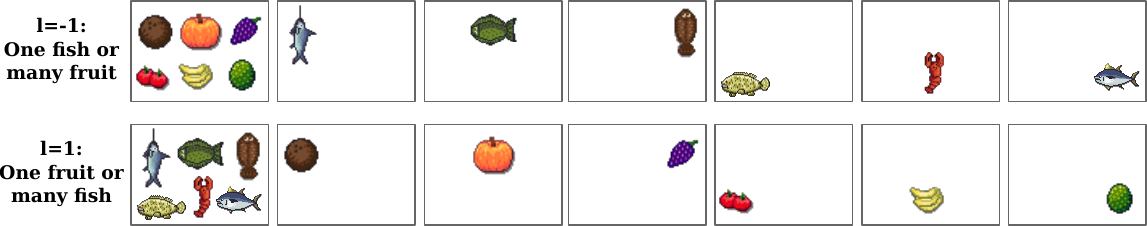}
    \vspace{-0.3cm}
    \caption{\label{fig:afc}\small{Example of a dataset with an AFC $\kappa=6$. The ``fruit'' features are concentrated in one image for class $l=-1$ but spread out over six images for $l=1$ (vice versa for the ``fish'' features). Each individual feature is not indicative of the class as it appears exactly once in each class. Nevertheless, Arthur and Merlin can exchange ``fruits'' to indicate ``$l=1$'' and ``fish'' for ``$l=-1$''. The images where this strategy fails or can be exploited by Morgana are the two images on the left. Applying~\Cref{thm:minmax}, we get $\epsilon_M = \frac{1}{7}$ and the set $D^{\prime}$ corresponds to all images with a single feature. Restricted to $D^{\prime}$, the features determine the class completely.}}
\end{figure*}

\subsection{Merlin-Arthur Classification}

For a feature classifier $A$ (Arthur) and two feature selectors $M$ (Merlin) and $\widehat{M}$ (Morgana) we define
\begin{equation}\label{eq:am:min_max}
 E_{M,\widehat{M},A} := \skl{x \in D\,\middle|\,
 \begin{array}{c}
 A\kl{M(\bfx)} \neq c(\bfx) ~\odr\\  A\kl{\widehat{M}(\bfx)} = -c(\bfx)
 \end{array}}
\end{equation}
as the set of data points for which Merlin fails to convince Arthur of the correct class or Morgana is able to trick him into returning the wrong class, in short, the set of points where Arthur fails. We can now state the following theorem connecting the competitive game between Arthur, Merlin and Morgana to the class conditional entropy.

\begin{restatable}{thm}{minmax}[Min-Max]\label{thm:minmax}
Let $M\in \CM(D)$ be a feature selector and let
 \[
 \epsilon_M = \min_{A \in \CA} \max_{\widehat{M} \in \CM} \,\P_{\bfx\sim \CD}\ekl{\bfx \in E_{M,\widehat{M},A}}.
 \]
 Then a set $D^{\prime}\subset D$ with $\P_{\bfx\sim \CD}\ekl{\bfx \in D^\prime} \geq 1-\epsilon_M$ exists such that for $\CD^\prime = \CD|_{D^\prime}$ we have
 \[
  \ap_{\CD^\prime}(M) = 1, \quad \text{thus}\quad H_{\bfx,\bfy\sim\CD^\prime}(c(\bfy) \;|\; M(\bfy) \subseteq \bfx) = 0.
 \]
\end{restatable}

The proof is in~\Cref{apx:theory}. 
This theorem states that if Merlin's strategy allows Arthur to classify almost perfectly, i.e., small $\epsilon_M$, then there exists a set that covers almost the entire original dataset and on which the class entropy conditioned on the selected features is zero.
Note that these guarantees are for the set $D^\prime$ and not the original set $D$. A bound for the set $D$, such as
\(
  \ap_{\CD}(M) \geq 1-\epsilon_{M},
\)
is complicated by a factor we call \emph{asymmetric feature correlation (AFC)}, and which we explain in~\Cref{sec:afc}.

This bound is tight, and we provide an example of a dataset and Merlin that achieve it in~\Cref{fig:afc}. The example shows a rather unintuitive image dataset with classes ``One fish or many fruit'' and ``One fruit or many fish''. Merlin selects a fish feature for the first and a fruit feature for the second class. The best strategy for Arthur is then to accept these features as proof for the respective class. The only images where this strategy fails are the two images with many fish or fruit, leading to a small $\epsilon_M$ no matter what Morgana does. Note, however, that each individual feature appears once in each class, which means the precision is $0.5$ and the conditional entropy is $1$. On the other hand, when we restrict the dataset to all images with only a single fruit or fish as $D^\prime$, then  covers almost the whole dataset and restricted to $D^\prime$ the features determine the class completely. This illustrates why the restriction to $D^\prime$ is necessary: It allows us to connect the informativeness of a set of features (e.g., each fish feature) to the informativeness of each single feature.

\subsection{Asymmetric Feature Correlation}\label{sec:afc} AFC describes a possible quirk of datasets, where a set of features is strongly concentrated in a few data points in one class and spread out over almost all data points in another. We give an illustrative example in~\Cref{fig:afc}. 
If a data space $\mathfrak{D}$ has a large AFC $\kappa$, Merlin can use features that individually appear equally in both classes (low precision) to indicate the class where they are spread over almost all points. Morgana can only fool Arthur in the other class where these features are highly concentrated, thus only in a few data points. This ensures a small $\epsilon_M$ even with uninformative features. 

For a given set of features $F\subset D_p$, we define the set
\[
 F^\ast := \skl{\bfx \in D~|~ \exists~ \bfz \in F: \bfz \subseteq \bfx},
\]
i.e., all data points that contain a feature from $F$.
\begin{dfn}[Asymmetric feature correlation]\label{def:afc}
Let $(D,\CD,c)$ be a two-class data space, then the asymmetric feature correlation $\kappa$ is defined as
\[
\kappa = \max_{l\in\skl{-1,1}} \max_{F \subset D_p} \E_{\bfy \sim \CD_l|_{F^*}}\ekl{\max_{\substack{\bfz \in F \\ \text{s.t. }\bfz \subseteq \bfy}}\kappa_l(\bfz, F)}
\]
with
\[
  \kappa_l(\bfz, F) = \frac{\P_{\bfx \sim \CD_{-l}}\ekl{\bfz \subseteq \bfx \,\middle|\, \bfx \in F^*}}{\P_{\bfx \sim \CD_l}\ekl{\bfz \subseteq \bfx \,\middle|\, \bfx \in F^*}}.
\]
\end{dfn}

We derive this expression in more detail in~\Cref{apx:afc}, but give an intuition here.
The probability $\P_{\bfx \sim \CD_{l}}\ekl{\bfz \subseteq \bfx \,\middle|\, \bfx \in F^*}$ for $\bfz\in F$ is a measure of how correlated the features are. If all features appear in the same data points this quantity takes a maximal value of 1 for each $\bfz$. If no features share the same data point the value is minimally $\frac{1}{\bkl{F}}$ for the average $\bfz$.
The $\kappa_l(\bfz, F)$ thus measures the difference in correlation between the two classes. In the example in~\Cref{fig:afc} the worst-case $F$ for $l=-1$ correspond to the ``fish'' features and $\kappa_l(\bfz, F)=6$ for each feature.
To take an expectation over the features $\bfz$ requires a distribution, so we take the distribution of data points that have a feature from $F$, i.e., $\bfy \sim \CD_l|_{F^*}$, and select the worst-case feature from each data point. Then we maximise over class and the possible feature sets $F$. 
Since in~\Cref{fig:afc}, the ``fish'' and ``fruit'' features are the worst case for each class respectively, we arrive at an AFC of 6.

Though it is difficult to calculate the AFC for complex datasets, we show that it can be bounded above by the maximum number of features per data point in $D$.
\begin{restatable}{lem}{afcbound}\label{lem:afc_bound}
  Let $\mathfrak{D}$ be a two-class data space with AFC of $\kappa$. Let $K = \max_{\bfx \in D}\, \bkl{\skl{\bfz \in D_p \,|\, \bfz \subseteq \bfx}}$ be the maximum number of features per data point. Then
 \(
  \kappa \leq K.
 \)
\end{restatable}
We prove this in~\Cref{apx:theory}. $K$ depends on the type of features one considers, e.g., for image data a rectangular cutout of given size, $K \sim d$, when any subset of pixels is allowed, then $K\sim 2^d$. See also~\Cref{apx:theory} for an example dataset with an exponentially large AFC.

\subsection{Realistic Algorithms and Relative Success Rate}\label{ssc:realistic}

In \Cref{thm:minmax}, we make use of a perfect Morgana. For complex classifiers this implies exhaustive search, which is indeed possible for low-dimensional data often used in recruitment and criminal justice, where interpretability is crucial. Consider the UCI Census Income dataset~\cite{Dua:2019} with 14 dimensions. When restricting features to a maximal size of seven, the search space is at most $ \binom{14}{7} = 3432$, well within range for exhaustive search. 
Contrary to this, modelling the UCI data distribution explicitly is still an involved task, and when done incorrectly, leads to incorrect explanations~\cite{frye2020shapley}. 

However, we also aim to apply our setup to high-dimensional datasets, where exhaustive search is not possible.
It turns out we can relax the requirement for Morgana to play optimally in two important ways: (i) She only has to find the features that can also be found by Merlin (ii) She only has to do so with a success rate comparable to Merlin.  

 \begin{dfn}[Relative Success Rate]
  Let $\mathfrak{D}=(D, \CD, c)$ be a two-class data space. Let $A\in \CA$ and $M, \morg \in \CM(D)$.
 Then the relative success rate $\alpha$ of $\morg$ with respect to $A,M$ and $\mathfrak{D}$ is defined as
 \[
     \alpha := \min_{l\in \skl{-1,1}} \frac{\P_{\bfx\sim \CD_{-l}}\ekl{A(\morg(\bfx))=l \,|\, \bfx \in F_l^\ast}}{\P_{\bfx\sim \CD_{l}}\ekl{A(M(\bfx))=l \,|\, \bfx \in F_l^\ast}},
 \]

  where $F_{l}^\ast := \skl{\bfx \in D \,|\, \exists \bfz\subseteq \bfx:~ \bfz\in M(D_l), A(\bfz)=l}$.
\end{dfn}

The set $F_{l}$ is the set of all features that Merlin uses in class $l$ to successfully convince Arthur, and $F_{l}^\ast$ is the set of all data points containing such a feature. Thus, we only evaluate Morgana's performance on data points where, in principle, she can identify one of Merlin's features. The question is then how the context of the other features makes this computationally easier or harder. We discuss this idea in more depth in~\Cref{apx:theory} and give a worst-case example in~\Cref{fig:relative_strength}.
We argue that realistically, we can assume a large $\alpha$ when using an algorithm for Morgana that is at least as powerful as the one for Merlin.
Together with the AFC, this allows us to state the following theorem.

\begin{restatable}{thm}{main}\label{thm:main}
Let $\mathfrak{D}=(D, \CD, c)$ be a two-class data space with AFC of $\kappa$ and class imbalance $B$. Let $A\in \CA$, and $M, \widehat{M}\in\CM(D)$ such that $\widehat{M}$ has a relative success rate of $\alpha$ with respect to $A, M$ and $\mathfrak{D}$.
Define

1. Completeness:
\[
\min\limits_{l \in \skl{-1,1}}\P_{\bfx \sim \CD_l}\ekl{A\kl{M(\bfx)} = c(\bfx) } \geq 1- \epsilon_c,
\]
2. Soundness 
\[
\max\limits_{l \in \skl{-1,1}} \P_{\bfx \sim \CD_l}\ekl{A\kl{\morg(\bfx)} = -c(\bfx) } \leq  \epsilon_s.
\]

Then it follows that
\[
 \ap_{\CD}(M) \geq 1 - \epsilon_c - \frac{ \kappa \alpha^{-1}\epsilon_s}{1 - \epsilon_c+ \kappa \alpha^{-1}B^{-1}\epsilon_s}.
\]
\end{restatable}

The proof is provided in~\Cref{apx:theory}. This bound is asymptotically tight. Consider again the example in~\Cref{fig:afc}, but generalised to $n$ fish and fruit features instead of six. Then we have $\ap_{\CD}(M)=\frac{1}{2}$, $\epsilon_c = \frac{1}{n+1}$, $\epsilon_s = \frac{1}{n+1}$, $\kappa = n$, $B = 1$ and $\alpha = 1$, since both Merlin and Morgana will always succeed in finding a feature to convince Arthur if it exists in the data point.
Then we have
\[
\frac{1}{2} \geq 1 - \frac{1}{n+1} - \frac{n\frac{1}{n+1}}{1 - \frac{1}{n+1}+n\frac{1}{n+1}} = \frac{1}{2} - \frac{1}{n+1},
\]
which approaches equality as the number of features grows larger.

The core assumption we make when comparing our lower bound with the measured average precision in~\Cref{sec:numerics} is the following:
\begin{ass}\label{ass:realistic}
The AFC $\kappa$ of $\mathfrak{D}$ and the relative success rate $\alpha$ of $\morg$ w.r.t.\ A, M, $\mathfrak{D}$ are $\CO(1)$.
\end{ass}

Currently, we cannot confirm whether a dataset exhibits a small AFC. However, we conjecture that, even in cases where a dataset may include a feature set that realises large AFC, identifying such a set poses a computationally challenging task for Merlin. We leave this issue open for further investigation.

\paragraph{Finitely Sampled and Biased Dataset} We usually have access to only finitely many samples of a dataset. Additionally, the observed samples can be biased as compared to the true distribution. We prove bounds for both cases in~\Cref{apx:finit_and_biased_datasets}. We show that any exchanged feature is either informative, or it is incorrectly represented in the dataset---thus highlighting the bias!

In conclusion, the theoretical results presented show that 
\begin{enumerate*}[label=(\roman*)]
    \item For optimal feature selectors and classifiers, we can guarantee highly informative features without the need to model the data distribution, see~\Cref{thm:minmax}.
    \item For suboptimal agents we can still assure informative features as long as the success probability of Morgana is comparable to the one of Merlin, see~\Cref{thm:main}.
    \item We can certify feature quality with measurable quantities soundness and completeness.

\end{enumerate*}

\begin{figure*}[t]
    \centering
     \resizebox{0.99\textwidth}{!}{

\input{img/census_data/graphic}

        }
        \vspace{-0.4cm}
    \caption{\label{fig:gender}\small{
    The objective is hiring only men while hiding ``sex'' as the explanation. 1. 
    No soundness is required ($\gamma = 0$). 
    a) Merlin has no punishment for showing ``sex'' ($\beta = 0$). He sends Arthur the ``sex'' feature and they discriminate successfully (high completeness). b) Merlin is incentivised not to use ``sex'' ($\beta > 0$). He successfully communicates the ``sex'' to Arthur via different features, here ``hours per week'' and ``education''. Morgana can exploit this strategy with the same features switched.
    2. High soundness is now required ($\gamma = 0.67$). Merlin either a) shows the ``sex'' feature despite the punishment ($\beta=3.5$) and achieves high completeness, or b) hides the ``sex'' feature ($\beta=10$) and reduces completeness to below 50\%, ceasing the discrimination.
    }}  
\end{figure*}
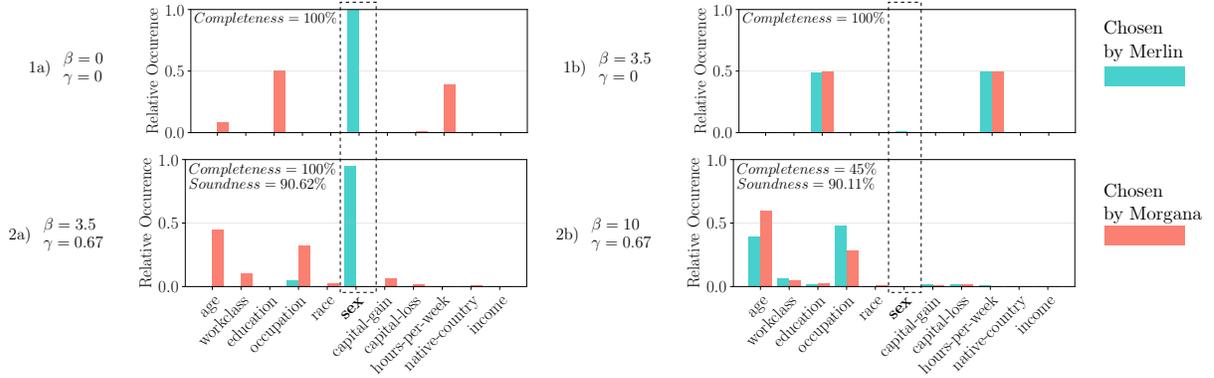

\section{Numerical Implementation}
\label{sec:numerics}

Let us describe how to train the agents Arthur, Merlin, and Morgana in a general $n$-class interactive learning setting for image data of dimension $d$, where $n, d \in \N$. 
We explain in~\Cref{apx:game_design} why we chose a multi-class neural network for Arthur and compare with the approaches of \cite{chang2019game} and \cite{anil2021learning}.
The training process for tabular data is equivalent, a detailed overview is provided in the appendix.\footnote{The code is available at \url{https://github.com/ZIB-IOL/merlin-arthur-classifiers}.} 

Arthur is modelled by a feed-forward neural network. He returns a probability distribution over his possible answers, so let $A: [0,1]^d \rightarrow [0,1]^{(n+1)}$, corresponding to the probabilities of stating a class or ``Don't know''.
The provers select a set $S$ of at most $k$ pixels from the image via a \emph{mask} $\bfs \in B_k^d$, where  $B_k^d$ is the space of $k$-sparse binary vectors of dimension $d$. A masked image $\bfs\cdot\bfx$ has all its pixels outside of $S$ set to a baseline or a random value.
We define the Merlin-loss $L_M$ as the cross-entropy loss with regard to the correct class, whereas the Morgana-loss $L_{\morg}$ considers the total probability of either answering the correct class or the ``I don't know'' option, so
\begin{align*}
& L_M(A,\bfx,\bfs) = -\log\kl{A(\bfs\cdot\bfx)_{c(\bfx)}} \quad\text{and}\\
&L_{\morg}(A,\bfx,\bfs) = -\log\kl{A(\bfs\cdot\bfx))_{0} + A(\bfs\cdot\bfx)_{c(\bfx)}}.
\end{align*}
Arthur's total loss is then $L = \E_{\bfx \sim \mathcal{D}}[L(\bfx)]$, where
\[
 L(\bfx) =  (1- \gamma) L_{M}(A,\bfx,M(\bfx)) + \gamma L_{\morg}(A,\bfx,\morg(\bfx)),
\]
and $\gamma \in [0,1]$ is a tunable parameter. In our experiments, we choose $\gamma > 0.5$ since we always want to ensure good soundness.
Note that Merlin wants to minimise $L_M$, whereas Morgana aims to maximise $L_{\morg}$. In an ideal world, they would solve
\begin{equation}
\label{eq:losses}
\begin{aligned}
 &M(\bfx) = \argmin_{\bfs \in B_k^d} L_M(A,\bfx, \bfs) \quad \text{and}\\
 &\morg(\bfx) = \argmax_{\bfs \in B_k^d} L_{\morg}(A,\bfx,\bfs).
 \end{aligned}
\end{equation}
The above solutions can be obtained either by solving the optimisation problem (Frank-Wolfe solver~\citep{macdonald2021interpretable}) or by using U-Nets to predict the solutions. We describe the training algorithm in~\Cref{alg:training}:  For $N$ epochs we iterate over the dataset and alternately train Arthur on masked images and on the original, unmasked images. The update steps for Merlin and Morgana (steps 8 and 9) only apply when the feature selectors are realised by U-Nets. 

 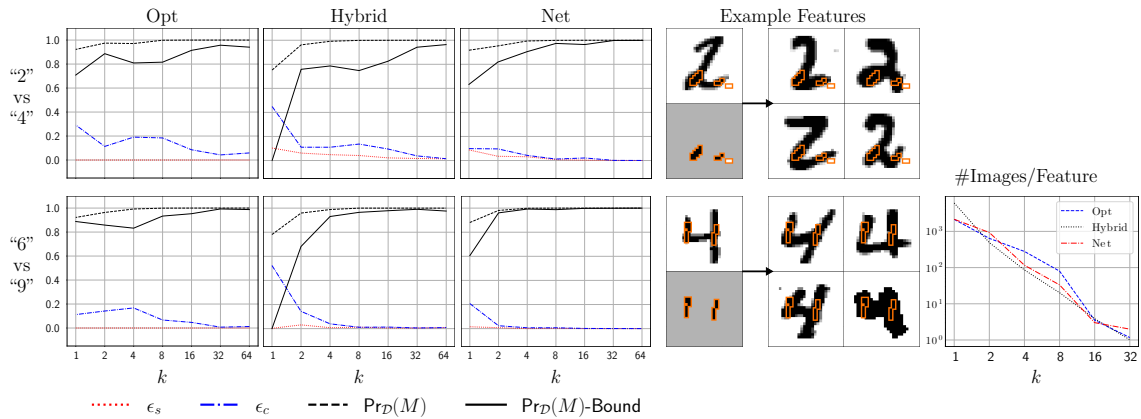
\begin{figure*}
 \centering
\resizebox{0.9\textwidth}{!}{

\input{img/bound_comparison/new_graphic}
 }
 \caption{\label{fig:bound}\small{\emph{Left:} For four different setups of Merlin and Morgana, we compare the lower bound on $\ap_{\CD}(M)$ with the experimental results on the MNIST dataset. The top row is for the labels $\skl{\text{``2''},\text{``4''}}$, and the bottom row for $\skl{\text{``6''}, \text{``9''}}$. The bound is tight for large masks, but loosens sharply for very small mask sizes $k$.
 \emph{Middle:} Examples of the features selected by Merlin for two images. For the ``4'' feature, there are 13 images in MNIST that share it, all of them of class ``4'' (we show four here). For the ``2'' there are 16 images, all of them in ``2''. \emph{Right:} The average number of images found for a feature selected by Merlin. These images were used to estimate $\ap_{\mathcal{D}}(M)$.
 }}
 \end{figure*}

\begin{algorithm}[H]
    \centering
    \begin{algorithmic}[1]
        \STATE \textbf{Input:} dataset: $D_{\text{train}}$, Epochs: $N$, $\gamma$
        \STATE \textbf{Output:} Classifier Network (A), Optional: Masking Networks Merlin $(M)$ and Morgana $(\morg)$
        \FOR{$i \in [N]$}
        \FOR{$\bfx_j, l_j \in D_{\text{train}}$}
            \STATE $\bfs_{M} \gets M(\bfx_j, l_j), \bfs_{\morg} \gets \morg(\bfx_j, l_j)$ 
            \STATE $L_A(\bfx_j, l_j) = (1- \gamma)L_M(A(\bfs_M \cdot \bfx_j), l_j) + \gamma L_{\morg}(A(\bfs_{\morg} \cdot \bfx_j), l_j)$
            \STATE $\theta_A \leftarrow \theta_A - \alpha \nabla_{\theta} L_A(\bfx_j, l_j)$
                \STATE $\theta_M \leftarrow \theta_M - \alpha \nabla_{\theta} L_M(A(M(\bfx_j)\cdot \bfx_j), l_j)$ %
                \STATE $\theta_{\morg} \gets \theta_{\morg} - \alpha \nabla_\theta L_{\morg}(A(\morg(\bfx_j)\cdot \bfx_j), l_j)$ 
        \ENDFOR
        \FOR{$\bfx_j, l_j \in D_{\text{train}}$}

            \STATE $\theta_A \gets \theta_A - \alpha \nabla_{\theta} L(A(\bfx_j), l_j))$ 
        \ENDFOR
    \ENDFOR
    \end{algorithmic}
    \caption{Merlin-ArthurTraining}
    \label{alg:training}
\end{algorithm}

\subsection{Preventing Manipulation}\label{sec:cheating}

XAI algorithms without guarantees cannot exclude the possibility for manipulation. Indeed, arbitrarily changing the interpretation by slightly modifying the classifier has been demonstrated for many XAI approaches.
\citeauthor{slack2020fooling} fool LIME and ShAP by making use of the fact that these methods sample off-manifold evaluations of the classifier. We are robust against this approach, since the Merlin-Arthur classifier only takes on-manifold inputs, see~\Cref{apx:off_manifold} for further details. \citeauthor{dimanov2020you}, \citeauthor{heo2019fooling} and \citeauthor{anders2020fairwashing} optimise manipulated classifier networks to give the desired explanations by penalising any deviation.
They introduce a penalty term in the network objective that measures the distance between the produced and the desired relevance scores. 
The equivalent for our setup is to put a penalty on Merlin to hide the true (potentially biased) explanations in the exchanged features.
Our theory predicts this scheme to be unsuccessful, and we confirm numerically that either: 
\begin{enumerate*}[label=(\roman*)]
\item The bias becomes visible, 
\item Morgana can exploit the setup, i.e., soundness is low or
\item the completeness, i.e., prediction accuracy, drops. 
\end{enumerate*}

We demonstrate this on a modified UCI Census Income dataset\footnote{For this experiment, we have removed the “relationship” and “marital status” features, since they strongly correlate with sex. See~\Cref{apx:uci_appendix} for experiments that include them.}. We train a classifier to discriminate by ``sex'' --- one of the remaining 12 features. We change the Merlin-loss to
    \[
 L^\prime_M(A,\bfx,\bfs) := L_M(A,\bfx,\bfs) + \beta \nkl{\bfs_{\text{sex}}},
\]
where $\beta>0$, to train Merlin to hide the sensitive ``sex'' feature. We then train the Merlin-Arthur classifier to accept only men and set $k=1$ to only send one feature. We compare a setup where soundness is ignored to one where we choose $\gamma$ such that a soundness of at least 90\% is achieved.

The results are illustrated in~\Cref{fig:gender}. When soundness is ignored ($\gamma=0$), setting $\beta=3.5$ causes Merlin to hide the ``sex'' feature and communicate ``sex'' via selecting ``education'' for men, ``working hours'' for women.
When soundness is required ($\gamma=0.67$), Merlin starts to reveal the ``sex'' feature again. Increasing the penalty to $\beta=10$, causes him to hide ``sex'', but now the prediction accuracy reduces to $0.45$, i.e., close to chance and thus ceasing the discrimination. When high soundness is required, Merlin is unable to both discriminate and hide his bias. 

In~\Cref{apx:off_manifold}, we apply this approach to different post-hoc XAI methods and show that they are indeed susceptible to manipulation.

\subsection{Evaluation of Theoretical Bounds}
\label{ssc:bound}

 Low-dimensional datasets serve to evaluate our mutual information bounds. For small features there are multiple data points that contain them and the ground truth information of the feature can be estimated.
 We evaluate \Cref{ass:realistic} on the MNIST dataset restricted to two classes for three setups of feature selectors, one with Frank-Wolfe optimisers (\textbf{Opt}), one where Merlin is a U-Net and Morgana an optimiser (\textbf{Hybrid}), and with U-Nets for both (\textbf{Net}). We want to stress that this is not a comparison to other XAI methods, which do not generally make predictions about the precision of the highlighted features. Standard classifiers might be easier to train or achieve higher accuracy compared to a Merlin-Arthur classifier. But they are not interpretable, as there are no post-hoc methods robust to manipulation.

In~\Cref{fig:bound}, the lower bound is tight for larger masks, but drops off when $k$ is small. One reason is that for small masks, Arthur tends to give up on one of the classes, while keeping the completeness in the other class high. Regularising Arthur to maintain equal completeness is a potential solution. When Merlin and Morgana are realised by the same method (both optimisers or NNs), the bound is the tightest. In our hybrid approach, the bound is pessimistic since Merlin is at a disadvantage. He needs to learn on the training set to select good features, whereas Morgana can optimise directly on the test set.
In~\Cref{apx:numerics}~\Cref{fig:avp_error}, we show error bars sampled over 10 training runs. Our lower bound is always below the empirical estimate, which is evidence that~\Cref{ass:realistic} is correct.
This must be evaluated more extensively on different datasets.

\section{Discussion and Limitations}\label{sec:discussion}

We can draw a connection between soundness and Adversarial Robustness~\citep{goodfellow2014explaining}.
Consider the generation of adversarial examples :
\[
 \boldsymbol{\delta}^\ast = \argmin_{\nkl{\boldsymbol{\delta}}\leq \epsilon} L(\bfx + \boldsymbol{\delta}).
\]
The intuition is that minuscule changes to the input, imperceptible to humans, should not change the classifier decision.
Likewise, the intuition behind soundness, i.e., robustness with respect to Morgana, is that hiding parts of an object should not convince the classifier of a different class.
At most, one could hide the whole object, which is reflected in the ``Don't know!'' option. 
In this sense, soundness should be expected of classifiers that generalise to partially hidden objects.

AFC seems to be more generally relevant to interactive interpretability, even in different setups than ours. In \cite{yu2019rethinking}, a prover sends part of an image to a cooperative classifier and the rest to an adversarial classifier. The goal is to allow correct classification for the cooperator and prevent it for the adversary. However, as in our example in~\Cref{fig:afc}, the prover can use completely uninformative features (``fish'' and ``fruit'') and the adversary is unable to exploit this except for a small number of image, inversely proportional to the AFC. This means the AFC would need to be part of an information bound, assuming it is formulated in terms of the accuracy of the cooperator and adversary on the whole dataset. For details, see~\Cref{apx:adv_classification}.

High completeness and soundness can be mandated for commercial classifiers, e.g., in the context of hiring decisions with past decisions by the Merlin-Arthur classifier as ground truth. An auditor uses their own Morgana to verify sufficient soundness. If Arthur is sound, the features selected by Merlin are verifiably the basis of the hiring decisions and can be inspected for protected attributes, e.g., race, sex or attributes that strongly correlate with them~\citep{mehrabi2021survey}. This hinges on the fact, as explained in in~\Cref{ssc:realistic} in terms of the relative success rate, whether the Morgana used by the auditor is computationally as powerful as the Merlin used by hiring department.

However, simply identifying features with high mutual information does not necessarily point to causal mechanisms, since they can include spurious correlations. While the adversary prevents such correlations in the masks, they might still be present in the original data. In the UCI dataset, the removed features ``marital status'' and ``relationship'' are correlated with sex and thus can be used by Merlin to communicate ``sex'' to Arthur when included, see~\Cref{apx:uci_appendix}. It is up to society to determine whether the exchanged features constitute discrimination. This is a problem shared generally by interpretability tools, however, there has been progress to adapt interactive classification to find causal features~\cite{chang2020invariant}.

In future work, we aim to move beyond the restriction to the deterministic two-class case. We discuss the training stability of the three-player game and numerical challenges in~\Cref{apx:stability}.

\section{Conclusion}\label{sec:conclusion}

We extend the framework of adversarial interactive classifiers to provide quantitative mutual information bounds on the exchanged features in terms of the measurable criteria completeness and soundness.
We also move beyond the common assumptions of optimally playing agents and of feature independence. Instead, we consider the relative strength of the provers and introduce Asymmetric Feature Correlation, which captures the relevant aspect of the feature dependence. 
Finally, we evaluate our results on the UCI Census Income and MNIST datasets. Our experiments show that the Merlin-Arthur classifier can prevent manipulation that is successful for other XAI methods, and that our theory matches well with our numerics.

\paragraph{Acknowledgement}
Funded by the Deutsche Forschungsgemeinschaft (DFG, German Research Foundation) under Germany's Excellence Strategy – The Berlin Mathematics Research Center MATH+ (EXC-2046/1, project ID: 390685689).

%% file: img/census_data/graphic.tex
\Large{
  \centering
  \begin{tabular}{@{}r@{\,\,}r@{}l@{}}
    1a) 
  \begin{tabular}{l}
       $\beta=0$  \\
       $\gamma=0$
  \end{tabular}
  \raisebox{-.5\height}{
  \includegraphics[width=13cm]{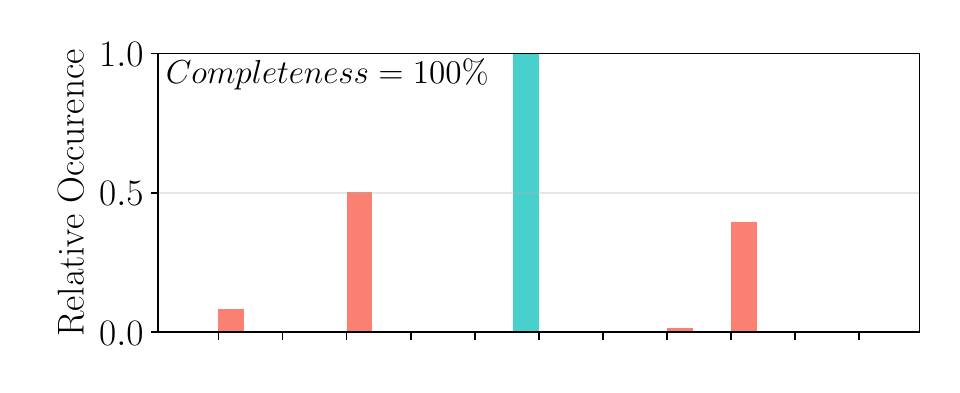}
  }
  &
    1b) 
  \begin{tabular}{l}
       $\beta=3.5$  \\
       $\gamma=0$
  \end{tabular}
  \raisebox{-.5\height}{
  \includegraphics[width=13cm]{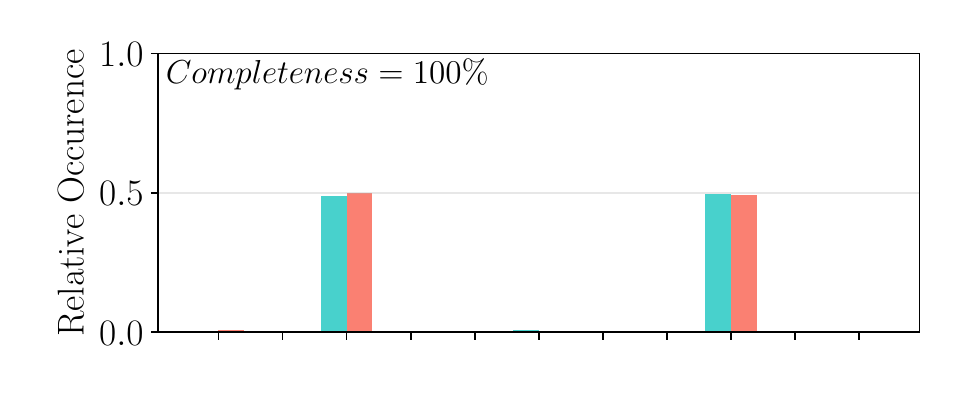}
  }
  &
  \raisebox{-.2\height}{
  \includegraphics[height=2cm]{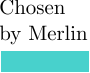}
  }
  \\[-1.5em]
    % c) $\beta=1, \gamma=1$ & d) $\beta=10, \gamma=1$
%   \\
    \hspace{1.5cm} 2a) 
  \begin{tabular}{l}
       $\beta=3.5$  \\
       $\gamma=0.67$
  \end{tabular}
    \raisebox{-.6\height}{
  \includegraphics[width=13cm]{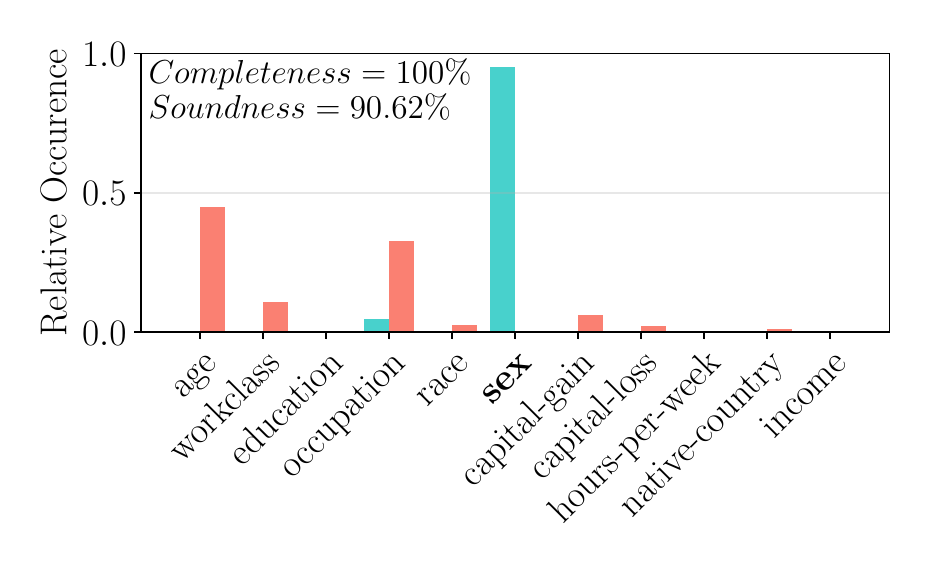}
  }
  &
    2b) 
  \begin{tabular}{l}
       $\beta=10$  \\
       $\gamma=0.67$
  \end{tabular}
  \raisebox{-.6\height}{
  \includegraphics[width=13cm]{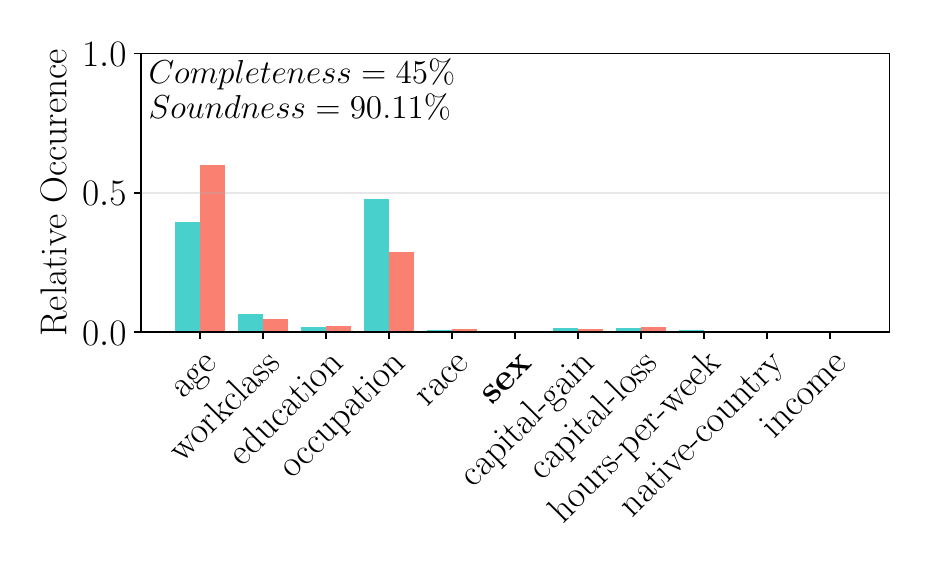}
  }
  &
  \raisebox{-.1\height}{
  \includegraphics[width=3cm]{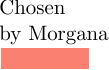}
  }
  \\
  
       % \multicolumn{2}{c}{
       % \includegraphics[width=13cm]{img/census_data/legend.pdf}}
    \end{tabular}
}

\begin{tikzpicture}[overlay]

\draw [draw=black,dashed] (-26.9,-3) rectangle (-25.8,5.9);
\draw [draw=black,dashed] (-10.1,-3) rectangle (-9.1,5.9);

\end{tikzpicture}

%% file: img/bound_comparison/new_graphic.tex
\Large{
  \centering
  \begin{tabular}%{@{}c@{\,}c@{}c@{\;}c@{\;}c@{\hspace{0.4cm}}c@{}m{0.6cm}c@{\,}}
  {@{}c@{\,}c@{}c@{\;}c@{\;}c@{\hspace{0.4cm}}c@{}c}
  &
  &
  Opt
  &
  Hybrid
  &
  Net
  &
  Example Features
  &
  \\
  \begin{tabular}[x]{@{}c@{}}``2''\\vs\\``4''\\ \end{tabular}
  &
  \raisebox{-.5\height}{
  \includegraphics[height=4cm]{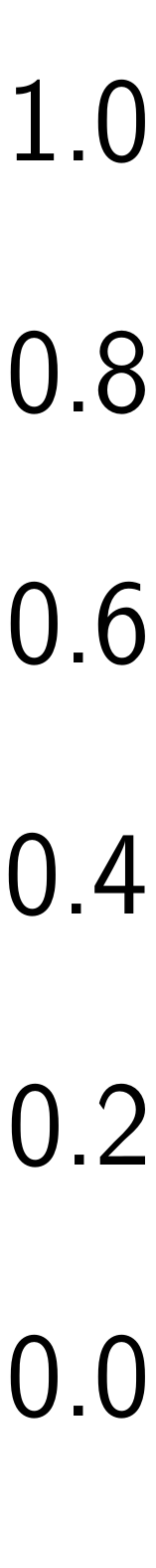}
  }
  &
  \raisebox{-.5\height}{\includegraphics[height=4cm]{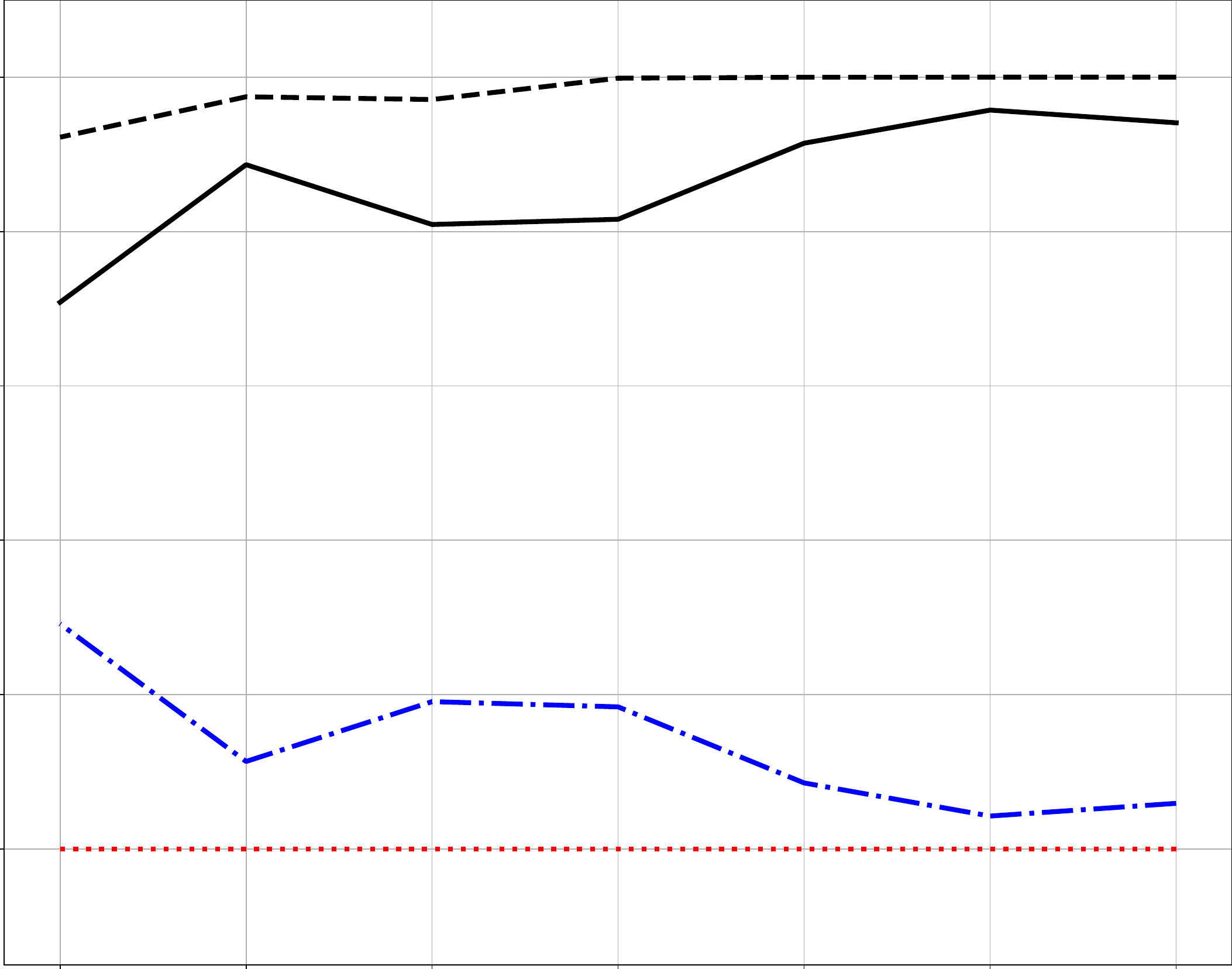}}
        &
        \raisebox{-.5\height}{\includegraphics[height=4cm]{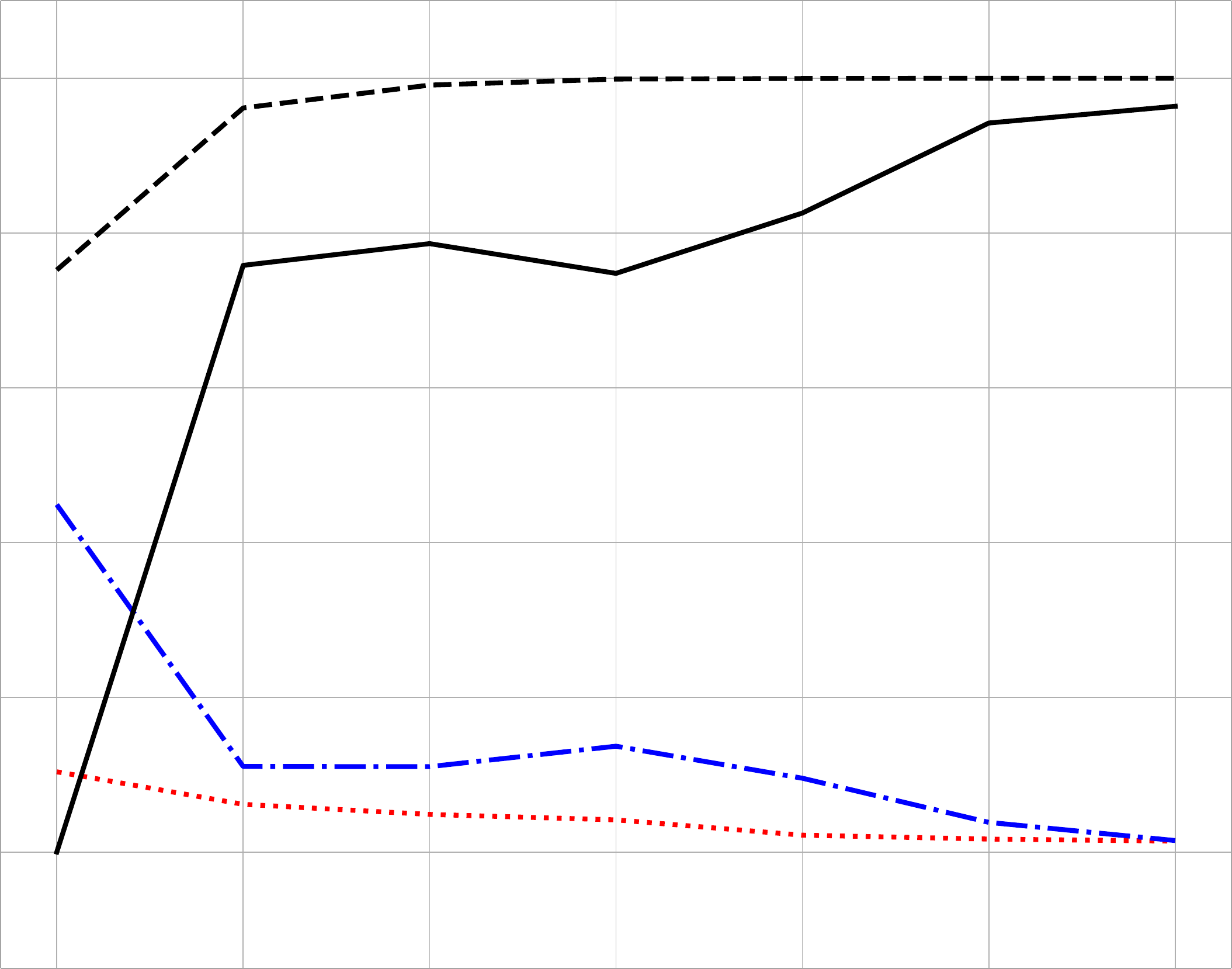}}
        &
\raisebox{-.5\height}{\includegraphics[height=4cm]{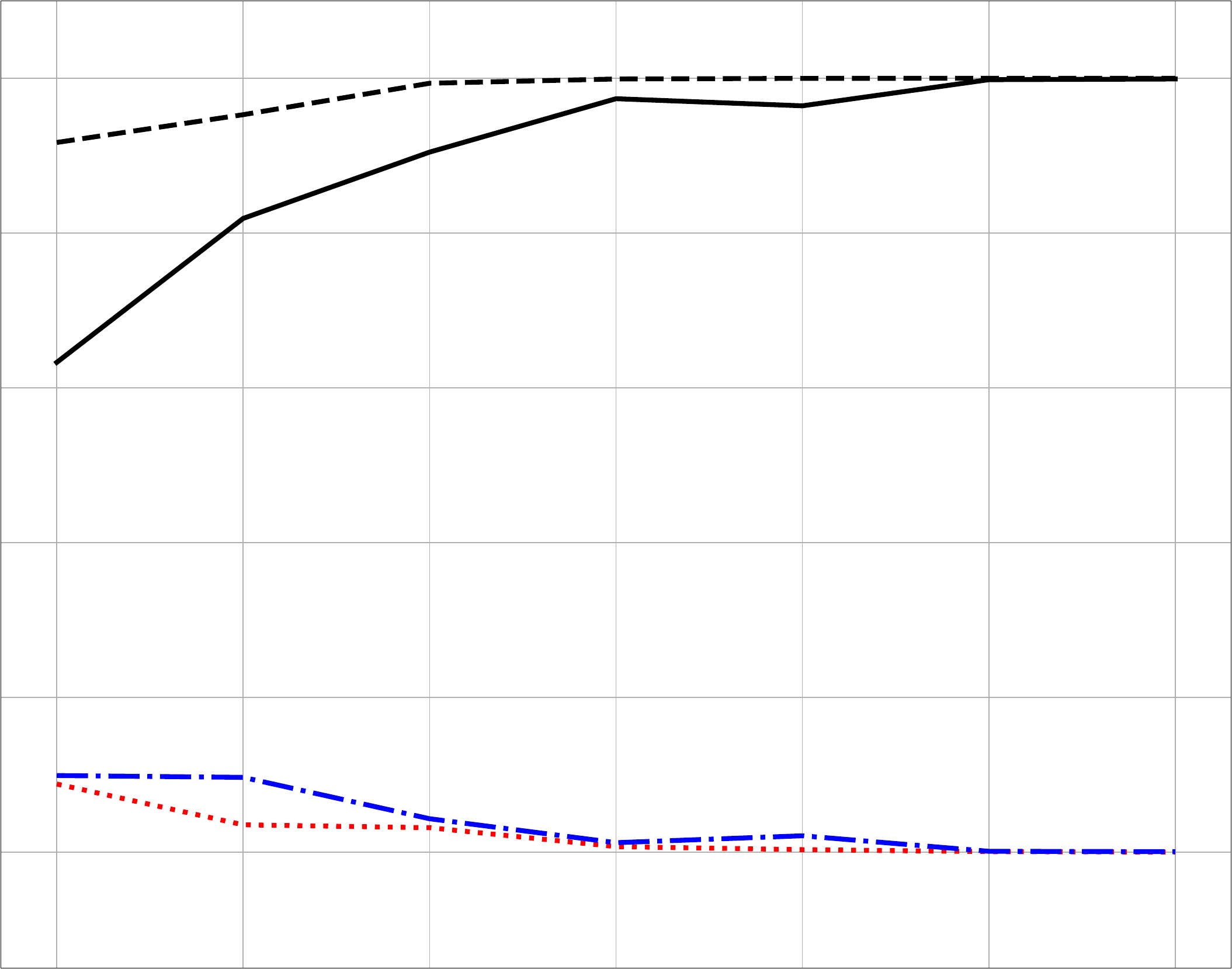}}
        &
\raisebox{-.5\height}{\includegraphics[height=4cm]{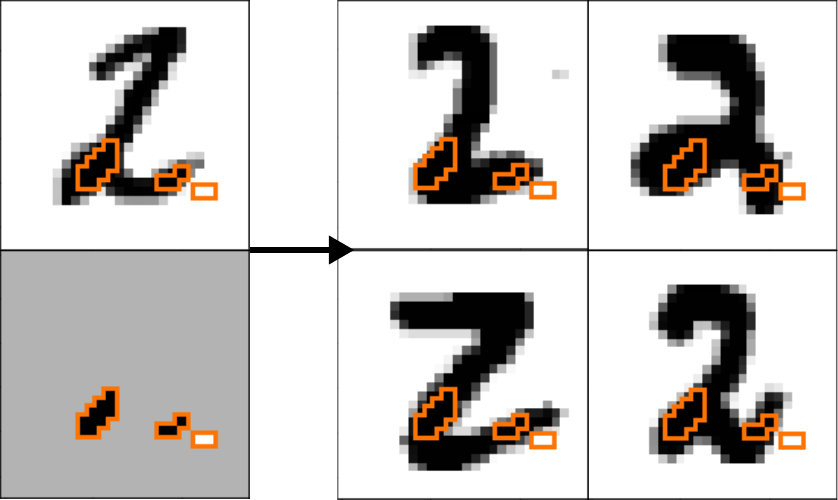}}
& 
\raisebox{-5.5\height}{$\#$Images/Feature}
        \vspace{0.5em}
        
        \\
        
       \begin{tabular}[x]{@{}c@{}}``6''\\vs\\``9''\\ \end{tabular}
        &
          \raisebox{-.5\height}{\includegraphics[height=4cm]{img/bound_comparison/bound_y_scale.pdf}}
                  &
        \raisebox{-.5\height}{\includegraphics[height=4cm]{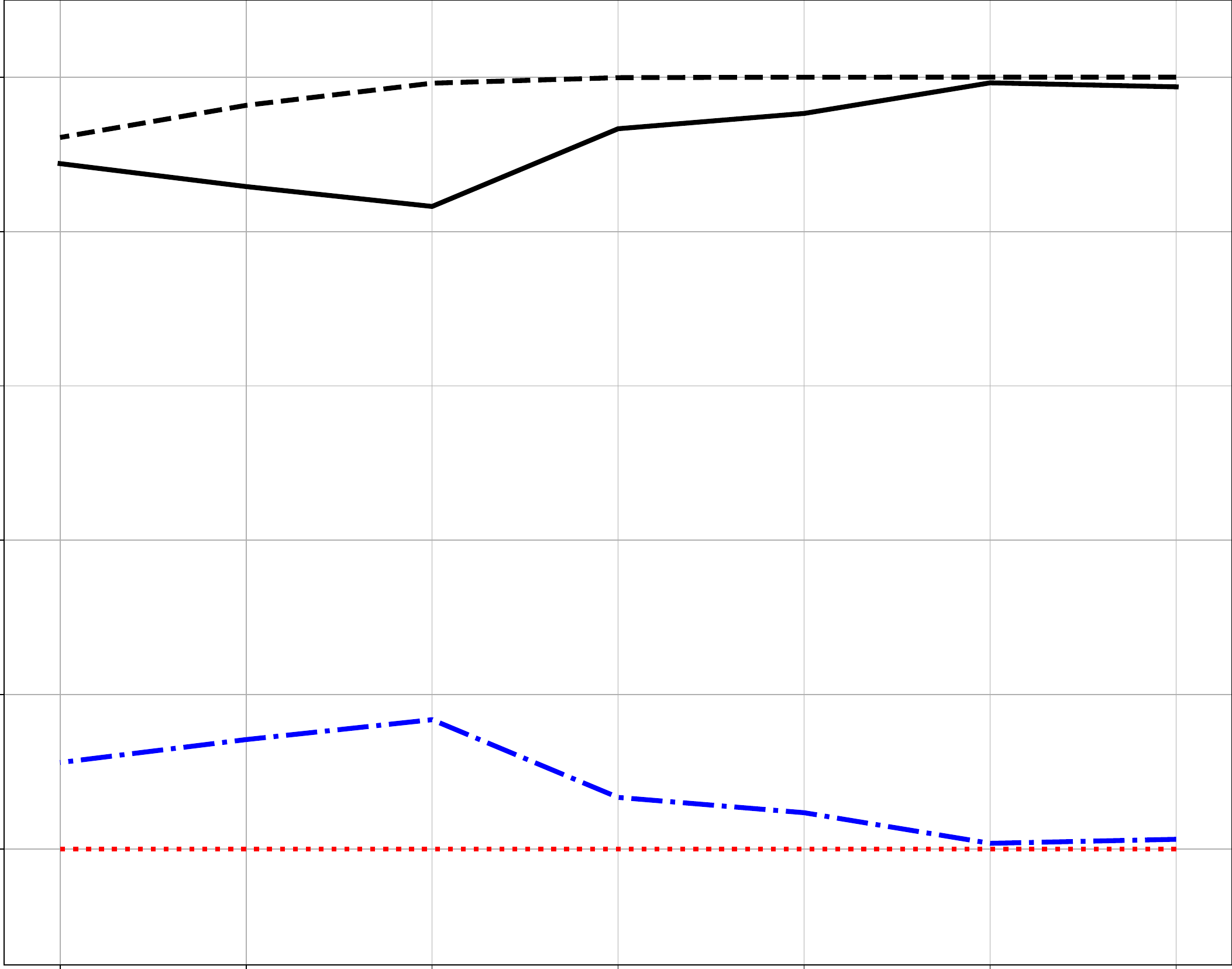}}
  &
  \raisebox{-.5\height}{\includegraphics[height=4cm]{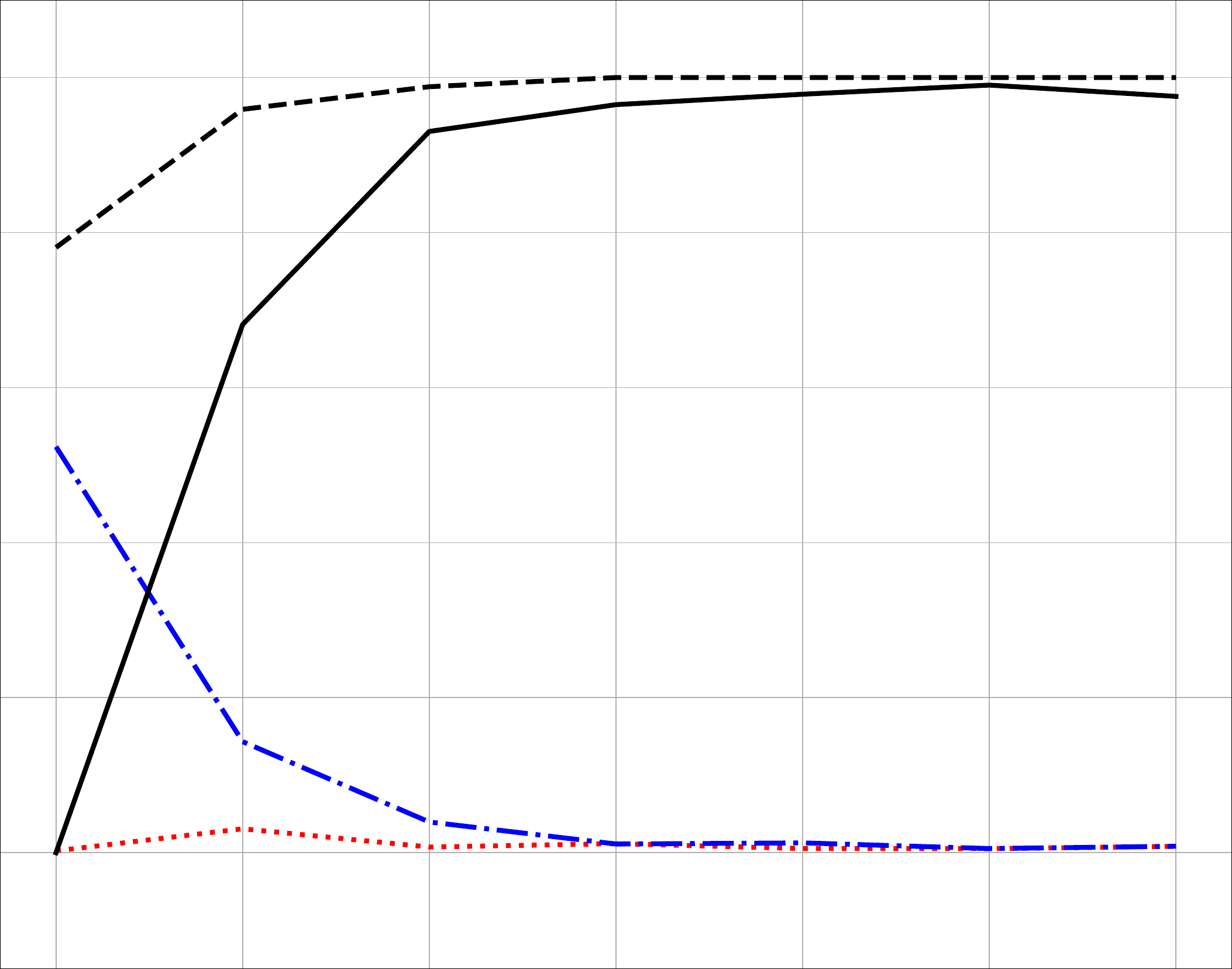}}
        &
\raisebox{-.5\height}{\includegraphics[height=4cm]{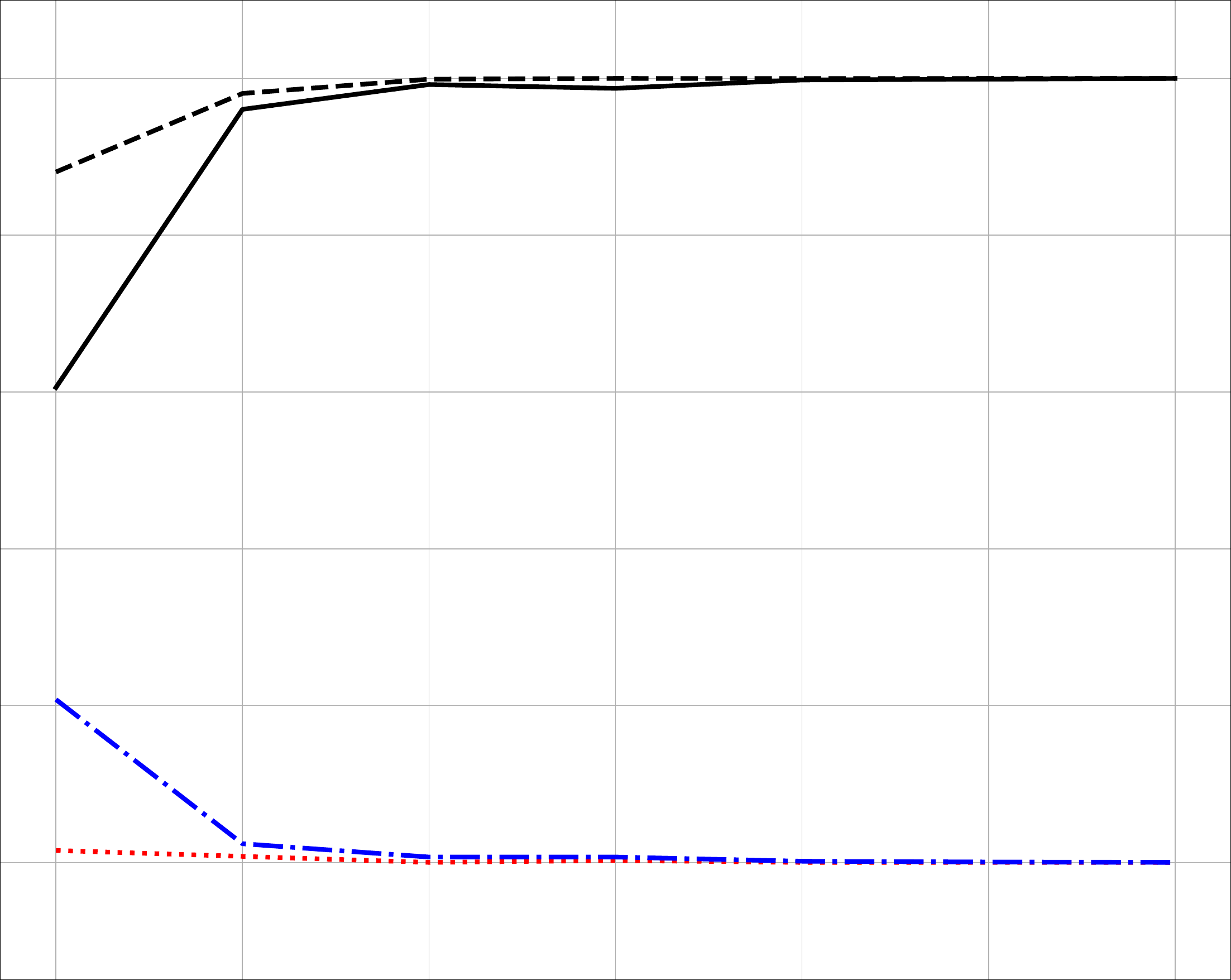}}
        &
\raisebox{-.5\height}{\includegraphics[height=4cm]{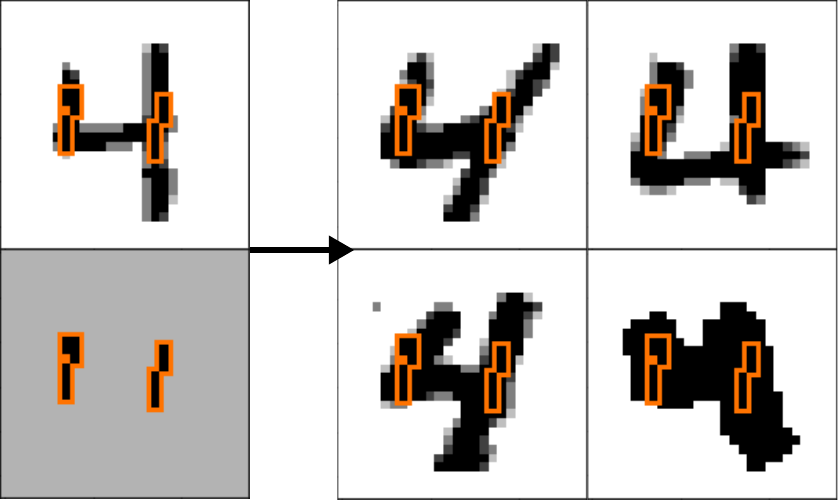}}
&
\raisebox{-.5\height}{
\includegraphics[height=4cm]{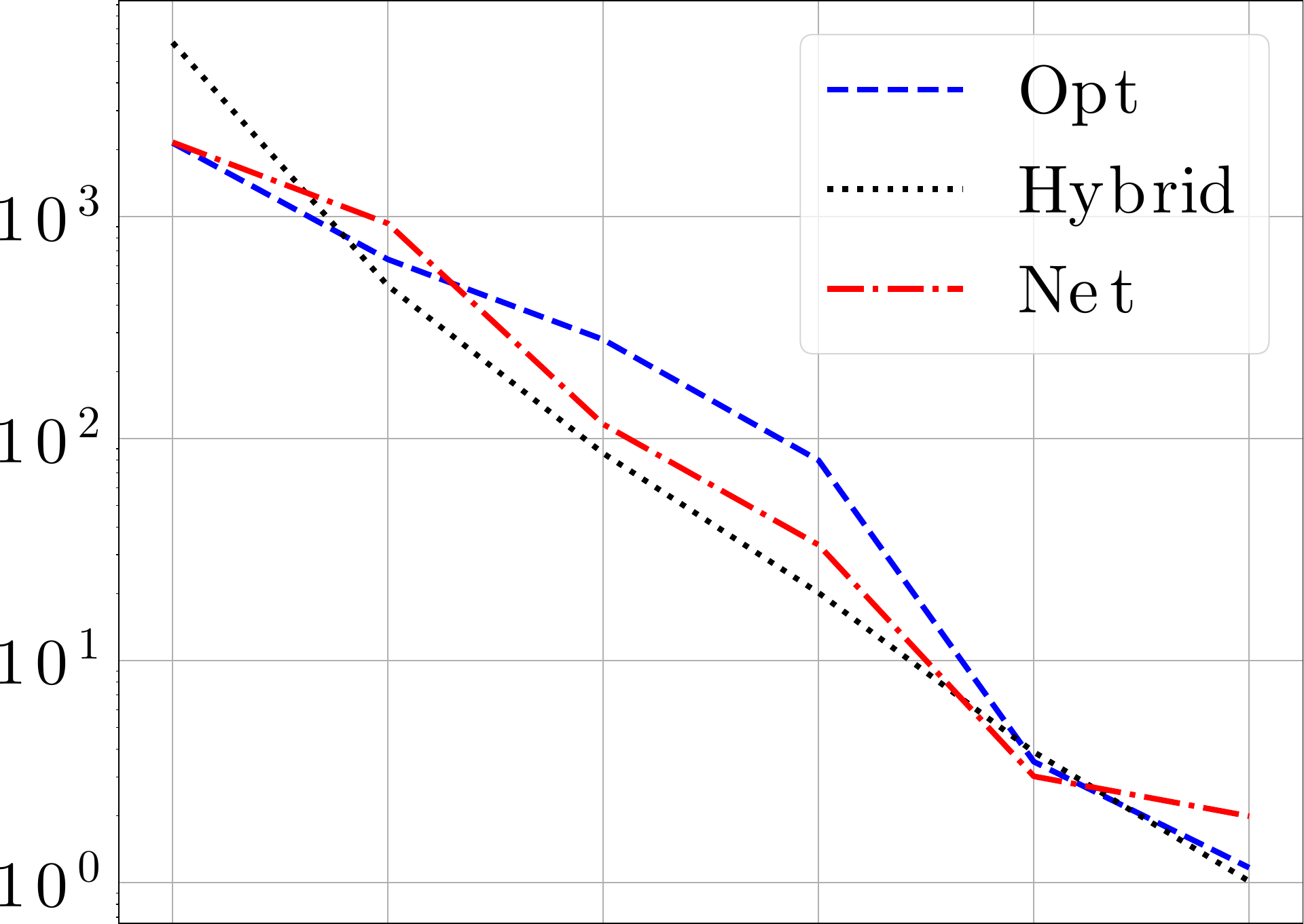}}
         \\
         &
         &
       \includegraphics[width=5cm]{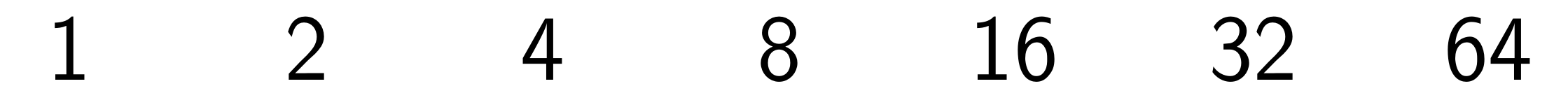}
                &
       \includegraphics[width=5cm]{img/bound_comparison/bound_x_scale.pdf}
                &
       \includegraphics[width=5cm]{img/bound_comparison/bound_x_scale.pdf}
                &
                &
                 \hspace{1.5em}\includegraphics[width=5.1cm]{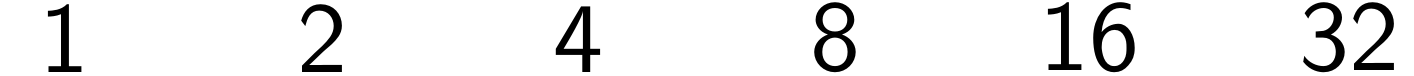}
             \\
         &
         &
       $k$
                &
       $k$
                &
       $k$
                &
                &
         \hspace{0.75em}$k$        
       \\
       &
       &
       \multicolumn{3}{c}{\includegraphics[width=15cm]{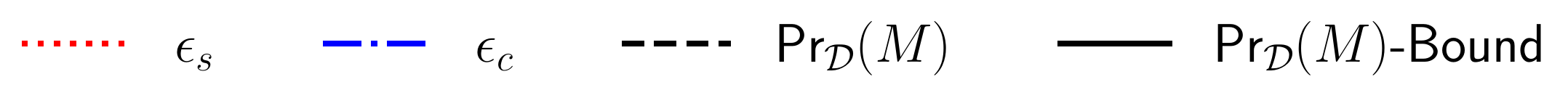}}
       &
    \end{tabular}
}

%% file: online_supplement.tex
\section{Conceptual Overview}\label{apx:concept}

Formal interpretability faces two hurdles: a complexity barrier, as well as a modelling problem. Here, we further explain these challenges and how we overcome them.
Furthermore, we also compare our architecture in detail with that proposed in ~\cite{chang2019game} and~\cite{anil2021learning} and provide an explanation for the differences in~\Cref{apx:game_design}.
For two alternative interactive classification setups with an adversary, the debate setup~\citep{irving2018ai} and the adversarial classifier setup~\cite{yu2019rethinking,dabowski2017unets}, we show that they cannot be used to derive bounds with the same generality as in our work. Instead, it would require stronger assumptions on either the classifier or the data space to exclude our counterexamples, see~\Cref{apx:alternative}.

\subsection{Computational Complexity}

Prime Implicant Explanations~\citep{shih2018symbolic}, a concept from logical abduction, can be efficiently computed for simple classifiers like decision trees~\citep{izza2020explaining} and monotonic functions~\citep{marques2021explanations}.
This concept has been extended to NNs~\citep{ignatiev2019abduction} in the form of \emph{probabilistic} prime implicants, which correspond to features with high precision.
However, it has been shown that even approximating small implicants within any non-trivial factor is $\SNP$-hard~\citep{waeldchen2021computational} for networks of two layers or more. 
In \cite{blanc2021provably}, the authors construct an algorithm that circumvents these hardness results by further relaxation of the problem. While this is a noteworthy theoretical breakthrough, the polynomial bound on the feature size grows so quickly with the dimension of the data space that the algorithm does not guarantee useful features for real-world data. For reasonably sized images, one would get guarantees only for features that cover the whole image.

We circumvent the hardness of this problem using a method that is very typical of Deep Learning: Use a heuristic and verify success afterwards! Our approach can be put alongside the regular training of classifiers, which is a theoretically hard problem as well. A heuristic like Stochastic Gradient Descent is not a priori guaranteed to produce a capable classifier. However, we can check the success of the procedure by evaluating the accuracy on a test dataset. In our case, training the Merlin-Arthur classifier is not guaranteed to converge to an equilibrium with informative features. But we can check whether this is the case via the test dataset, where soundness and completeness take the role of the accuracy.

\subsection{Modelling the True Data Distribution}\label{apx:modelling}

We introduce Merlin-Arthur classification as it provides us a way to measure the feature quality via the completeness and soundness values over a test dataset.
This would not be necessary if we could directly measure the feature quality over the dataset (though it would still be faster than measuring every individual feature).
The reason we need the Merlin-Arthur setup is that for general datasets the conditional entropy
\[
 H_{\bfy\sim\CD}(c(\bfy)\,|\,\bfz \subseteq \bfy) = H_{\bfy\sim\CD|_{\bfz \subseteq \bfy}}(c(\bfy)),
\]
is difficult to measure, since we do not generally know the conditional distribution $\CD|_{\bfz \subseteq \bfy}$.
This measurement is possible for MNIST for small features since the dataset is very simple. However, for more complex data, a feature which is large enough to be indicative of the class will in all likelihood not appear more than once in the same dataset. We will now discuss some existing approaches that aim to approximate the conditional data distribution and what problems they face.

Modelling the conditional data distribution has been pursued in the context of calculating Shapley values.
These are different interpretability method based on \emph{characteristic functions} from cooperative game theory that assign a value to every subset of a number of features~\citep{shapley201617}. 
We will shortly discuss the approach proposed in~\cite{lundberg2017unified}, where features correspond to partial vectors supported on sets.

Let $f: [0,1]^d \rightarrow \skl{-1,1}$ be a classifier function. Then we can naturally define a characteristic function $\nu_{f,\bfx}:\CP([d]) \rightarrow [-1,1]$ as
\begin{align*}
    \nu_{f,\bfx}(S) = \E_{\bfy \sim \CD}\ekl{f(\bfy)\,|\, \bfy_S = \bfx_S } = \int f(\bfy_S, \bfx_{S^c}) d\P_{\bfy\sim\CD}(\bfy_{S^c} ~|~ \bfy_S = \bfx_S).
\end{align*}
The Shapley value for the input component $x_i$ is then defined as
\[
 \text{Shapley Value}(x_i) = \frac{1}{d} \sum_{S \subseteq [d]\setminus\skl{i}} \begin{pmatrix} d-1 \\ \bkl{S} \end{pmatrix}^{-1}(\nu_{f,\bfx}(S\cup \skl{i}) - \nu_{f,\bfx}(S)).
\]
In~\cite{macdonald2021interpretable}, the characteristic function is instead used to define a feature selection method as
\[
 \bfx_{S^*}\quad\text{where}\quad S^* = \argmin_{\bkl{S}\leq k}\,\dist(\nu([d]),\nu(S)),
\]
where $k\in [d]$ is a cap on the set size and $\dist$ is an appropriate distance measure.

\begin{figure}[t]
    \centering
    \includegraphics[width=0.45\textwidth]{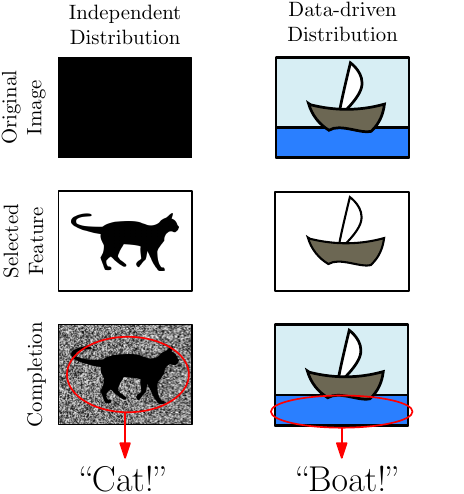}
    \caption{\label{fig:failures}Different failure modes of unrepresentative distributions. \emph{Left:} Independent, random inpainting, similar to~\cite{macdonald2019rate}. From a black image, the shape of a cat is selected, and the rest is filled with uniform noise. The shape of a cat is detected by a classifier.
    \emph{Right:} Data-driven inpainting, similar to~\cite{agarwal2020explaining}. The image of a ship is given and the ship-feature is selected. The data driven distribution inpaints the water back into the image, since in the dataset, ships are always on water. The faulty classifier that relies on the water feature is undetected, as the ship-feature indirectly leads to the correct classification.} 
\end{figure}

As in our setup, the problem is that these approaches depend on how well the conditional probability $\P_{\bfy\sim\CD}(\bfy_{S^c} ~|~ \bfy_S = \bfx_S)$ is modelled. Modelling the data distribution incorrectly makes it possible to manipulate many existing XAI-methods. This is done by changing the classifier in such a way that it gives the same value on-manifold, but arbitrary values off-manifold. To get feature-based explanations independent of the off-manifold behaviour, one needs to model the data manifold very precisely~\citep{aas2021explaining,dombrowski2019explanations}.
The authors of~\cite{anders2020fairwashing, heo2019fooling} and~\cite{dombrowski2019explanations} demonstrate this effect for existing techniques, such as sensitivity analysis, LRP, Grad-Cam, IntegratedGradients and Guided Backprop. They are able to manipulate relevance scores at will and demonstrate how this can be used to obfuscate discrimination inside a model.
LIME and SHAP can be manipulated as well~\citep{slack2020fooling} by using a classifier that behaves differently outside off-distribution if the wrong distribution for the explanation. For RDE~\citep{macdonald2019rate} it is assumed that features are independent and normally distributed, and it was demonstrated that the off-manifold optimisation can create new features that weren't in the original image~\citep{waldchen2022training}.

We now discuss two approaches proposed to model the data distribution and why each leads to a different problem by under- or over-representing correlation in the data respectively.

\paragraph{Independent distribution:}
Which means that the conditional probability is modelled as
 \[
 \P_{\bfy\sim\CD}(\bfy_{S^c} ~|~ \bfy_S = \bfx_S) = \prod_{i \in S^c} p(y_i),
 \]
 where $p(y_i)$ are suitable probability densities on the individual input components. 
 This approach has been used in~\cite{fong2017interpretable} and~\cite{macdonald2021interpretable}, where optimisers are employed to find small features that maximise the classifier score. In fact, \cite{macdonald2021interpretable} has to make a new approximation of the data distribution in every layer and it has been shown that for neural networks one cannot do much better than applying either this or sampling~\cite{macdonald2022complete}. 
 It was highlighted in~\cite{waldchen2022training} how this approach, when modelling the data distribution incorrectly, will create artificial new features that were not present in the original image. 
 Employing an optimisation method with this distribution can result in masks that generate new features that were not present in the original image. We illustrate this problem in~\Cref{fig:failures}.
 Cutting a specific shape out of a monochrome background will with high likelihood result in an image where this shape is visible.
If the distribution was true, a monochrome shape would likely lead to an inpainting that is monochrome in the same colour, destroying the artificial feature. But an independent distribution under-represents these reasonable correlations.

\paragraph{Taking a data-determined distribution via generative model:}
Which means that the conditional probability is modelled as
 \[
 \P_{\bfy\sim\CD}(\bfy_{S^c} ~|~ \bfy_S = \bfx_S) = G(\bfy_{S^c}; \bfx_S),
 \]
 where $G$ is a suitable generative model.
 Generative models as a means to approximate the data distribution in the context of explainability have been proposed in a series of publications~\citep{agarwal2020explaining,chang2018explaining,liu2019generative,mertes2020not}. This setup introduces a problem. If the network and the generator were trained on the same dataset, the biases learned by the classifier will appear might be learned by the generator as well (see~\Cref{fig:failures} for an illustration)! The important cases will be exactly the kind of cases that we will not be able to detect. If the generator has learned that horses and image source tags are highly correlated, it will inpaint an image source tag when a horse is present. This  allows the network to classify correctly, even when the network only looks for the tag and has no idea about horses. The faulty distribution over-represents correlations that are not present in the real-world data distribution.

\subsection{Design of the Three-Way Game}\label{apx:game_design}

The basic setup for a prover-verifier game for classification was proposed by \citeauthor{chang2019game} with a verifier, a cooperative prover and an adversarial prover for one specific class. The verifier either accepts the evidence for the class or rejects it. Both provers try to convince the verifier, the cooperative prover operates on data from the class, the adversary on data from outside the class.
The authors suggest that the way to scale to multiple classes is to have three agents for every class.

In our work, we combine the agents over all classes, to have a single verifier (Arthur), cooperator (Merlin), and adversary (Morgana).
The verifier rejecting all the classes in their paper corresponds to our ``Don't know!'' option. In our design in~\Cref{sec:numerics}, we make the implicit assumption that the class of the data point is unique.
Combining the verifiers gives us a numerical advantage for two reasons. First, since a lot of lower-level concepts (e.g. edges and corners for image data) are shared over classes, the lower levels of the neural network benefit by being trained on more and more diverse data. Second, we can leverage the knowledge that the class is unique by outputting a distribution over classes (and ``Don't know!''). Both lines of reasoning are standard for deep learning \cite{bridle1989training}. 

\citeauthor{anil2021learning} further combine Merlin and Morgana into a single prover that probabilistically produces a certificate for a random class. This has the advantage that it allows for further weight-sharing among the provers. However, the probabilistic nature of the certificate is also a disadvantage.
The probability of generating the certificate for the correct class is the inverse of the total number of classes.
When applied after training, one only occasionally gets a valid classification. 
In our case, we can always use Merlin together with Arthur to obtain the correct class together with an interpretable feature

\subsection{Alternative setups}
\label{apx:alternative}

Here we discuss alternatives to the Merlin-Arthur setup. Both these alternatives present an interactive classification setup as well. However, we show that they cannot prove bounds with the same generality as we have proven.

\subsubsection{Debate Model}

The debate setting introduced in~\cite{irving2018ai} is an intriguing alternative to our proposed setup. However, we are now going to present an example data space on which, in debate mode, Arthur and Merlin can cooperate perfectly without using informative features. For this, we use the fact that in the debate setting, Arthur receives features from both Merlin and Morgana for each classification. Our example illustrates that the debate setting would need stronger requirements on either the data space or Arthur to produce results similar to ours.

\begin{figure}[t]
    \centering
    \resizebox{0.5\textwidth}{!}{
    \input{img/debate_new}}
    \caption{\label{fig:debate}Schematic of $\mathfrak{D^{\text{ex}}}$ as defined in~\Cref{ex:debate}. a) The data space forms a bipartite graph, where every data point shares exactly one feature each with two data points from the opposite class. b) Classification on data point $[3,4]$. Merlin chooses the feature with the smallest 1-norm from this data point, so $[3,*]$. Arthur chooses the class of the data point with the highest 1-norm compatible with the presented features, so correctly $[3,4]$. Morgana can choose $\varnothing$, $[*,2]$ or $[3,*]$, but in all cases Arthur can correctly identify the original data point and return class $l=-1$.} 
\end{figure}
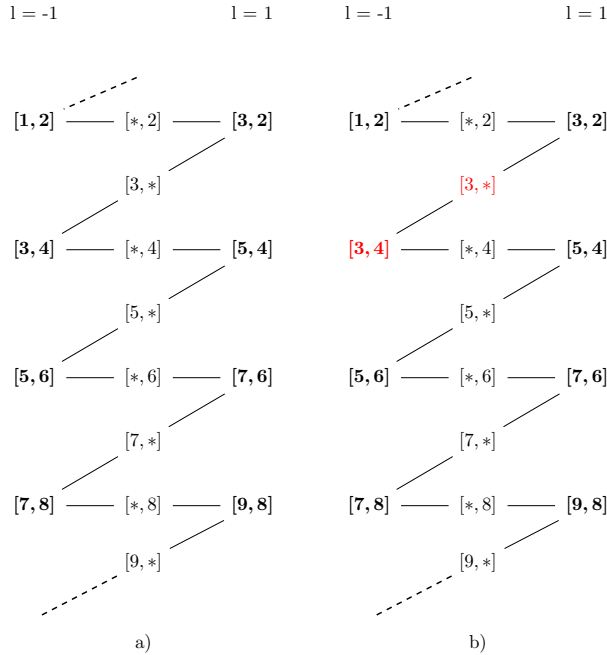

Consider the following example of a data space $\mathfrak{D^{\text{ex}}}$, illustrated in~\Cref{fig:debate}.
\begin{expl}
Given $n\in \mathbb{N}, n\geq 4$, we define the data space $\mathfrak{D^{\text{ex}}} = (D, \CD, c)$ with
\begin{itemize}
 \item $D = D_{-1} \cup D_1 \quad\text{where}\quad D_{s} = \bigcup^n_{k=1} [2k+s,2k]$
 \item for $T \in \CP(D): \CD(T) = \frac{\bkl{T}}{N}$,\vspace{-5pt}
 \item $c(\bfx) = \begin{cases} -1 & \bfx \in D_{-1} \\ 1 & \bfx \in D_{1}. \end{cases}$\vspace{-5pt}
\end{itemize}\label{ex:debate}
\end{expl}
None of the features in $D_p$ are informative of the class and the mutual information $I(c(x);\bfz\in \bfx)$ for any $\bfz\in D_p$ is zero.
Nevertheless, in a debate setting, Arthur can use the following strategy after receiving a total of two features from Merlin and Morgana
\[
 A(\skl{\bfz_1, \bfz_2}) =
 \begin{cases}
 c(\bfx^*) &\text{where}~ \bfx^* = \argmax_{\bfx\in D} \nkl{\bfx}_1~\text{s.t.}~ \bfz_1 \subseteq \bfx, \bfz_2 \subseteq \bfx, \\
 0 & \text{if} \not\exists \bfx\in D: \bfz_1 \subseteq \bfx, \bfz_2 \subseteq \bfx.
 \end{cases}
\]
This means he returns the class of the data point with the largest 1-norm that fits the presented features.
But now Merlin can use the strategy
\[
 M(\bfx) = \argmin\limits_{\bfz} \nkl{\bfz}_1  \,\text{s.t.}\, \bfz \subseteq \bfx,
\]
which returns the feature with the smaller 1-norm. It is easy to verify that no matter what Morgana puts forward, a feature with smaller or larger entry, nothing can convince Arthur of the wrong class. If she gives the same feature as Merlin, the data point will be correctly determined by Arthur. If she gives the other feature, the true data point is the unique one that has both features. Arthur's strategy works as long as \emph{someone} gives him the smaller feature.

In a setting where Arthur has to evaluate every feature individually, the best strategy that Arthur and Merlin can use achieves $\epsilon_{c} = \epsilon_{s} = \frac{1}{3}$, by making use of the asymmetric feature concentration. The AFC for $\mathfrak{D^{\text{ex}}}$ is $\kappa=2$, as can be easily verified by taking $F=\skl{[*,2], [3,*], [*,6], [7,*]}$ in the definition of the AFC, see~\Cref{def:afc}, and observing that they cover 4 data points in class $l=-1$ and only two in class $l=1$.
But since the AFC-constant appears in the bound, the lower bound for $\ap_{\mathcal{D}}(M)$ is $\frac{1}{6}$, well below the actual average precision of $\frac{1}{2}$.

This example demonstrates that Arthur and Merlin can successfully cooperate even with uninformative features, as long as Arthur does not have to classify on features by Morgana alone.
This implies that to produce similar bounds as in our setup, the debate mode needs stronger restrictions on either the allowed strategies of Arthur or the structure of the data space, such that this example is excluded.

\subsubsection{Adversarial Classifier}\label{apx:adv_classification}

\begin{figure}[b]
    \centering
    \begin{tabular}{c@{\hspace{3cm}}c}
    \includegraphics[height=3cm]{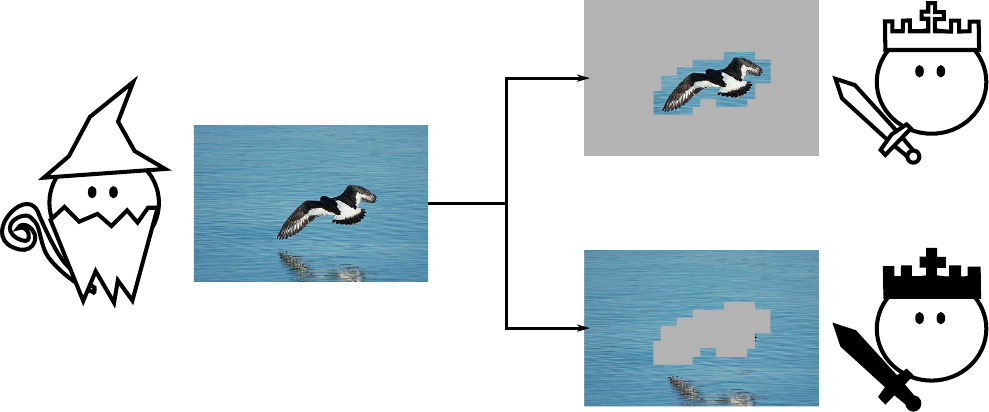}
    &
    \includegraphics[height=3cm]{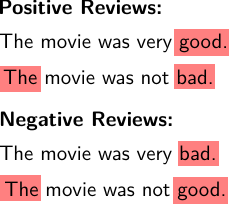}
    \\
    a) & b)
    \end{tabular}
    
    \caption{\label{fig:adv_classifier}\small a) Illustration of the adversarial classification setup. Here, the prover selects a feature from the data point and sends it the cooperative verifier that decides the class. The data point without the feature is send to an adversary that also aims to decide the class. b) An example dataset that illustrates why this setup might fail to select for informative features. All selected features appear once in each class, thus have zero mutual information. Nevertheless, it is possible to classify perfectly with the selection, but not better than chance with the leftover.}  
\end{figure}

An alternative interactive setup has been proposed in~\cite{yu2019rethinking, dabowski2017unets}, see~\Cref{fig:adv_classifier} a) for an illustration. In this setup, a single prover selects a feature from the data point and sends it to a cooperative classifier that decides the class. The rest of the data point is sent to an adversarial classifier that also tries to classify correctly. The aim of the prover is to maximise the probability that the cooperator classifies correctly, and that the adversary cannot perform much better than chance. This setup prevents cheating (in the sense illustrated in~\Cref{fig:cheating}), because selecting uninformative features would leave the informative features for the adversary.
The optimal selections thus captures all the features that are sufficient to decide the class, whereas our Merlin-Arthur setup captures just the features that are necessary to decide the class. We should expect the latter set to be contained in the former.

\paragraph{Asymmetric Feature Correlation} We can analogously define completeness and soundness values corresponding to the classification accuracy of the cooperator and adversary respectively. But our argument laid out in~\Cref{fig:afc} still holds. Merlin can use the uninformative ``fish'' and ``fruit'' features to communicate the class to the cooperator. This works for all images except the two on the left with the many features. These are also the only images where the adversary can then still decide the class. Thus, completeness and soundness can be made arbitrarily good even with uninformative features as long as the AFC can be made arbitrarily large. This implies that the AFC constant of the dataset plays a role in this setup as well.

But even when accounting for the AFC, more care is needed to state bounds. This is illustrated by a counterexample in~\Cref{fig:adv_classifier} b). Here, four reviews of a movie are classified as positive or negative. Merlin can use the highlighted sections as features that are sent to the cooperative classifier. Three things are true for this strategy:
\begin{enumerate}
 \item The mutual information between the features and the class is zero. Every feature appears once in each class. %\vspace{-0.6cm}
 \item It is possible to classify perfectly with the features since Merlin's selection is unique for each class. %\vspace{-0.2cm}
 \item It is impossible to classify the leftover text better than chance. Each leftover appears once in each class.
\end{enumerate}
The trick here is of course that Merlin uses the word ``The'' to indicate negation to his cooperator. This trick works as long as there is a XOR-like relationship in the data that flips the meaning of another feature, in this case the ``good'' and ``bad''. These XOR-like features are especially likely in text, but can exist in images as well.
Even though the mutual information of the features is zero, they still have a significant overlap with features that actually have high mutual information. So this issue might be resolved by a careful reformulation of the objective of the interactive setup. We also do not claim that this will very likely happen in real-world applications, but it must be dealt with to derive theoretical bounds.

\section{Theoretical Details}\label{apx:theory}

We now give further explanations to our theoretical investigation in the main part as well as provide definitions and proofs for the previously stated theorems and lemmas.

\subsection{Conditional entropy and Average Precision}\label{apx:entropy}
We restate the definition of the average precision and average class conditional entropy to show how one can be bound by the other.
The average precision of a feature selector $M$ is defined as
\[
\ap_{\CD}(M) := \E_{\bfy\sim \CD} \ekl{\P_{\bfx\sim\CD}\ekl{c(\bfx) = c(\bfy) \,|\, M(\bfy) \subseteq \bfx}}.
\]
The average class conditional entropy with respect to a feature selector is defined as
 \[
H_{\bfy,\bfx\sim\CD}(c(\bfx)\,|\, M(\bfy) \subseteq \bfx) := \E_{\bfy\sim \CD}\ekl{H_{\bfx \sim \CD}(c(\bfx)\,|\, M(\bfy) \subseteq \bfx)}.
\]
We can expand the latter and reorder that expression in the following way:
\begin{align*}
 H_{\bfy,\bfx\sim\CD}(c(\bfx)\,|\, M(\bfy) \subseteq \bfx)
  &= - \E_{\bfy\sim \CD} \ekl{\sum_{l\in \skl{-1,1}}P(c(\bfx)=l \,|\, M(\bfy) \subseteq \bfx) \log\kl{P_{\bfx\sim\CD}(c(\bfx)=l \,|\, M(\bfy) \subseteq \bfx)}}\\
  &=
  - \E_{\bfy\sim \CD} \ekl{\sum_{l\in \skl{c(\bfy),-c(\bfy)}}P(c(\bfx)= l \,|\, M(\bfy) \subseteq \bfx) \log\kl{P_{\bfx\sim\CD}(c(\bfx)=l \,|\, M(\bfy) \subseteq \bfx)}} \\
  &= \E_{\bfy\sim \CD} \ekl{H_b(P_{\bfx\sim\CD}(c(\bfx)=c(\bfy) \,|\, M(\bfy) \subseteq \bfx)},
\end{align*}
where $H_b(p)= -p\log(p) - (1-p)\log(1-p)$ is the binary entropy function.
Since $H_b$ is a concave function,  we can use Jensen's inequality and arrive at the bound
\[
 H_{\bfy,\bfx\sim\CD}(c(\bfx)\,|\, M(\bfy) \subseteq \bfx) \leq H_b(\ap_{\CD}(M)).
\]

We now give a short proof for~\Cref{lem:feature_prob}. 

\featureprob*

\begin{proof}
The proof follows directly from Markov's inequality, which states that for a non-negative random variable $Z$ and $\delta>0$
\[
 \P\ekl{Z \geq \delta} \leq \frac{\E\ekl{Z}}{\delta}.
 \]
 Choosing $Z= 1 - \P_{\bfx\sim\CD}\ekl{c(\bfx) = c(\bfy) \,|\, M(\bfy) \subseteq \bfx}$ with $\bfy \sim \CD$ leads to the result.
\end{proof}

\subsection{Min-Max Theorem}

We now present the proof for~\Cref{thm:minmax} which we restate here.
\minmax*

\begin{proof}
 From the definition of $\epsilon_M$ it follows that there exists a not necessarily unique $A^\sharp \in \CA$ such that
 \begin{equation}\label{eq:art_mer:minimal_morgana}
       \max_{\widehat{M} \in \CM} \,\P_{\bfy\sim \CD}\ekl{\bfy \in E_{M,\widehat{M},A^\sharp}} = \epsilon_M.
 \end{equation}
 Given $A^\sharp$, an optimal strategy by Morgana is certainly
 \[
   \widehat{M}^{\sharp}(\bfy) = \begin{cases}
   \bfz\;\text{s.t.}\; A(\bfz) = -c(\bfy) & \text{if possible},  \\
   \varnothing & \text{otherwise},
   \end{cases}
 \]
 and every optimal strategy differs only on a set of measure zero.
 Thus, we can define
 \[
  D^\prime = D\setminus E_{M,\widehat{M}^\sharp,A^\sharp},
 \]
 and have $\P_{\bfy\sim \CD}\ekl{\bfy \in D^\prime} \geq 1-\epsilon_M$.
 We know that $A(M(\bfy))\neq 0$ when $\bfy \in D^\prime$ and thus can finally assert that
 \[
 \forall \bfy,\bfx \in D^\prime:  M(\bfy)\subseteq \bfx \Rightarrow c(\bfx) = c(\bfy).
 \]
 Otherwise there would be at least one $\bfx \in D^\prime$ that would also be in $E_{M,\widehat{M}^\sharp,A^\sharp}$.
 Thus, we conclude $\ap_{\CD^\prime}(M) = 1$, and from
 \[
  0 \leq H_{\bfy,\bfx \sim \CD^\prime}(c(\bfx)\;|\; \bfx \in M(\bfy)) \leq H_b(\ap_{\CD^\prime}(M)) = 0,
 \]
 it follows that $H_{\bfy,\bfx \sim \CD^\prime}(c(\bfx)\;|\;  M(\bfy)\subseteq \bfx) = 0$.
\end{proof}

This theorem states that if Merlin uses a strategy that allows Arthur to classify almost always correctly, thus small $\epsilon_M$, then there exists a dataset that covers almost the entire original dataset and on which the class entropy conditioned on the features selected by Merlin is zero.

This statement with a new set $D^\prime$ appears convoluted at first, and we would prefer a simple bound, such as
\[
  \ap_{\CD}(M) \geq 1-\epsilon_{M},
\]
where we take the average precision over the whole dataset. This is, however, not straightforwardly possible due to a principle we call \emph{asymmetric feature correlation} and which we introduce in the next section.

\subsection{Asymmetric Feature Correlation}\label{apx:afc}

\begin{figure}
    \centering
    \begin{tabular}{cc}
         a)&\raisebox{-.5\height}{\includegraphics[width=0.8\textwidth]{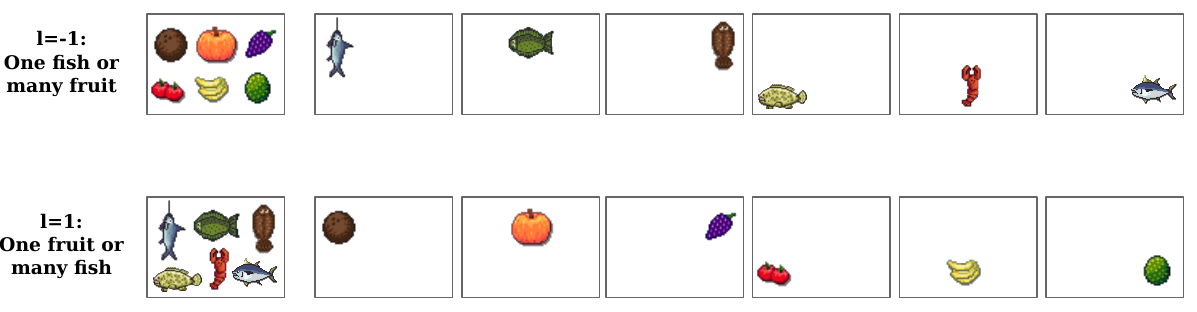}}  \vspace{0.3cm}\\[1cm]\hline\\\vspace{0.3cm}
         b) & \raisebox{-.5\height}{\includegraphics[width=0.8\textwidth]{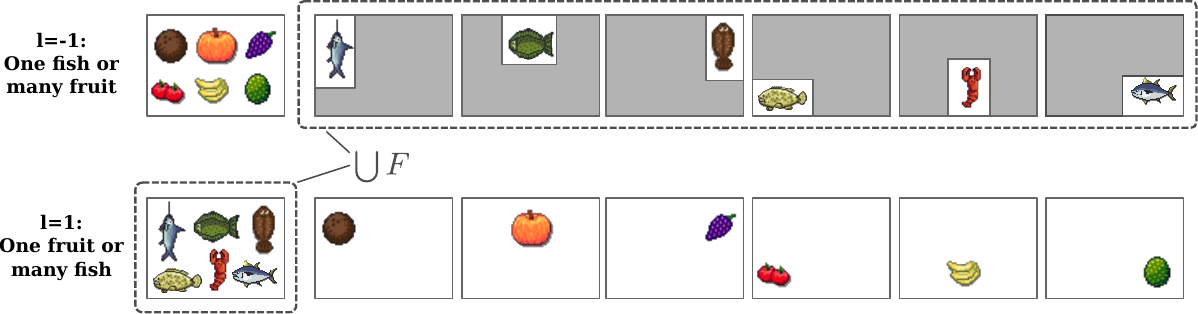}}
    \end{tabular}
    
    \caption{\label{fig:afc_supp}\small Illustration of a data space with strong asymmetric feature correlation. \textit{a)}
    A dataset with fish and fruit features. The features are asymmetrically correlated, because all fruit features are maximally correlated in class $-1$ (they are all in the same image) and maximally uncorrelated in $1$ (no two fruits share the same image). The reverse is true for the fruits. See~\Cref{fig:afc_strategy} for a strategy for Merlin that ensures strong completeness and soundness with uninformative features.\\
     \textit{b)}
     Asymmetric feature correlation for a specific feature selection. For the class $-1$, we select the set $F$ of all ``fish'' features.
     Each individual fish feature in $F$ covers a fraction of $\frac{1}{6}$ of $(F^\ast)\cap D_{-1}$ and all images (one) in $(F^\ast)\cap D_{1}$. The expected value in~\Cref{eq:am:afc_expect} thus evaluates to $k=6$. This is also the maximum AFC for the entire dataset as no different feature selection gives a higher value.}
\end{figure}

\emph{Asymmetric feature correlation (AFC)} is a concept that will allow us to state our main result.
It measures if there is a set of features that are concentrated in a few data points in one class, but spread out over almost all data points in the other class.
This represents a possible quirk of datasets that can result in a scenario where Arthur and Merlin cooperate successfully with high probability, Morgana is unable to fool Arthur except with low probability-- and yet the features exchanged by Arthur and Merlin are uninformative for the class.
An illustration of such an unusual dataset is given in~\Cref{fig:afc_supp}.

For an illustration of the asymmetric feature correlation, consider two-class data space $\mathfrak{D} = \skl{D, \CD, c}$, e.g. the ``fish and fruit'' data depicted in~\Cref{fig:afc_supp}. 
Let us choose $F \subset D_p$ to be all the ``fish'' features.
We see that these features are strongly anti-concentrated in class $l=-1$ (none of them share an image) and strongly concentrated in class $l=1$ (all of them are in the same image).

For now, let us assume $F$ is finite and let $\CF = \CU(F)$, the uniform measure over $F$. Let us recall that $F^\ast := \skl{\bfx \in D~|~ \exists~ \bfz \in F: \bfz \subseteq \bfx}$.
We have strong AFC if the class-wise ratio of what each feature covers individually is much larger than what the features cover as a whole:
\[
\E_{\bfz \sim \CF}\ekl{\frac{\P_{\bfx \sim \CD_{-l}}\ekl{\bfz \subseteq \bfx}}{\P_{\bfx \sim \CD_l}\ekl{\bfz \subseteq \bfx}}} \gg \frac{\P_{\bfx \sim \CD_{-l}}\ekl{\bfx \in F^\ast}}{\P_{\bfx \sim \CD_{l}}\ekl{\bfx \in F^\ast}}.
\]
In our example, every individual ``fish'' feature covers one image in each class, so the left side is equal to 1. As a feature set, they cover six images in class $-1$ and one in class $1$, so the right side is $\frac{1}{6}$.
Using
\[
  \frac{\P\ekl{\bfz \subseteq \bfx}}{\P\ekl{\bfx \in F^\ast}} = \P\ekl{\bfz \subseteq \bfx \,|\, \bfx \in F^\ast},
\]
we can restate this expression as
\begin{equation}\label{eq:am:afc_expect}
\E_{\bfz \sim \CF} \ekl{ \kappa_l(\bfz, F)} \gg 1,
\end{equation}
where
\[
 \kappa_l(\bfz,F) = \frac{\P_{\bfx \sim \CD_{-l}}\ekl{\bfz \subseteq \bfx \,|\, \bfx \in F^\ast}}{\P_{\bfx \sim \CD_l}\ekl{\bfz \subseteq \bfx \,|\, \bfx \in F^\ast}}.
\]
For our ``fish'' features we have $\kappa_{-1}(\bfz,F)=6$ for every feature $\bfz\in F$. Considering an infinite set $F$, we need a way to get a reasonable measure $\CF$, where we don't want to ``overemphasise'' features that appear only in very few data points. We thus define a feature selector $f_F \in \CM(F^\ast)$ as
\begin{equation}\label{eq:am:feature_selection}
   f_{F}(\bfx) = \argmax_{\substack{\bfz \in F \\ \text{s.t. } \bfz \subseteq \bfx}} \kappa(\bfz, F), 
\end{equation}
and we can define the push-forward measure $\CF = f_{F*}\CD_l|_{F^\ast}$. For our fish and fruit example, where $F$ is the set of all fish features,  $f_{F*}$ would select a fish feature for every image in class $-1$ that is in $F^\ast$.
Putting everything together, we get the following definition.

\begin{dfn}[Asymmetric feature correlation]%\label{def:afc}
Let $(D, \CD, c)$ be a two-class data space, then the asymmetric feature correlation $\kappa$ is defined as 
\[
  \kappa = \max_{l\in \skl{-1,1}} \max_{F\subset D_p}\,\E_{\bfy \sim \CD_l|_{F^\ast}} \ekl{\max_{\substack{\bfz \in F \\ \text{s.t. }\bfz \subseteq \bfy}}\frac{\P_{\bfx \sim \CD_{-l}}\ekl{\bfz \subseteq \bfx \,\middle|\, \bfx \in F^\ast}}{\P_{\bfx \sim \CD_l}\ekl{\bfz \subseteq \bfx \,\middle|\, \bfx \in F^\ast}}}.
\]
\end{dfn}

\begin{figure}[t]
    \centering
    \includegraphics[width=0.3\textwidth]{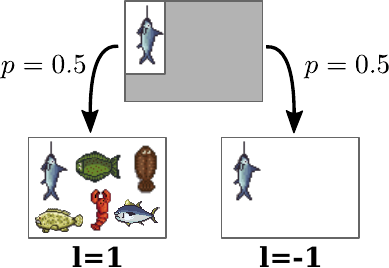}
    \caption{\label{fig:afc_strategy}\small In the dataset presented in~\Cref{fig:afc_supp}, Merlin can use the strategy to always select the fish features for class $l=-1$ and the fruit features for class $l=1$ if they exist and choose something arbitrary otherwise. Arthur can then guess $l=1$ if he gets a fish and $l=-1$ for a fruit. This strategy fails only for the images containing all fruits or fish, and can only be exploited by Morgana for those same two images out of 14.  The completeness and soundness constants in this case are both $\frac{1}{7}$.
    But as illustrated here, each ``fish'' feature is completely uninformative of the class. Conditioned on the selected fish, it could either be the image from class $l=-1$ or from $l=1$.
    }   
\end{figure}

As we have seen in the ``fish and fruit'' example, the AFC can be made arbitrarily large, as long as one can fit many individual features into a single image. We can prove that the maximum amount of features per data point indeed also gives an upper bound on the AFC.
We now come back to~\Cref{lem:afc_bound} and prove it.
\afcbound*

\begin{proof}
 Let $l\in\skl{-1,1}$ and let $F \subset D_p$. We define $f_F \in \CM(F^\ast)$ as in~\Cref{eq:am:feature_selection} as well as $\CF= f_{F*}\CD_l|_{F^\ast}$. We can assert that
\begin{equation}\label{eq:am:maxfeatures}
\P_{\bfx \sim \CD_l} \ekl{\bfz \subseteq \bfx \,|\, \bfx\in F^\ast} \geq \P_{\bfx \sim \CD_l} \ekl{f(\bfx) = \bfz \,|\, \bfx\in F^\ast} = \CF(\bfz).
\end{equation}
 We then switch the order of taking the expectation value in the definition of the AFC:
  \begin{align*}
  \kappa_l(F)
  &=
  \E_{\bfz \sim \CF}\ekl{\frac{\P_{\bfx \sim \CD_{-l}}\ekl{\bfx \in F \,|\, \bfx \in F^\ast}}{\P_{\bfy \sim \CD_l}\ekl{\bfz \subseteq \bfy \,|\, \bfy \in F^\ast}}} \\
  &=
  \E_{\bfz \sim \CF}\ekl{\frac{\E_{\bfx \sim \CD_{-l}}\ekl{\mathbf{1}\kl{\bfz \subseteq \bfx }\,|\, \bfx \in F^\ast}}{\P_{\bfy \sim \CD_l}\ekl{\bfz \subseteq \bfy \,|\, \bfy \in F^\ast}}} \\
  &=
 \E_{\bfx \sim \CD_{-l}}\ekl{\E_{\bfz \sim \CF}\ekl{\frac{\mathbf{1}\kl{\bfz \subseteq \bfx} }{\P_{\bfy \sim \CD_l}\ekl{\bfz \subseteq \bfy \,|\, \bfy \in F^\ast}}}\,\middle|\, \bfx \in F^\ast}.
  \end{align*} 
  Since there are only finitely many features in a data point we can express the expectation value over a countable sum weighted by the probability of each feature:
 \begin{align*}
  \kappa_l(F)
  &=
 \E_{\bfx \sim \CD_{-l}}\ekl{\sum_{\bfz\in F:\, \bfx\in F}\ekl{\frac{ \CF(\bfz) }{\P_{\bfy \sim \CD_l}\ekl{\bfz \subseteq \bfy \,|\, \bfy \in F^\ast}}}\,\middle|\, \bfx \in F^\ast} \\
  &\leq 
 \E_{\bfx \sim \CD_{-l}}\ekl{\sum_{\bfz\in F:\, \bfx\in F}1\,\middle|\, \bfx \in F^\ast} \\
 &\leq
  \E_{\bfx \sim \CD_{-l}}\ekl{K\,\middle|\, \bfx \in F^\ast} \\
  &= K,
  \end{align*} 
  where in the first step we used~\Cref{eq:am:maxfeatures} and the definition of $K$ in the second.
  Then we see that 
  \[
  \kappa = \max_{l\in\skl{-1,1}}\max_{F\subset D_p}~ \kappa_l(F) \leq K.
  \]
\end{proof}
\begin{figure}[t]
    \centering
    \includegraphics[width=0.8\textwidth]{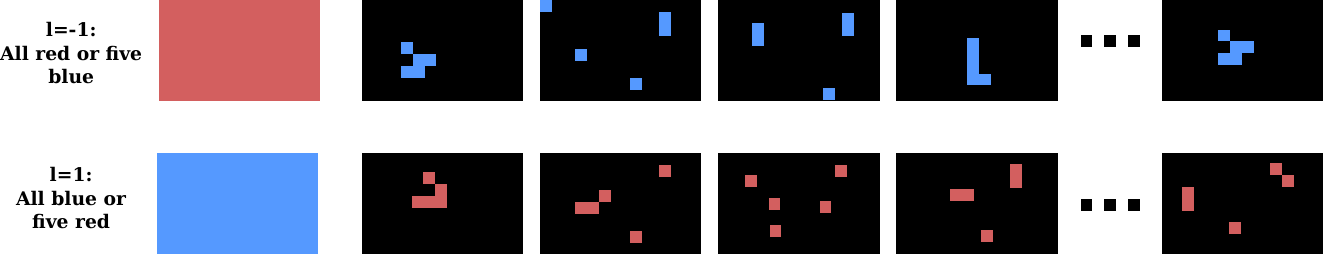}
    \caption{\label{fig:red_vs_blue}\small An example of a dataset with very high asymmetric feature correlation. The completely red image shares a feature with each of the $m$-red-pixel images (here $m=5$), of which there are $\binom{d}{m}$ many. In the worst case $m=\frac{d}{2}$, resulting in $k=\binom{d}{d/2}$ thus exponential growth in $d$.}  
\end{figure}

The number of features per data point is dependent on which kinds of features we consider. Without limitations, this number can be $2^d$, i.e., exponentially high. See~\Cref{fig:red_vs_blue} for an example of exponentially large AFC parameters. If we consider only features of a fixed size and shape, such as in image data, the number of features per data point drops to $\approx d$.

We now prove an intermediate lemma that will later allow us to prove~\Cref{thm:main}.

\begin{lem}\label{thm:am:main}
 Let $(D, \CD, c)$ be a two-class data space with asymmetric feature correlation of $\kappa$ and class imbalance $B$. Let $A: [0,1]^d \rightarrow \skl{-1,0,1}$ be a feature classifier and $M \in \mathcal{M}(D)$ a feature selector for $D$. From
\begin{enumerate}
    \item Completeness:\\
    \[
    \min_{l \in \skl{-1,1}}\P_{\bfx \sim \CD_l}\ekl{A\kl{M(\bfx)} = l } \geq 1- \epsilon_c,
    \]
    \item Soundness:\\
    \[
    \max\limits_{\widehat{M}\in \CM(D)} \max\limits_{l \in \skl{-1,1}} \P_{\bfx \sim \CD_l}\ekl{A\kl{\widehat M(\bfx)} = -l } \leq  \epsilon_s,
    \]
\end{enumerate}
follows
% \[
% \]
\[
 \ap_{\CD}(M) \geq 1 - \epsilon_c - \frac{\kappa \epsilon_s}{1 - \epsilon_c+\kappa\epsilon_sB^{-1}}.
\]
\end{lem}
This lemma gives a bound on the probability that data points with features selected by Merlin will also belong to the same class. This probability is large as long as we have a bound on the AFC of the dataset.

\begin{proof}
    
The proof of our lemma is fairly intuitive, although the notation can appear cumbersome. Here we give a quick overview over the proof steps.
\begin{enumerate}
    \item
    In the definition of the AFC, we maximise over all possible features sets.
    We will choose as a special case (for each class $l\in\skl{-1,1}$) the features that Merlin selects for data points that Arthur classifies correctly.
    \item
    These feature sets cover the origin class at least up to $1-\epsilon_c$, and the other class at most up to $\epsilon_s$, which is required by the completeness and soundness criteria, respectively.
    This gives us a high precision for the whole feature set.
    \item
    The AFC upper bounds the quotient of the precision of the whole feature set and expected precision of the individual features, which finally allows us to state our result.
\end{enumerate}

Let us define a partition of $D$ according to the true class and the class assigned by Arthur. From now on, let $l\in\skl{-1,1}$ and $m\in\skl{-1,0,1}$. We introduce the datasets
\[
 D_{l,m} = \skl{\bfx \in D_l \,|\, A(M(\bfx)) = m},
\]
which means that $D_{l,l}$ are the data points that Arthur classifies correctly, and furthermore
\[
  F_l = M(D_{l,l}).
\]
To use the AFC bound we need a feature selector $f: D_l|_{F^\ast} \rightarrow F$. Merlin itself maps to features outside $F$ when applied to data points in $D_l|_{F^\ast_l}\setminus D_{l,l}$.
Let us thus define $\sigma: D_F\setminus D_{l,l} \rightarrow F$ which selects an arbitrary feature from $F$ (in case one is concerned whether such a representative can always be chosen, consider a discretised version of the data space which allows for an ordering).
Then we can define
\[
 f(\bfx) = \begin{cases}
 M(\bfx) &\bfx \in D_{l,l}, \\
 \sigma(\bfx) & \bfx\in D_l|_{F^\ast_l}\setminus D_{l,l},
 \end{cases}
 \quad\text{and}\quad \CF_l = f_\ast D_l|_{F^\ast_l}.
\]
This feature selector will allow us to use the AFC bound.
We now abbreviate 
\[
p(\bfx, f) = \P_{\bfy\sim\CD}\ekl{c(\bfy) \neq c(\bfx) \,|\, \bfy\in f(\bfx)}  \quad\text{and}\quad P_l=\P_{\bfx \sim \CD}\ekl{\bfx \in D_{l}}.
\]
Then
\begin{equation}\label{eq:main:class_expansion}
    1- \ap_{\CD}(M) = \E_{\bfx\sim \CD} \ekl{p(\bfx, M)} =\sum_{l\in\skl{-1,1}} \E_{\bfx\sim \CD_{l}} \ekl{p(\bfx, M)}P_l.
\end{equation}
Using the completeness criterion and the fact that $p(\bfx, M)\leq 1$ we can bound
\begin{align*}
   \E_{\bfx\sim \CD_{l}} \ekl{p(\bfx,M)}
   &=
   \E_{\bfx\sim \CD_{l}} \ekl{p(\bfx,M)\bbone\kl{\bfx \in D_{l,l}}} + \E_{\bfx\sim \CD_{l}} \ekl{p(\bfx,M)\bbone\kl{\bfx \notin D_{l,l}}} \\
   &\leq
   \E_{\bfx\sim \CD_{l}} \ekl{p(\bfx,M)\bbone\kl{\bfx \in D_{l,l}}} + \epsilon_c \\
   &\leq
   \E_{\bfx\sim \CD_{l}} \ekl{p(\bfx,M)\bbone\kl{\bfx \in D_{l,l}}} + \epsilon_c + \E_{\bfx\sim \CD_{l}} \ekl{p(\bfx, \sigma)\bbone\kl{\bfx \in D_l|_{F^\ast_l}\setminus D_{l,l}}} \\
   &\leq
   \frac{\E_{\bfx\sim \CD_{l}} \ekl{p(\bfx,M)\bbone\kl{\bfx \in D_{l,l}} + p(\bfx, \sigma)\bbone\kl{\bfx \in D_l|_{F^\ast_l}\setminus D_{l,l}}}}{\P_{\bfx \sim \CD_l}\ekl{\bfx \in D_l|_{F^\ast_l}}} + \epsilon_c \\
    &=
   \E_{\bfx\sim \CD_l|_{F^\ast_l}} \ekl{p(\bfx, f)} +\epsilon_c \\
   &=\E_{\bfz \sim \CF_l} \ekl{\P_{\bfy\sim\CD}\ekl{c(\bfy)=-l \,|\, \bfz \subseteq \bfy}} +\epsilon_c.
\end{align*}
We can expand the expression in the expectation using Bayes' Theorem:
\begin{align*}
   \P_{\bfy\sim\CD}\ekl{c(\bfy) = -l \,|\, \bfy\in\bfz}
   &=\frac{\P_{\bfy\sim\CD_{-l}}\ekl{\bfz \subseteq \bfy}\P_{-l}}{\P_{\bfy\sim\CD_{-l}}\ekl{\bfz \subseteq \bfy}P_{-l} + \P_{\bfy\sim\CD_l}\ekl{\bfz \subseteq \bfy}P_{l}} \\
   &= h\kl{\frac{\P_{\bfy\sim\CD_{-l}}\ekl{\bfz \subseteq \bfy}\P_{-l}}{\P_{\bfy\sim\CD_l}\ekl{\bfz \subseteq \bfy}P_{l}}},
\end{align*}
where $h(t) = (1+t^{-1})^{-1}$. Since $h(t)$ is a concave function for $t\geq0$, we know that for any random variable $R$ have $\E\ekl{h(R)} \leq h(\E\ekl{R})$, so
\begin{equation}\label{eq:main:concave}
     \E_{\bfx\sim \CD_{l}} \ekl{p(\bfx,M)} \leq h\kl{\E_{\bfz \sim \CF_l}\ekl{\frac{\P_{\bfy\sim\CD_{-l}}\ekl{\bfz \subseteq \bfy}}{\P_{\bfy\sim\CD_l}\ekl{\bfz \subseteq \bfy}}}\frac{P_{-l}}{P_{l}}} + \epsilon_c.
\end{equation}
From the definition of the AFC $\kappa$ we know that
\begin{equation}\label{eq:am:main:afc}
\E_{\bfz \sim \CF_l}\ekl{\frac{\P_{\bfy\sim\CD_{-l}}\ekl{\bfz \subseteq \bfy}}{\P_{\bfy\sim\CD_l}\ekl{\bfz \subseteq \bfy}}} \leq \E_{\bfx\sim \CD_l|_{F^\ast_l}}\ekl{\max_{\substack{\bfz \in F \\ \text{s.t. }\bfz \subseteq \bfx}} \frac{\P_{\bfy\sim\CD_{-l}}\ekl{\bfz \subseteq \bfy}}{\P_{\bfy\sim\CD_l}\ekl{\bfz \subseteq \bfy}}} \leq \kappa \frac{\P_{\bfy\sim\CD_{-l}}\ekl{\bfy\in F^\ast}}{\P_{\bfy\sim\CD_l}\ekl{\bfy\in F^\ast}}.
\end{equation}
Now we make use of the fact that we can lower bound $\P_{\bfy\sim\CD_l}\ekl{\bfy\in F}$ by the completeness property
\[
 \P_{\bfy\sim\CD_l}\ekl{\bfy\in F^\ast} \geq 1 - \epsilon_c,
\]
and upper bound $\P_{\bfy\sim\CD_{-l}}\ekl{\bfy\in F^\ast}$ with the soundness property
\[
 \P_{\bfy\sim\CD_{-l}}\ekl{\bfy\in F^\ast} \leq \epsilon_s.
\]
This is because $\bfy\in F^\ast$ implies that there are features Morgana can use to convince Arthur of class $l$ whereas $\bfy\sim\CD_{-l}$.
Together with~\Cref{eq:main:concave} and~\Cref{eq:am:main:afc} we arrive at
\[
 \E_{\bfx\sim \CD_{l}}\ekl{p(\bfx,M)} \leq h\kl{\kappa \frac{\epsilon_s}{1-\epsilon_c}\frac{P_{-l}}{P_{l}}}+ \epsilon_c = \frac{\kappa\epsilon_s \frac{P_{-l}}{P_{l}}}{1 - \epsilon_c + \kappa\epsilon_s\frac{P_{-l}}{P_{l}}} + \epsilon_c.
\]
Using $\frac{P_{l}}{P_{-l}} \leq B$ thus $\frac{P_{-l}}{P_{l}} \geq B^{-1}$, we can express
\[
 \E_{\bfx\sim \CD_{l}}\ekl{p(\bfx,M)} \leq \frac{\kappa\epsilon_s \frac{P_{-l}}{P_{l}}}{1 - \epsilon_c + \kappa\epsilon_s B^{-1}}+ \epsilon_c.
\]
Inserted back into equation~\Cref{eq:main:class_expansion} leads us to
\[
  1 - \ap_{\CD}(M) \leq \frac{\kappa\epsilon_s}{1 - \epsilon_c + \kappa\epsilon_s B^{-1}} + \epsilon_c.
\]
\end{proof}

Bounding the AFC is a hard task that is not possible for most real-world dataset. This is why we evaluate~\Cref{ass:realistic} on the UCI and MNIST dataset. Here, we want to give further intuition why we think that we are unlikely to encounter effects of high AFC when training Merlin-Arthur classifiers, even if the dataset theoretically has a high AFC:
\begin{enumerate}
    \item A learning barrier: Merlin and Arthur's strategy cannot be arbitrary, because it has to be learned on the training dataset and generalise to the test dataset. This precludes any set of features (for example an intricate set of pixels) that is over-optimised on the training data. 
    \item A computational barrier: We conjecture that for datasets where the AFC is computationally  hard to estimate, it will also be hard for Merlin and Arthur to exploit. There are evidence provided by ~\citeauthor{waldchen2023hardness}, which looks at the hardness of deceptive certificate selection.
\end{enumerate}

\subsection{Relative Success Rate and Realistic Algorithms} \label{sec:rel_success}

As discussed in~\Cref{ssc:realistic} realistic algorithms will not be optimal players as in~\Cref{thm:minmax}. It will turn out that we can relax the requirement for Morgana to play optimally in two important ways: (i) She is only required to find features that can also be found by Merlin (2) She only needs success on a similar rate as Merlin.
Thus, what's crucial is the relative strength between the algorithms utilized for Merlin and Morgana.
We can define \emph{relative success rate} in the following way:
\begin{dfn}[Relative Success Rate]
  Let $\mathfrak{D}=(D, \CD, c)$ be a two-class data space. Let $A\in \CA$ and $M, \morg \in \CM(D)$
 Then the relative success rate $\alpha$ of $\morg$ with respect to $A,M$ and $\mathfrak{D}$ is defined as
 \[
     \alpha := \min_{l\in \skl{-1,1}} \frac{\P_{\bfx\sim \CD_{-l}}\ekl{A(\morg(\bfx))=l \,|\, \bfx \in F_l^\ast}}{\P_{\bfx\sim \CD_{l}}\ekl{A(M(\bfx))=l \,|\, \bfx \in F_l^\ast}},
 \quad\text{where}\quad
 F_{l}^\ast := \skl{\bfx \in D \,|\, \exists \bfz\subseteq \bfx:~ \bfz\in M(D_l), A(\bfz)=l}
 \]
\end{dfn}
The set $F_{l}$ contains all features that Merlin selects in class $l$ that successfully convince Arthur of class $l$, and $F_{l}^\ast$ is the set of all data points containing such a feature. We condition on the fact that a feature that Merlin uses is in the data point and then compare the rates of Morgana and Merlin convincing Arthur. Morgana can of course also select different features but is not required to find features that also Merlin could not find.

\begin{figure}[t]
      \centering
      \includegraphics[width=0.6\textwidth]{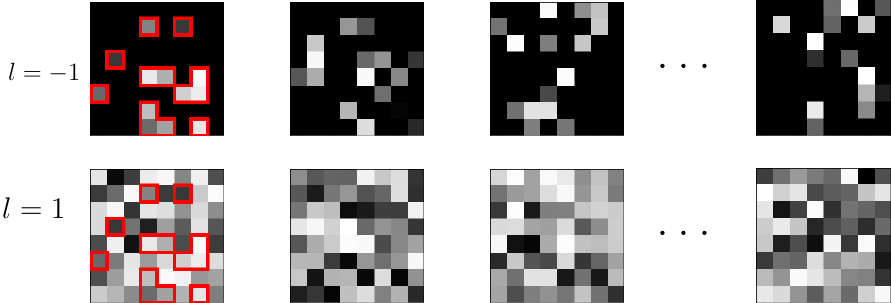}
      \caption{\label{fig:relative_strength}\small{Illustration of a dataset where the complexity of finding the relevant features depends strongly on the the irrelevant features. Class $-1$ consists of $k$-sparse images whose pixel values sum to some number $s$. For each of these images, there is a non-sparse image in class $1$ that shares all non-zero values (marked in red for the first image). Merlin can use the strategy to show all $k$ non-zero pixels for an image from class $-1$ and $k+1$ arbitrary non-zero pixels for class $1$. Arthur checks if the sum is equal to $s$ or if the number of pixels equal to $k+1$, otherwise he says ``I don't know!''. He will then classify with 100\% accuracy. Nevertheless, the features Merlin uses for class $-1$ are completely uncorrelated with the class label. To exploit this, however, Morgana would have to solve the $\SNP$-hard~\citep{kleinberg2006algorithm} subset sum problem to find the pixels for images in class 1 that sum to $s$. The question is not in which class we can find the features, but in which class we can find the features \emph{efficiently}.}}
  \end{figure}

Given that one of Merlins features is in the data point, the question is thus how much the context given by the other features affect the hardness of finding of the former.
We can easily construct scenarios in which the context matters very strongly, see~\Cref{fig:relative_strength} for an example. We expect that for most real-world data this dependence is only weak and can be upper bounded.

\begin{minipage}{0.52\textwidth}
\begin{expl}
 Let $\mathfrak{D}$ be a data space with maximum number of features per data point $K$. Let Morgana operate the algorithm described in~\Cref{alg:random}, in which she randomly searches for a convincing feature.
Then we have \[
 \alpha \geq \frac{K}{n_{\text{try}}},
\]
which corresponds to an upper bound on the probability that Morgana will succeed on an individual data point when there is only one convincing feature.
\end{expl}
  \end{minipage}\hfill
 \begin{minipage}{0.45\textwidth}
 \vspace{-1em}
\begin{algorithm}[H]
    \centering
    \begin{algorithmic}[1]
        \STATE \textbf{Input:} $\bfx\in D$, $n_{\text{try}}$
        \STATE \textbf{Output:} $\bfz \in D_p$
    \FOR{$i \in [n_{\text{try}}]$}
        \STATE Pick random feature $\bfz$ s.t. $\bfz \in \bfx$ 
        \IF{$A(\bfz)=-c(\bfx)$} 
        \STATE \textbf{return} $\bfz$
    \ENDIF
    \ENDFOR
    \STATE \textbf{return} $\varnothing$
    \end{algorithmic}
    \caption{Merlin-Arthur Training}
    \label{alg:random}
\end{algorithm}
 \end{minipage}

We generally want Morgana's algorithm to be at least as powerful as Merlin's so in case of an optimiser one can consider giving more iterations or more initial starting values.

We now want to prove~\Cref{thm:main} which we restate here.

\main*

\begin{proof}
 We follow the same proof steps and definitions as in the proof of~\Cref{thm:am:main} up to~\Cref{eq:am:main:afc}. Then we consider the following
 \begin{align}
    \alpha \frac{\P_{\bfy\sim\CD_{-l}}\ekl{\bfy\in F}}{\P_{\bfy\sim\CD_l}\ekl{\bfy\in F}} 
    &\leq
    \frac{\P_{\bfy\sim\CD_{-l}}\ekl{\bfy\in F}}{\P_{\bfy\sim\CD_l}\ekl{\bfy\in F}} \frac{\P_{\bfx\sim \CD_{-l}}\ekl{A(\morg(\bfx))=l \,|\, \bfx \in F_l^\ast}}{\P_{\bfx\sim \CD_{l}}\ekl{A(M(\bfx))=l \,|\, \bfx \in F_l^\ast}} \\
    &=
    \frac{\P_{\bfx\sim \CD_{-l}}\ekl{A(\morg(\bfx))=l,\, \bfx \in F_l^\ast}}{\P_{\bfx\sim \CD_{l}}\ekl{A(M(\bfx))=l,\, \bfx \in F_l^\ast}} \\
    &\leq
    \frac{\P_{\bfx\sim \CD_{-l}}\ekl{A(\morg(\bfx))=l}}{\P_{\bfx\sim \CD_{l}}\ekl{A(M(\bfx))=l}}
 \end{align}
 where in the last step we used that $\bfx\in D_l \und A(M(\bfx))=l\,\Rightarrow\, \bfx \in F_l^\ast$ by definition of $F_l^\ast$.
 We know by the soundness and completeness criteria that
 \[
  \P_{\bfx\sim \CD_{-l}}\ekl{A(M(\bfx,l)) = l} \leq
  \epsilon_s \quad\text{and}\quad \P_{\bfx\sim \CD_{l}}\ekl{M(\bfx,l)\in F} \geq 1 - \epsilon_c
 \]
 Putting everything together we arrive at
 \[
  \frac{\P_{\bfy\sim\CD_{-l}}\ekl{\bfy\in F}}{\P_{\bfy\sim\CD_l}\ekl{\bfy\in F}} \leq \frac{\alpha^{-1} \epsilon_s}{1-\epsilon_c},
 \]
 which allows us to continue the proof analogously to~\Cref{thm:am:main}.
\end{proof}

\subsection{Finite and Biased datasets} \label{apx:finit_and_biased_datasets}

Real datasets come with further challenges when evaluating the completeness and soundness.

Let us introduce two data distributions $\CT$ and $\CD$ on the same dataset $D$, where $\CT$ is considered the ``true'' distribution and $\CD$ a different, potentially biased, distribution.
We define this bias with respect to a specific feature $\bfz \in D_p$ as
\[
  d^{l}_{\bfz}(\CD,\CT) := \bkl{\P_{\bfx \in \CD}\ekl{c(\bfx)=l\,|\, \bfz \in \bfx} - \P_{\bfx \in \CT}\ekl{c(\bfx)=l\,|\, \bfz \in \bfx}}.
\]
Note that $d^{1}_{\bfz}(\CD,\CT) = d^{-1}_{\bfz}(\CD,\CT) =\colon d_{\bfz}(\CD,\CT)$ and $0\leq d_{\bfz}(\CD,\CT) \leq 1$.
This distance measures if data points containing $\bfz$ are distributed differently to the two classes for the two distributions.

For example, consider as $\bfz$ the water in the boat images of the PASCAL VOC dataset~\cite{lapuschkin2019unmasking}. The feature is a strong predictor for the ``boat'' class in the test data distribution $\CD$ but should not be indicative for the real world distribution which includes images of water without boats and vice versa. 
We now want to prove that a feature selected by $M$ is either an informative feature or is misrepresented in the test dataset.

\begin{lem}\label{lemma:bias}
Let $\mathfrak{D}, k, B, A, M, \alpha, \epsilon_c$ and $ \epsilon_s$ be defined as in~\Cref{thm:main}. Let $\CT$ be the true data distribution on $D$.
Then for $\delta\in[0,1]$
 we have
 \[
 \P_{\bfy \sim \CT}\ekl{c(\bfy)=c(\bfx)\,|\,M(\bfx) \subseteq \bfy} \geq 1 - \delta - d_{M(\bfx)}(\CD,\CT),
 \] 
 with probability
$
  1 - \frac{1}{\delta}\kl{\frac{k\alpha \epsilon_s}{1+k\alpha\epsilon_sB^{-1} - \epsilon_c} + \epsilon_c}
$ for $\bfx \sim \CD$.

\end{lem}
This follows directly from~\Cref{lem:feature_prob},~\Cref{thm:main}, the definition of $d_{\bfz}(\CD,\CT)$ and the triangle inequality. This means that if an unintuitive feature was selected in the test dataset, we can pinpoint to where the dataset might be biased. 

We provided~\Cref{lemma:bias} in the context of biased datasets.
The next iteration considers the fact that we only sample a finite amount of data from the possibly biased test data distribution. This will only give us an approximate idea of the soundness and completeness constants.

\begin{lem}\label{lem:am:finite}
Let $D, \sigma, \CD, c A, M$ and $\CT$ be defined as in~\Cref{lemma:bias}.
Let $D^\text{test} = \kl{\bfx_i}_{i=1}^N \sim \CD^N$ be $N$ random samples from $\CD$. Let
\[
    \epsilon^{\text{test}}_c = \max_{l\in \skl{-1,1}} \frac{1}{N} \sum_{\bfx \in D^\text{test}_l} \bbone(A\kl{M(\bfx,c(\bfx))} \neq c(\bfx)),
\]
and
\[
    \epsilon^{\text{test}}_s =\max_{l\in \skl{-1,1}} \frac{1}{N} \sum_{\bfx \in D^\text{test}_l} \bbone(A\kl{M(\bfx,-c(\bfx))} = -c(\bfx)).
\]
Then it holds with probability $1-\eta$ where $\eta\in[0,1]$ that on the true data distribution $\CT$ $A$ and $M$ obey completeness and soundness criteria with
\begin{align*}
    \epsilon_c &\leq \epsilon^{\text{test}}_c + \epsilon_{\text{dist}} + \epsilon_{\text{sample}} \quad \text{and} \\
    \epsilon_s &\leq \epsilon^{\text{test}}_s + \epsilon_{\text{dist}} + \epsilon_{\text{sample}}
\end{align*}
respectively, where $\epsilon_{\text{dist}} = \max\limits_{l\in \skl{-1,1}} \nkl{\CD_l - \CT_l}_{TV}$ and $\epsilon_{\text{sample}} = \sqrt{\frac{1}{2N} \log\kl{\frac{4}{\eta}}}$.
\end{lem}

The proof follows trivially from Hoeffding's inequality and the definition of the total variation norm.
\begin{proof}
We define $ E_{c,l} = \skl{\bfx \in D_l\,|\, A\kl{M(\bfx,c(\bfx))} \neq c(\bfx)}$ for $l\in\skl{-1,1}$ and let $E^{\CD}_{c,l}$ be the Bernoulli random variable for the event that $X \in E_{c,l}$ where $X\sim \CD$.
  Then
  \[
  \P_{\bfx \sim \CD_l}\ekl{A\kl{M(\bfx,c(\bfx))} \neq c(\bfx) } = \E\ekl{E^{\CD}_{c,l}} \, .
  \]
 Using Hoeffding's inequality, we can bound for any $t>0$
 \[
 \P\ekl{\bkl{\kl{\frac{1}{N} \sum_{\bfx \in D^\text{test}_l} \bbone(\bfx\in E_{c,l})} - \E\ekl{E^{\CD}_{c,l}}} > t} \leq e^{-2nt^2}.
 \]
  We choose $t$ such that $e^{-2t^2} = \frac{\eta}{4}$. We use a union bound for the maximum over $l\in\skl{-1,1}$ which results in a probability of  $2\frac{\eta}{4}=\frac{\eta}{2}$ we have
 \[
  \max_{l\in\skl{-1,1}}\E\ekl{E^{\CD}_{c,l}} > \epsilon^{\text{test}}_c + \sqrt{\frac{1}{2N}\log\kl{\frac{4}{\eta}}},
 \]
 and thus with $1-\frac{\eta}{2}$ we have $\max_{l\in\skl{-1,1}}\E\ekl{E^{\CD}_{c,l}} \leq \epsilon^{\text{test}}_c + \epsilon^{\text{sample}}$.
 Using the definition of the total variation norm
 \[
  \nkl{\CT-\CD}_{TV} = \sup_{J \subset D} \bkl{\CT(J)-\CD(J)},
 \]
 with $J = E_{c,l}$ we can derive $\E\ekl{E^{\CT}_{c,l}} \leq \E\ekl{E^{\CT}_{c,l}} + \nkl{\CT-\CD}_{TV}$ and thus
 \[
 \epsilon_c \leq \epsilon_c^{\text{test}} + \epsilon^{\text{sample}} + \epsilon^{\text{dist}}.
 \]
 We can treat $\epsilon_s$ analogously and take a union bound over both the completeness and soundness bounds holding which results in the probability of $1-\eta$.
\end{proof}

\section{Numerical Experiments}\label{apx:numerics}

For the numerical experiments, we implement Arthur, Merlin and Morgana in Python 3.8 using PyTorch (BSD-licensed). We perform our experiments on the UCI Census Income dataset and the MNIST dataset, which is licensed under the Creative \emph{Commons Attribution-Share Alike 3.0} licence. All experiments were executed efficiently, with each run completing in less than 15 minutes.

\subsection{UC Irvine Census Income Dataset}
\label{apx:uci_appendix}

In the following, we provide the technical details for the experiments performed on the UCI Census Income dataset, including preprocessing steps and training configurations. We also describe additional experiments that complements our analysis in~\Cref{sec:numerics}.

\subsubsection{Training Setup}

Please note that, when predicting income, a class map $c: D \rightarrow \skl{-1,1}$ cannot be defined for all data points, i.e., there are different incomes for exactly the same input. However, only small fraction of data points (0.6\%) have this issue.

\paragraph{Data Preprocessing}
The UCI Census Income dataset consists of both continuous and categorical features, 14 features in total.
The target class is chosen to be the feature ``sex'', which contains the categories ``male'' and ``female'', to indicate a possible case of discrimination. The feature ``fnlwgt'' is removed from the set of features, since it does not contain any meaningful information. In addition, the features ``marital-status'' and ``relationship'' are also removed as they strongly indicate the target class. After removal, 11 features remain for each data point. The continuous features are scaled according to the min-max scaling method. To simplify feature masking, all features are padded to a vector of the same fixed dimension. Continuous features are then repeated along the padding dimension, while categorical features are one-hot encoded to the length of the fixed dimension. Note, that the fixed dimension is determined by the categorical feature with the most categories. 

Finally, the train and test datasets are balanced with respect to the target class, resulting in 19564 train samples and 9826 test samples.

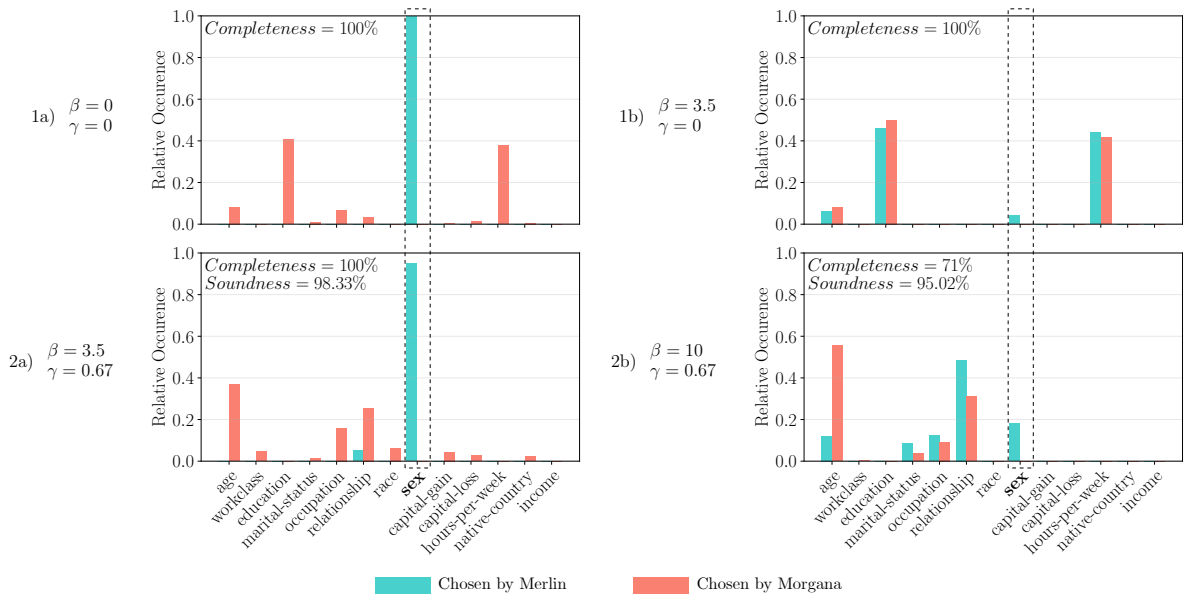
\begin{figure}[t]
    \centering
     \resizebox{0.95\textwidth}{!}{

\input{img/census_data/uci_corr_features}

        }
        % \vspace{-0.5cm}
    \caption{\label{fig:gender_corr_feat}\small{Results from the experiments in \Cref{sec:cheating}, with the features ``marital-status'' and ``relationship'' are included. 1) The selected features are the same as in the setup, excluding the two features.
    2a) Again, the requirement of soundness results in Merlin selecting  ``sex'' regardless of the punishment. 2b) Despite the higher punishment ($\beta=10$), Arthur and Merlin achieve completeness of 71\%, unlike the results presented in \Cref{sec:cheating}, due to the selection of ``relationship'', a feature strongly correlated with ``sex''.
    }}
\end{figure}

\paragraph{Model Description}
Arthur is modelled as a NN with a single hidden linear layer of size 50 followed by a ReLU activation function. The output is converted to a probability distribution via the softmax function with three output dimensions, where the third dimension corresponds to the ``Don't know!'' option. The resulting NN contains approximately 23k parameters. Merlin and Morgana, on the other hand, are modelled as Frank-Wolfe optimisers, which are discussed in more detail in the overview of the experiments conducted on the MNIST dataset.

\begin{table}[h]
    \centering
    \vspace{2mm}
    \resizebox{\columnwidth}{!}{
    \begin{tabular}{ccccc}
         \hline 
         \textbf{Model} &
         \textbf{Learning Rate} & \textbf{Batch Size} & \textbf{Frank-Wolfe Learning Rate}
         & \textbf{Epochs} \\ \hline
         Pre-training Arthur & $10^{-4}$ & 512 & - & $100$  \\
         Merlin-Arthur Classifier & $0.5$ & 512 & 0.1 & $20$ \\ \hline
    \end{tabular}}
    \caption{\label{tab:census-param}Configurations for pre-training Arthur and training the Merlin-Arthur classifiers on the UCI Census Income dataset.}
\end{table}

\paragraph{Model Selection}
The training and testing of the models is divided into two phases. First, we pre-train Arthur on the preprocessed UCI Census Income dataset without Merlin and Morgana. Second, we use the pre-trained model to perform further training including Merlin and Morgana. For the pre-training of Arthur, our experiments were conducted such that the models’ parameters were saved separately after each epoch. Consequently, for each experiment, we had access to a set of candidate models to choose from for further analysis. To ensure high completeness, we select the pre-trained candidate with the highest accuracy with respect to the test dataset. The selected model then represents the pre-trained classifier for all subsequent experiments involving Merlin and Morgana. 
Similar to the pre-training process, we also stored the model parameters after each epoch of training with Merlin and Morgana. The results presented in~\Cref{fig:gender} correspond to the models with the highest completeness at a soundness of at least 0.9 among all epochs. The training configurations regarding the pre-training of Arthur and the training of the  Merlin-Arthur classifiers can be obtained from~\Cref{tab:census-param}.

\begin{wrapfigure}{r}{0.4\textwidth}
    \vspace{-1cm}
    \centering
    \includegraphics[width=0.4\textwidth]{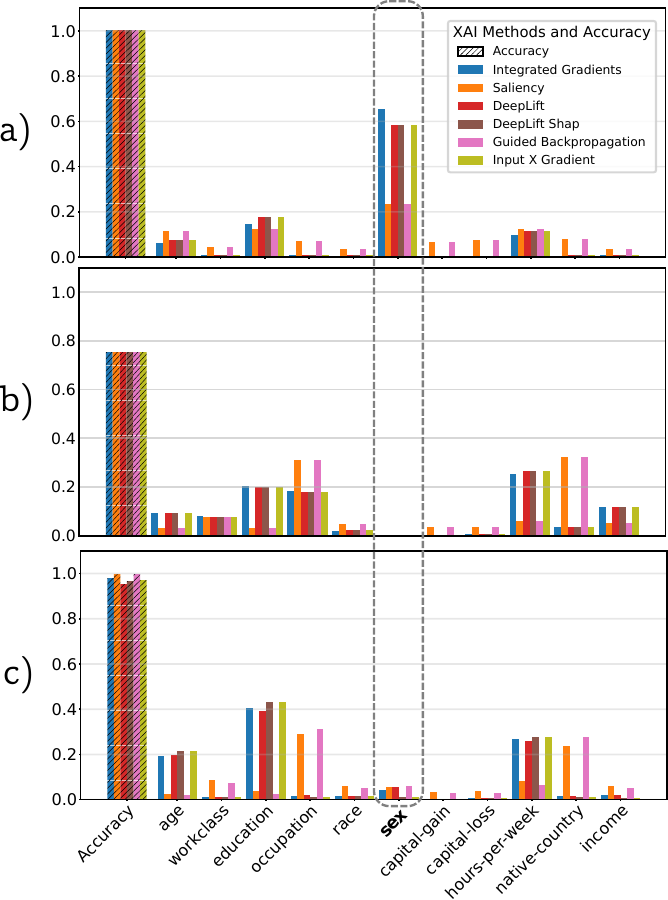}
    \caption{\small Normalised relevance scores for six XAI methods in the sex discrimination task, generated using the open-source library Captum. a) The unmanipulated network with sex feature as input achieves 100\% accuracy in the sex discrimination task, and all methods predominantly highlight the sex feature.
    b) The sex feature was omitted from the input. The network achieves an accuracy of 75\% to predict sex, and the interpretability methods predominantly highlight ``education'' and ``hours-per-week''.
    c) The sex feature is included in the input, but for every interpretability method a new network was trained with the additional objective to minimise the attribution to the sex feature. In each case, the network achieves above 98\% accuracy, but the sex feature is not strongly highlighted.}
    \label{fig:manipulation_XAI}
    \vspace{-2cm}
\end{wrapfigure}

\subsubsection{Marital-Status and Relationship Included}

In~\Cref{sec:cheating}, to show the effect most clearly, we kept the focus on a single sensitive feature (``sex'') and removed the ``relationship'' and ``marital status'' features that strongly correlate with it and are thus informative. We now show the results for the experiments including these features in~\Cref{fig:gender_corr_feat}, where Merlin uses these features to communicate ``sex'' with a completeness of 71\%. In the case 2b), where there is an incentive for soundness, but also to hide ``sex,'' Merlin manages to discriminate with 71\% accuracy, mostly by relying on the ``relationship'' feature. As the ``relationship'' feature is indeed informative about the ``sex'' of the applicant, this is in line with our theory. In such a case, further investigation is neccessary to decide whether such a feature should be protected.

\subsubsection{Other Manipulation Techniques}
\label{apx:off_manifold}

As mentioned in~\Cref{sec:cheating}, we want to further explain the argument why the off-manifold manipulations does not work on the Merlin-Arthur classifier.

LIME~\cite{ribeiro2016should} and ShaP~\cite{lundberg2017unified}  explanations are generated by sampling random inputs around the original data point and compare the classifier output on these points with the original output. The manipulation technique of \citeauthor{slack2020fooling} makes use of the fact that these random samples are off-manifold and change the off-manifold behaviour of the classifier. This allows to get the desired interpretation, while leaving the on-manifold behaviour constant (to continue to discriminate for example).

Changing the off-manifold behaviour of Merlin is without effect for our interpretation, since Merlin will always be used on-manifold.
Training Merlin to give off-manifold features to Arthur on which he changes behaviour also does not work. This behaviour is either solely off-manifold, in which case, again, it does not matter. Or it is on-manifold, in which case it is subject to potential exploits by Morgana. In case it cannot be exploited by Morgana, we already proved that the features then need to be informative.

\subsubsection{Manipulating XAI Methods through the Network Training Objective}
\label{apx:posthoc_manipulated}

In~\Cref{sec:cheating}, we demonstrate how an adversarial setup prevents a typical approach to manipulating explanations that has been demonstrated for numerous XAI methods in \citeauthor{dimanov2020you}, \citeauthor{heo2019fooling} and \citeauthor{anders2020fairwashing}. Here, we complement this analysis by showing that for the same setup, the manipulation method proposed in to obscure relevant features works for a wide range of popular post-hoc XAI methods. We used Integrated Gradients, Saliency, DeepLift, DeepShap, Guided Backpropagation, and Input X Gradient from the open-source Captum library~\citep{kokhlikyan2020captum}.

For this, we trained a regular classifier network $\Phi$ of the same architecture as Arthur on the sex discrimination task. The loss $L$ is the binary cross entropy loss for predicting ``male'' versus ``female''. We looked at three different scenarios, see~\Cref{fig:manipulation_XAI}: 
\begin{enumerate}
    \item[a)] The classifier has $L$ as its sole objective and has access to the sex feature. Unsurprisingly, it achieves perfect accuracy. All interpretability methods rate ``sex'' as the most important feature. ``Education'' and ``hours-per-week'', which correlate with ``sex'', are highlighted as well.
    \item[b)] The objective remains the same, but the sex feature is not part of the input anymore. The classifier achieves a maximal accuracy of 75\%, relying on the ``education'' and ``hours-per-week'' features.
    \item[c)] The training objective for $\Phi$ was changed to
        \[
 L^\prime(\Phi, \bfx) := L(\Phi, \bfx) + \beta_1 \nkl{\bfr_{\text{sex}}(\Phi, \bfx)} - \beta_2\nkl{\bfr_{\setminus\text{sex}}(\Phi, \bfx)},
\]
where $\bfr_{\text{sex}}$ and $\bfr_{\setminus\text{sex}}$ are the relevance scores for the ``sex'' feature and every other feature respectively. We trained a different network for every XAI method and varied $\beta_1$ and $\beta_2$ in the range of $[2,20]$, choosing the best performing.
\end{enumerate}

We observe that the inclusion of the ``sex'' feature raises the accuracy to predict ``male'' vs.\ ``female'' from 75\% to almost 98\%. At the same time, our ``manipulated'' network was able to almost perfectly hide the crucial feature for all of these common XAI approaches. This demonstrates that without an adversarial aspect as in the Merlin-Arthur framework, post-hoc interpretability is not robust against manipulation.

% \clearpage

\subsection{Training Setup for MNIST}

Here, we give a detailed description of our training setup for MNIST and show the error bars of the numerical results presented in the main part of the paper over 10 different training runs.

\emph{Structure of Arthur}. Arthur is modelled using a neural network. 
Specifically, we consider a convolutional NN with a ReLU activation function. 
For the case of two classes, we consider a NN with two convolution layers, whereas for the five class case we consider three convolution layers. 
The output of the convolution is then passed through two linear layers before being used for the output.
\Cref{tab:arthur_architecture} describes the used architecture in detail.

\begin{table}[h]
    \centering
    \vspace{2mm}
    \begin{tabular}{|c|c|}
        \hline 
        \textbf{Layer Name} & \textbf{Parameters} \\
        \hline
         Conv2D & I=3, O=32, K=3 \\ \hline
         ReLU & \\ \hline
         Conv2D & I=32, O=64, K=3 \\ \hline
         ReLU & \\ \hline
         Conv2D & I=64, O=64, K=3 \\ \hline
         ReLU & \\ \hline
         MaxPool2d & K=2\\ \hline
         Linear & I=7744, O=1024\\ \hline
         ReLU & \\ \hline
         Linear & I=1024, O=128\\ \hline
         ReLU & \\ \hline
         Linear & I=128, O=1\\ \hline
         \hline
    \end{tabular}    
    \caption{\label{tab:arthur_architecture}Description of the NN architecture used for feature classifier Arthur in the MNIST experiments.}
\end{table}
\emph{Structure of Merlin and Morgana}. Recall that Merlin and Morgana aim to ideally solve
\[
 M(\bfx) = \argmin_{\bfs \in B_k^d} L_M(A,\bfx, \bfs) \quad \text{and} \quad \morg(\bfx) = \argmax_{\bfs \in B_k^d} L_{\morg}(A,\bfx,\bfs),
\]
respectively, where $L_M$ and $L_{\morg}$ are the loss functions defined in~\Cref{sec:numerics}, and $B_k^d$ is the space of $k$-sparse binary vectors.
Thus, Merlin and Morgana take an image as input and produce a mask of the same dimension with $k$ one-entries and zero-entries otherwise.
We additionally added a regularisation term in the form of \(
\lambda\nkl{\bfs}_1\)
to both of the objectives, and set $\lambda = 0.25$.
We realise the pair Merlin and Morgana in four different ways, which we explain in the following. All of these approaches return a mask $\bfs \in [0,1]^d$ that we then binarise by setting the $k$ largest values to one and the rest to zero.

\paragraph{Frank-Wolfe Optimisers}
In the first case, Merlin and Morgana are modelled by an optimiser using the Frank-Wolfe algorithm~\citep{jaggi2013revisiting}. We follow the approach outlined by~\cite{macdonald2021interpretable} with the Frank-Wolfe package provided by~\cite{pokutta2020deep}.
The optimiser searches over a convex relaxation of $B_k^d$, i.e.,
\[
  \CB_k^d = \skl{\bfv \in [0,1]^d \,\middle|\, \nkl{\bfv}_1 \leq k },
\] 
the $k$-sparse polytope with radius $1$ limited to the positive octant. To optimise the objective we use the solver made available at \url{https://github.com/ZIB-IOL/StochasticFrankWolfe} with 200 iterations.

\paragraph{U-Net Approach}
For the second case, we model Merlin and Morgana using NNs, specifically a U-Net that has already been used in the XAI domain to create saliency maps, see \citep{dabowski2017unets}. We copy the U-Net design by~\cite{ronneberger2015u} since it achieves good results in image segmentation, a task reasonably similar to ours. We predict mask values between zero and one for each image pixel, and rescale the mask should it lie outside of $\CB_k^d$.
The binarisation of the mask is ignored during the backpropagation that trains the U-Nets and only employed to produce the masks that Arthur is trained on.

\paragraph{Hybrid Approach}
In the Hybrid approach, Merlin is modelled by a U-Net, whereas Morgana is still modelled by the FW-optimiser.
This approach is useful since for a sound Arthur that cannot be fooled, the training of the Morgana U-Net becomes difficult and the U-Net diverges. It then takes a while of training for the U-Net to adapt, should Arthur open himself to possible adversarial masks. The optimiser is applied to each individual image instead and can find possible weaknesses instantly.

\paragraph{Class-Networks}
One of the alternatives to Merlin and Morgana that we propose are class-specific U-Nets. Instead of Merlin and Morgana each being represented by a network, there is a U-Net associated with each class that is trained to produce a feature mask that convinces Arthur of its own class for any input image, i.e., for $l\in [C]$ try to solve
\[
 M_l(\bfx) = \argmin_{\bfs \in B_k^d} -\log\kl{A(\bfs\cdot\bfx)_{l}} + \lambda \nkl{\bfs}_1.
\]
Merlin is then implemented as an algorithm to choose the U-Net corresponding to the true class, so
\[
 M(\bfx) = M_{c(\bfx)}(\bfx).
\]
Morgana instead uses for each individual image the output of the U-Net that most convinces Arthur of a wrong class (maximises the Morgana-loss), i.e.,
\[
 M(\bfx) = M_{l}(\bfx)\quad\text{with}\quad l = \argmax_{l \in C\setminus \skl{c(\bfx)}} L_{\morg}(A,\bfx,M_{l}(\bfx)).
\]

Ideally, this training setup would be more stable than the normal U-Net approach.
When Arthur cannot be fooled, the class-U-Nets still have a reasonable objective in convincing him of the correct class, which hopefully prevents divergence as for the Morgana U-Net.
Experimentally, however, the class-networks proved to be much less stable than the simple U-Net approach, see~\Cref{fig:avp_error}.

We give an overview over the parameters used for the four different approaches in~\Cref{tab:training_param}.
\begin{table}[h]
    \centering
    \vspace{2mm}
    \begin{tabular}{cc}
         \hline 
         \textbf{Parameter} & \textbf{Value} \\ \hline
         Batch Size & 128 \\
         Baseline Value & 0.3 \\
         Max FW Iterations & 200 \\
         FW Momentum & 0.9\\
         Regularisation $\lambda$ & 0.25\\
         Max NN Passes & 5 \\
         Arthur LR & 1e-4 \\
         Merlin LR & 1e-4 \\
         Morgana LR & 1e-6 \\ \hline
    \end{tabular}
    \caption{\label{tab:training_param}Training parameters for the Merlin-Arthur classifier on the MNIST dataset.}
\end{table} 

\paragraph{Merlin-Arthur Classifier Training}

The overall training procedure proceeds as outlined in~\Cref{alg:training}.
We initially train Arthur directly on the training data.
In the optimiser approach, this pre-trained network is used to search for the optimal masks for Merlin and Morgana. In the U-Net approach, these masks are directly produced by the U-Nets.
Arthur is then trained on the masked images over the whole dataset.
The U-Nets are then trained on the dataset with a fixed Arthur according to their respective objectives.
We cycle through this process for five epochs. 
The learning rate is 1e-4 for the Arthur and Merlin network and 1e-6 for the Morgana and the class-specific networks. 

\subsubsection{Purely Cooperative Setup and ``Cheating'' for MNIST}

Here, we discuss the Merlin-Arthur classifier when only Merlin is used with no Morgana. Our results demonstrate that the inclusion of Morgana is necessary for Merlin and Arthur to exchange meaningful features and abstain from ``cheating''.
For a purely cooperative setup, information about the class $c(\bfx)$ that Arthur infers is dominated by the fact that Merlin chose that feature, rather than the feature itself, i.e., $H(c(\bfx) | M(\bfx) = \bfz) \ll H(c(\bfy) | \bfz \subseteq \bfy)$.
We can upper bound $H(c(\bfx) | M(\bfx) = \bfz)$ through the classification error $P_e$ that Arthur and Merlin achieve via Fano's inequality~\citep{fano1961transmission}:
\[
 H(c(\bfx) | M(\bfx) = \bfz)  \leq H_b(P_e) + P_e\log(\bkl{C} - 1),\]
where $\bkl{C}$ is the number of classes. We can then bound the amount of information that is transferred by the choice of the mask rather than the feature itself as
\begin{align*}
I(c(\bfx);M(\bfx)=\bfz) - I(c(\bfy);\bfz \subseteq \bfy)
 &= H(c(\bfy)~|~\bfz \subseteq \bfy) - H(c(\bfx)~|~M(\bfx) = \bfz) \\
 &\geq   H(c(\bfy)~|~\bfz \subseteq \bfy) - (H_b(P_e) + P_e\log(\bkl{C} - 1)).
\end{align*}
We define the \emph{cooperative information} as
\begin{equation*}
  I_{\text{coop}} := \max\left(0, H(c(\bfy)~|~\bfz \subseteq \bfy) - (H_b(P_e) + P_e\log(\bkl{C} - 1))\right),
\end{equation*}
which lower bounds how much Arthur and Merlin ``cheat''. We train Merlin and Arthur on the MNIST dataset and show the results in~\Cref{fig:fano_error} that in the purely cooperative case $I_{\text{coop}}$ keeps the classification error low despite exchanging uninformative features. Including Morgana on the other hand pushes $I_{\text{coop}}$ to zero even for small masks. In this case Merlin produces highly informative features.

\Cref{fig:fano_error} depicts ``Cheating'' measured by $I_{coop}$ over the MNIST test dataset (restricted to 5 classes) for different setups. In a purely cooperative setup, Arthur and Merlin learn to communicate over uninformative features. Thus, the classification error $P_e$ stays low, while the conditional entropy of the class with respect to the feature goes up. When Morgana is present and soundness is high (in our analysis $\epsilon_s < 0.05$),  Arthur and Merlin cannot cheat, the classification error increases for small $k$.

Why is this observation important? Note that the purely cooperative setup could already be seen as interpretable. With Merlin as a network it appears as a version of a self-interpreting network. With Merlin as an optimiser, it is similar to Rate-Distortion Explanations (RDE)~\citep{macdonald2019rate}. In fact, RDE have been criticised in~\citep{waldchen2022training} for producing ``cheating'' masks that artificially create features that were not present in the original image. We connect this to the lack of robustness with respect to Morgana. Designers of self-interpreting setups need to explain how they circumvent the problem of ``cheating'' masks.

\begin{figure}[h!]
    \centering
    \resizebox{0.6\textwidth}{!}{

\input{img/fano_error/graphic}
    }
    \caption{\label{fig:fano_error}\small{This figure depicts the mean and standard deviation over 10 training runs for the error probability, cooperative information and the class entropy. This was obtained from 5-class classification with classes 1,2,3,4 and 5 with $\gamma = 0.75$ for the purely cooperative setup (\emph{left}) and the adversarial setup (\emph{right}), where Merlin was realised as an optimiser (\emph{top}) or as a neural network (\emph{bottom}).}}   
\end{figure}
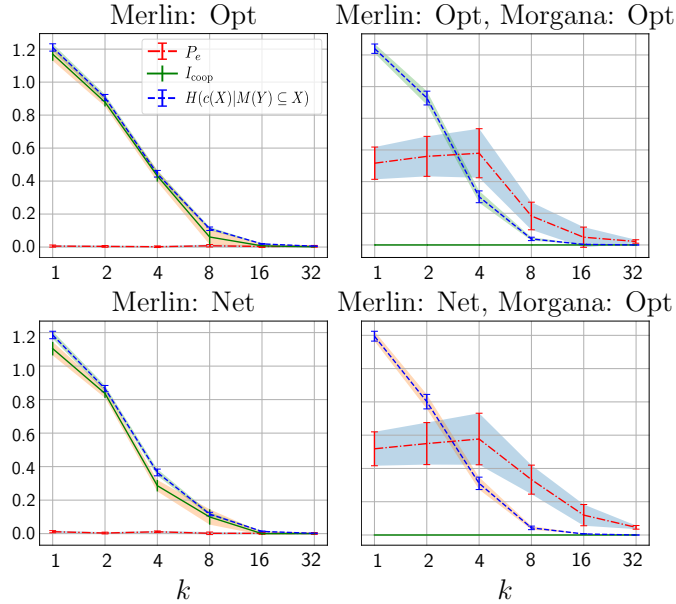
\begin{figure}[h!]
    \centering
    \resizebox{0.6\textwidth}{!}{

\input{img/bound_error/graphic}}
    \caption{\label{fig:avp_error}\small{We show the mean and standard deviation over 10 training runs for completeness and soundness, along with the average precision and its bound as obtained from 2-class classification with classes 6 and 9 with $\gamma = 0.75$ for different settings.}}
    \label{fig:training_variance}
\end{figure}
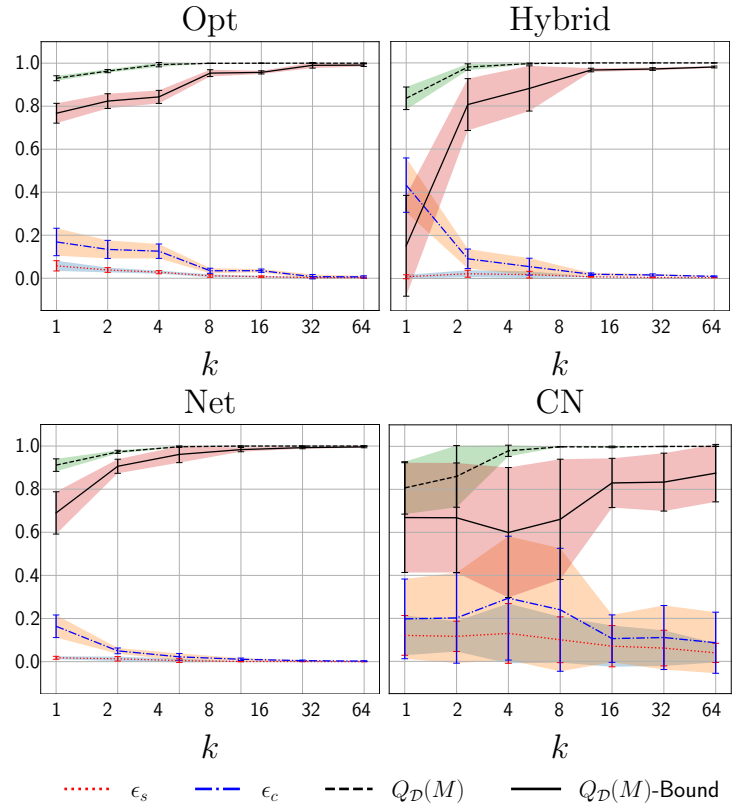

\Cref{fig:avp_error} depicts the results from the main paper in more detail.
Specifically, it depicts averages and the standard deviation as error bars for each parameter using samples from 10 different training runs. 
The results presented here are consistent with those presented in the main body of the paper, except for the class networks. 
The error bars are large for small mask sizes, but shrink as the mask size increases.
The class-network approach is considerably less stable than our other implementations, even though it was conceived as a stabler alternative to the simple network approach. One possibility might be that since each U-Net both cooperates with Arthur and wants to fool him, they are more sensitive when Arthur switches between achieving good soundness and completeness. We hope that further research will determine if this realisation can be trained in a stable manner.

\subsubsection{Stability of the Three-player Game}\label{apx:stability}

Min-max games are indeed challenging to optimise, nevertheless they are common in state-of-the art approaches, e.g., in adversarial robustness or GAN-training~\citep{roth2017stabilizing, wiatrak2019stabilizing}. More sophisticated training routines that have already been developed in these areas are a way to stabilise Merlin-Arthur training for more complex datasets.

\Cref{fig:training_variance} shows the error bars for soundness and completeness for 10 independent runs with 12 rounds of training, respectively. One can see that the variance for the Opt, Hybrid and Network case is quite small.
We find that the more information Merlin is allowed to send to Arthur, the more stable the training. 
For large features, we come close to the regime studied by \citeauthor{anil2021learning}. They show that when a strategy is assumed to have perfect completeness and soundness, there is no reason for the agents to change their strategies and the system is in equilibrium. However, in Figure 6, we also investigate the more interesting regime where the mask size is so small that no perfect strategies exist, leading to a trade-off between soundness and completeness.
In our experiments, the agents vary along this trade-off during training. We push this trade-off strongly in the direction of soundness ($\gamma > 0.5$ in Arthur's objective), and obtain greater stability.

Another factor is that stability is not a prime concern for our setup. The advantage of our theoretical bound is that, compared to \citeauthor{anil2021learning}, we do not assume that the system is in equilibrium. Even if the agents do not converge to the equilibrium, one can take a well-performing snapshot during training, and our bounds still apply. However, in our experiments this was not necessary, and we always evaluate soundness and completeness after 12 rounds of training.

%% file: img/debate_new.tex
\begin{tabular}{c@{\hspace{1cm}}c}
\begin{tikzpicture}
% \begin{scope}[every node/.style={circle,thick,draw,fill=white}]
    \node (c1) at (0,0) {l = -1};
    \node (c2) [right=\hor of c1] {l = 1};
    \node[circle,draw=none,fill=white] (x1) [below=\ver of c1] {$\bm{[1,2]}$};
    \node[circle,draw=none,fill=white] (x2) [below=\ver of c2] {$\bm{[3,2]}$};
    \node[circle,draw=none,fill=white] (x3) [below=\ver of x1]  {$\bm{[3,4]}$};
    \node[circle,draw=none,fill=white] (x4) [below=\ver of x2] {$\bm{[5,4]}$};
    \node[circle,draw=none,fill=white] (x5) [below=\ver of x3]  {$\bm{[5,6]}$};
    \node[circle,draw=none,fill=white] (x6) [below=\ver of x4] {$\bm{[7,6]}$};
    \node[circle,draw=none,fill=white] (x7) [below=\ver of x5]  {$\bm{[7,8]}$};
    \node[circle,draw=none,fill=white] (x8) [below=\ver of x6]  {$\bm{[9,8]}$};
    \node[] (x9) [below=1.4cm of x7]  {};
    
    \draw[-] (x1) -- (x2) node[midway,circle,draw=none,fill=white] (phi1) {$[*,2]$};
    \draw[-] (x2) -- (x3) node[midway,circle,draw=none,fill=white] (phi2) {$[3,*]$};
    \draw[-] (x3) -- (x4) node[midway,circle,draw=none,fill=white] (phi3) {$[*,4]$};
    \draw[-] (x4) -- (x5) node[midway,circle,draw=none,fill=white] (phi4) {$[5,*]$};
    \draw[-] (x5) -- (x6) node[midway,circle,draw=none,fill=white] (phi5) {$[*,6]$};
    \draw[-] (x6) -- (x7) node[midway,circle,draw=none,fill=white] (phi6) {$[7,*]$};
    \draw[-] (x7) -- (x8) node[midway,circle,draw=none,fill=white] (phi7) {$[*,8]$};
    \node[] (phi0) [above=1.4cm of phi2]  {};
    
    \node[circle,draw=none,fill=white] (phi8) at ($(x8)!0.5!(x9)$) {$[9,*]$};
    %\node[circle,thick,draw,fill=white] (phi8) [below=\ver of phi6]  {$\phi_8$};
    \draw[-] (phi8) -- (x8);
    \draw[-, dashed, thick] (phi8) -- (x9);
    \draw[-, dashed, thick] (phi0) -- (x1);
\end{tikzpicture}
&
\begin{tikzpicture}
% \begin{scope}[every node/.style={circle,thick,draw,fill=white}]
    \node (c1) at (0,0) {l = -1};
    \node (c2) [right=\hor of c1] {l = 1};
    \node[circle,draw=none,fill=white] (x1) [below=\ver of c1] {$\bm{[1,2]}$};
    \node[circle,draw=none,fill=white] (x2) [below=\ver of c2] {$\bm{[3,2]}$};
    \node[circle,draw=none,red,fill=white] (x3) [below=\ver of x1]  {$\bm{[3,4]}$};
    \node[circle,draw=none,fill=white] (x4) [below=\ver of x2] {$\bm{[5,4]}$};
    \node[circle,draw=none,fill=white] (x5) [below=\ver of x3]  {$\bm{[5,6]}$};
    \node[circle,draw=none,fill=white] (x6) [below=\ver of x4] {$\bm{[7,6]}$};
    \node[circle,draw=none,fill=white] (x7) [below=\ver of x5]  {$\bm{[7,8]}$};
    \node[circle,draw=none,fill=white] (x8) [below=\ver of x6]  {$\bm{[9,8]}$};
    \node[] (x9) [below=1.4cm of x7]  {};
    
    \draw[-] (x1) -- (x2) node[midway,circle,draw=none,fill=white] (phi1) {$[*,2]$};
    \draw[-] (x2) -- (x3) node[midway,circle,red,draw=none,fill=white] (phi2) {$[3,*]$};
    \draw[-] (x3) -- (x4) node[midway,circle,draw=none,fill=white] (phi3) {$[*,4]$};
    \draw[-] (x4) -- (x5) node[midway,circle,draw=none,fill=white] (phi4) {$[5,*]$};
    \draw[-] (x5) -- (x6) node[midway,circle,draw=none,fill=white] (phi5) {$[*,6]$};
    \draw[-] (x6) -- (x7) node[midway,circle,draw=none,fill=white] (phi6) {$[7,*]$};
    \draw[-] (x7) -- (x8) node[midway,circle,draw=none,fill=white] (phi7) {$[*,8]$};
    \node[] (phi0) [above=1.4cm of phi2]  {};
    
    \node[circle,draw=none,fill=white] (phi8) at ($(x8)!0.5!(x9)$) {$[9,*]$};
    %\node[circle,thick,draw,fill=white] (phi8) [below=\ver of phi6]  {$\phi_8$};
    \draw[-] (phi8) -- (x8);
    \draw[-, dashed, thick] (phi8) -- (x9);
    \draw[-, dashed, thick] (phi0) -- (x1);
\end{tikzpicture}
\\
a) & b)
\end{tabular}

%% file: img/census_data/uci_corr_features.tex
\Large{
  \centering
  \begin{tabular}{@{}r@{\,\,}r@{}}
    1a) 
  \begin{tabular}{l}
       $\beta=0$  \\
       $\gamma=0$
  \end{tabular}
  \raisebox{-.5\height}{
  \includegraphics[width=13cm]{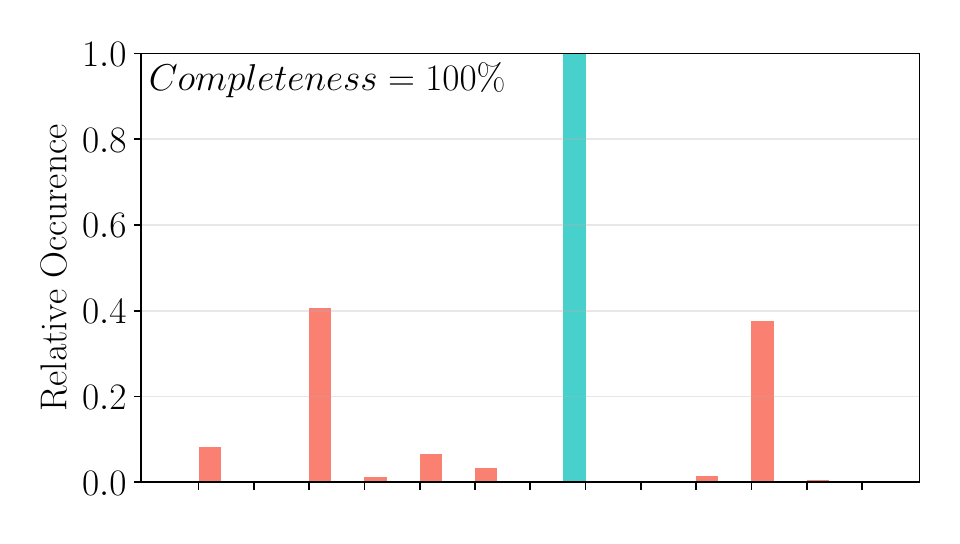}
  }
  &
    1b) 
  \begin{tabular}{l}
       $\beta=3.5$  \\
       $\gamma=0$
  \end{tabular}
  \raisebox{-.5\height}{
  \includegraphics[width=13cm]{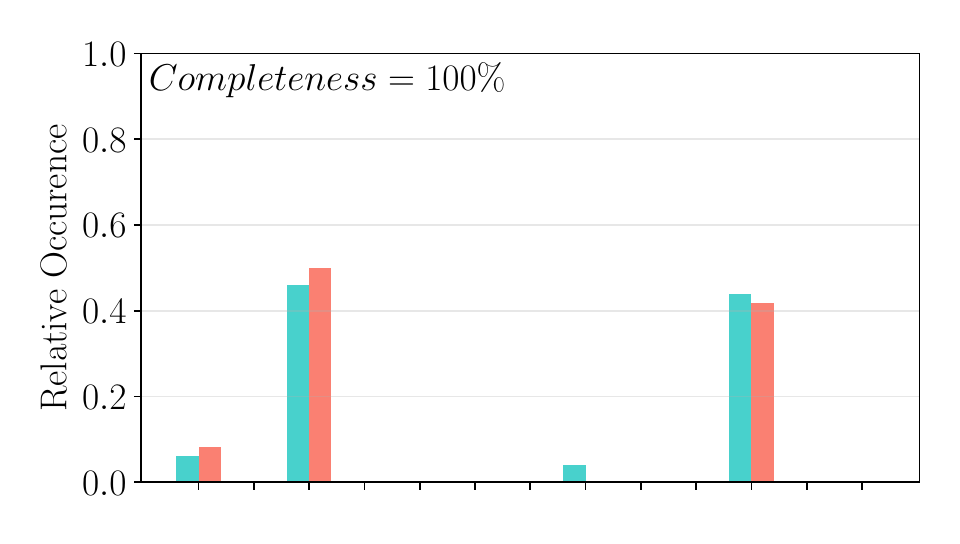}
  }\\[-1.5em]
    % c) $\beta=1, \gamma=1$ & d) $\beta=10, \gamma=1$
%   \\
    2a) 
  \begin{tabular}{l}
       $\beta=3.5$  \\
       $\gamma=0.67$
  \end{tabular}
    \raisebox{-.6\height}{
  \includegraphics[width=13cm]{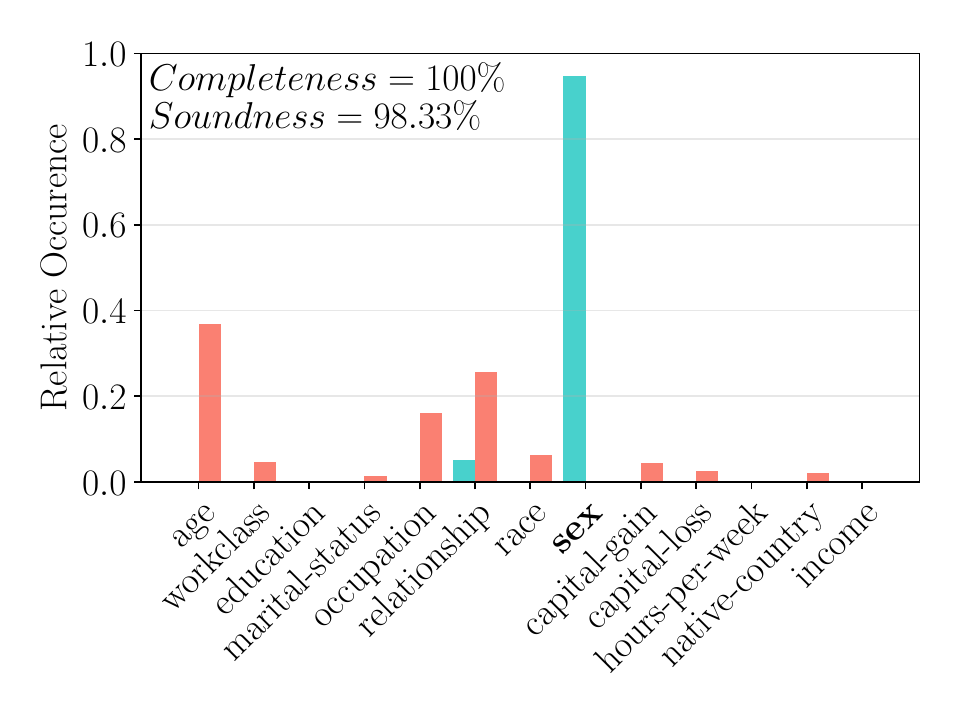}
  }
  &
    2b) 
  \begin{tabular}{l}
       $\beta=10$  \\
       $\gamma=0.67$
  \end{tabular}
  \raisebox{-.6\height}{
  \includegraphics[width=13cm]{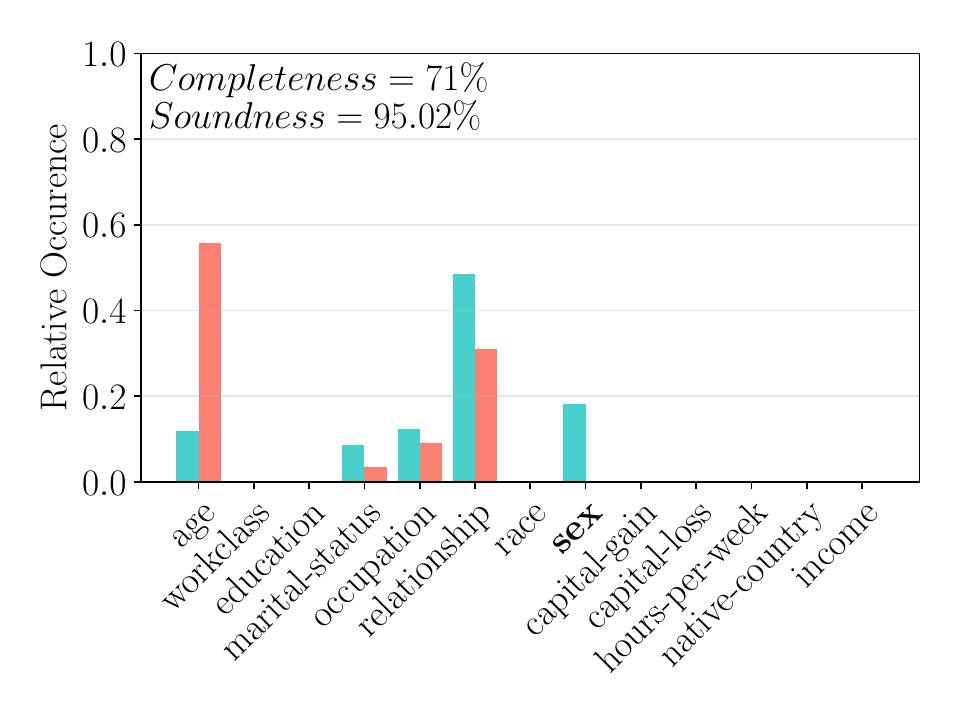}
  }\\
  
       \multicolumn{2}{c}{
       \includegraphics[width=13cm]{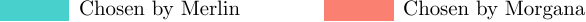}}
    \end{tabular}
}

\begin{tikzpicture}[overlay]

\draw [draw=black,dashed] (-22.7,-4.7) rectangle (-22,8.1);
\draw [draw=black,dashed] (-5.9,-4.7) rectangle (-5.2,8.1);

\end{tikzpicture}

%% file: img/fano_error/graphic.tex
\Large{
\centering
\begin{tabular}{@{}c@{\hspace{-0.2em}}c@{\;}c@{\hspace{1.2cm}}m{0.6cm}}
%%%%%%%%%%%%%%%%%%%%%%%%%%%%%%%%%%%%%%%%%%%%
& Merlin: Opt & Merlin: Opt, Morgana: Opt  \\
%%%%%%%%%%%%%%%%%%%%%%%%%%%%%%%%%%%%%%%%%%%%
\raisebox{-.5\height}{
\includegraphics[height=4cm]{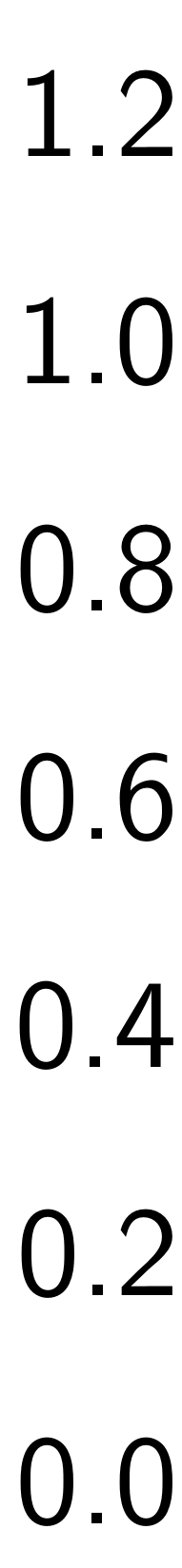}
}
&
\raisebox{-.5\height}{\includegraphics[height=4cm]{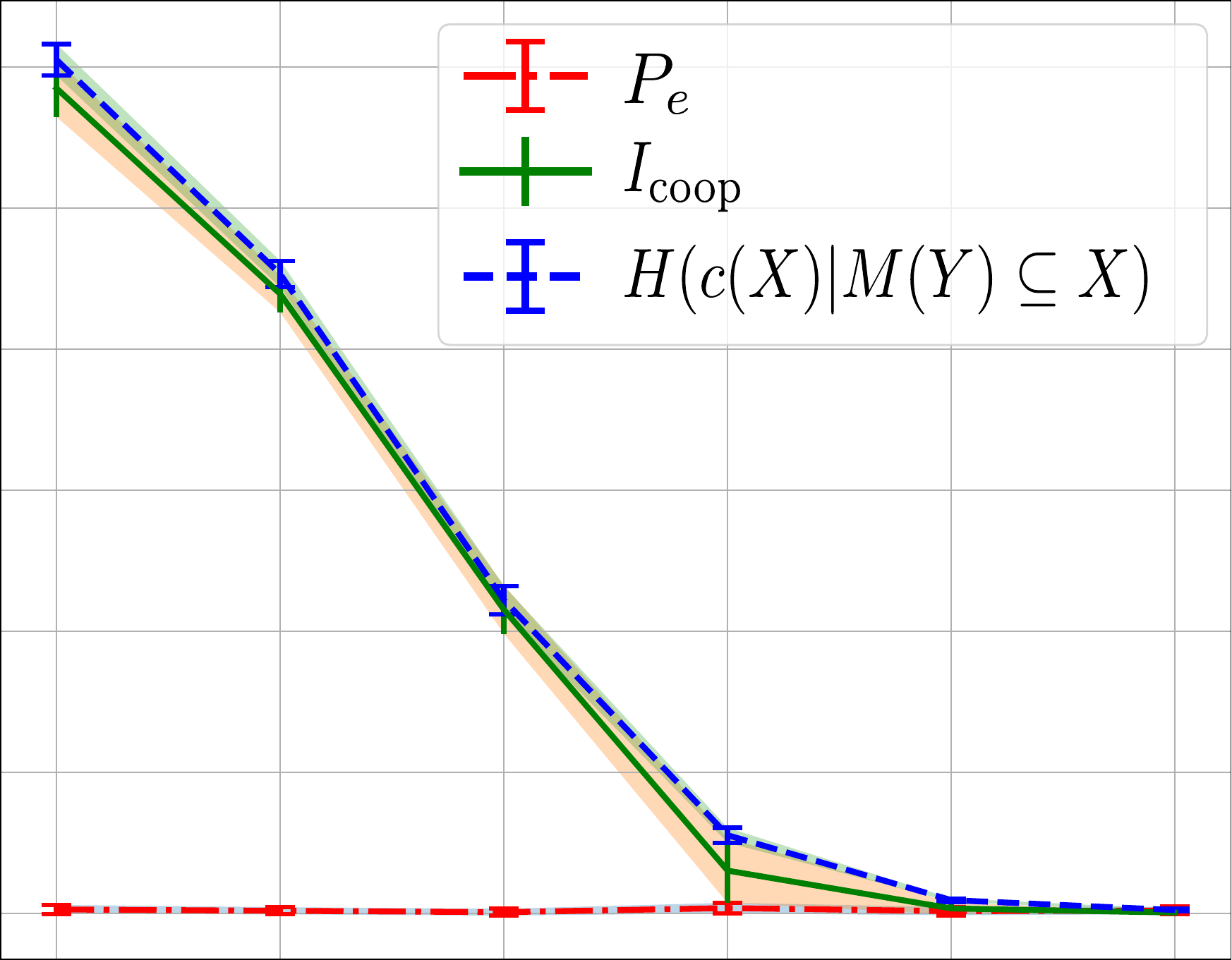}}
&
\raisebox{-.5\height}{\includegraphics[height=4cm]{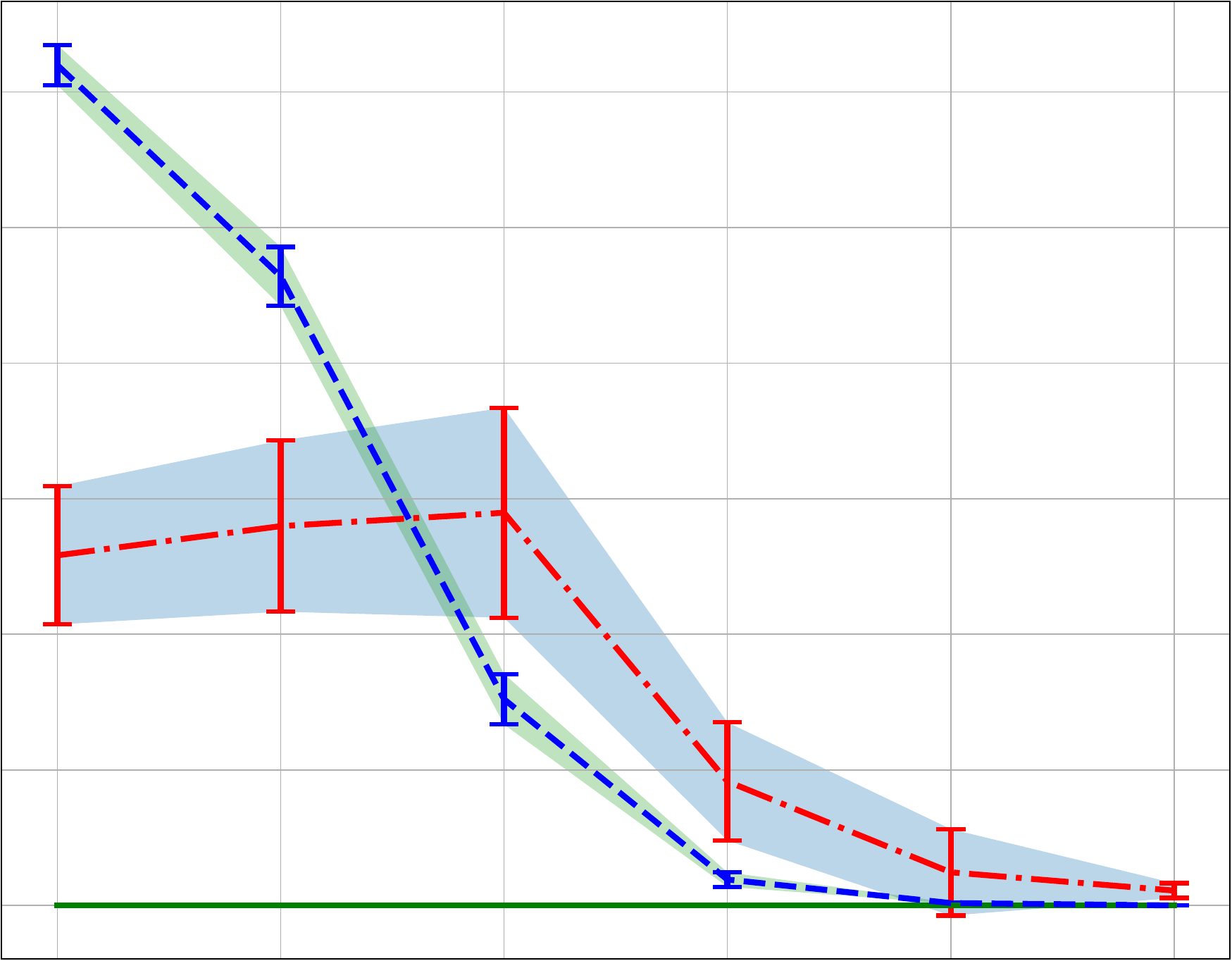}}
\\
%%%%%%%%%%%%%%%%%%%%%%%%%%%%%%%%%%%%%%%%%%%%
&
\includegraphics[width=5cm]{img/fano_violation/fano_x_scale.pdf}
&
\includegraphics[width=5cm]{img/fano_violation/fano_x_scale.pdf}
\\
%%%%%%%%%%%%%%%%%%%%%%%%%%%%%%%%%%%%%%%%%%%%
&  Merlin: Net & Merlin: Net, Morgana: Opt \\
%%%%%%%%%%%%%%%%%%%%%%%%%%%%%%%%%%%%%%%%%%%%
\raisebox{-.5\height}{
\includegraphics[height=4cm]{img/fano_violation/fano_y_scale.pdf}
}
&
\raisebox{-.5\height}{\includegraphics[height=4cm]{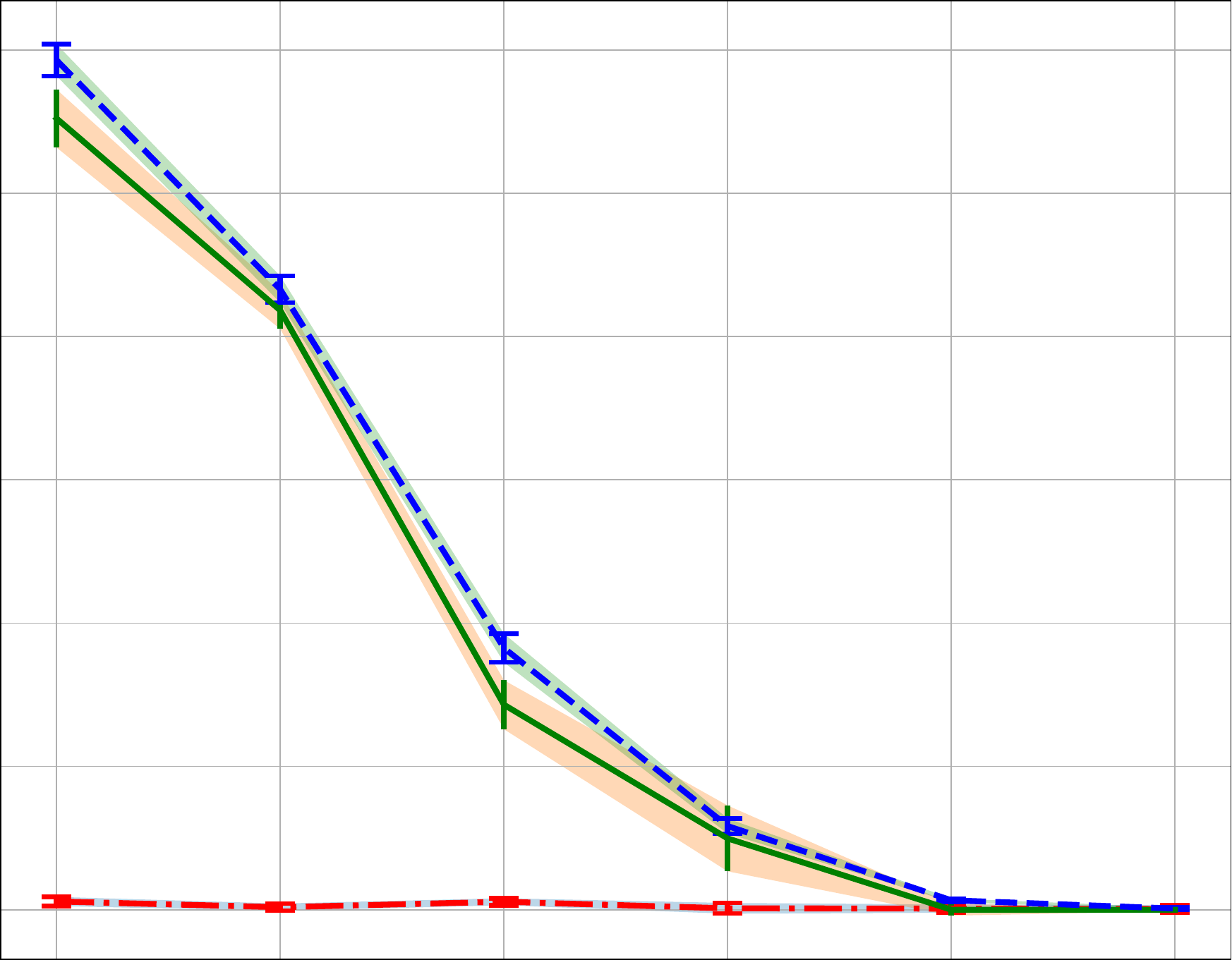}}
&
\raisebox{-.5\height}{\includegraphics[height=4cm]{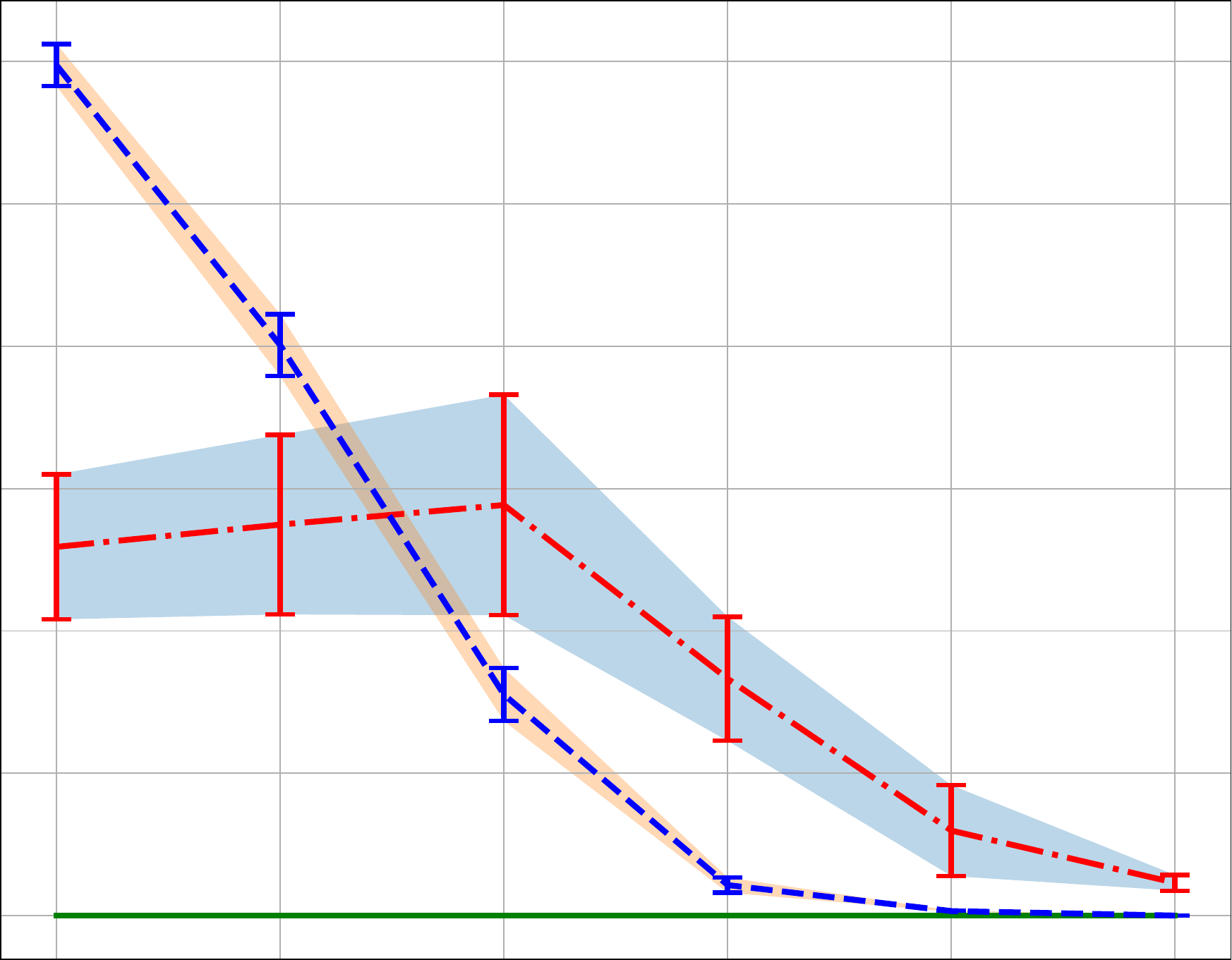}}
\\
%%%%%%%%%%%%%%%%%%%%%%%%%%%%%%%%%%%%%%%%%%%%
&
\includegraphics[width=5cm]{img/fano_violation/fano_x_scale.pdf}
&
\includegraphics[width=5cm]{img/fano_violation/fano_x_scale.pdf}
\\
%%%%%%%%%%%%%%%%%%%%%%%%%%%%%%%%%%%%%%%%%%%%
&
$k$
&
$k$

%%%%%%%%%%%%%%%%%%%%%%%%%%%%%%%%%%%%%%%%%%%%
\end{tabular}
}

%% file: img/bound_error/graphic.tex
\Large{
  \centering
  \begin{tabular}{c@{}c@{\;}c@{\;}m{0.6cm}}
  &
  Opt
  &
  Hybrid
  \\
          \raisebox{-.5\height}{\includegraphics[height=4cm]{img/bound_comparison/bound_y_scale.pdf}}
                  &
        \raisebox{-.5\height}{\includegraphics[height=4cm]{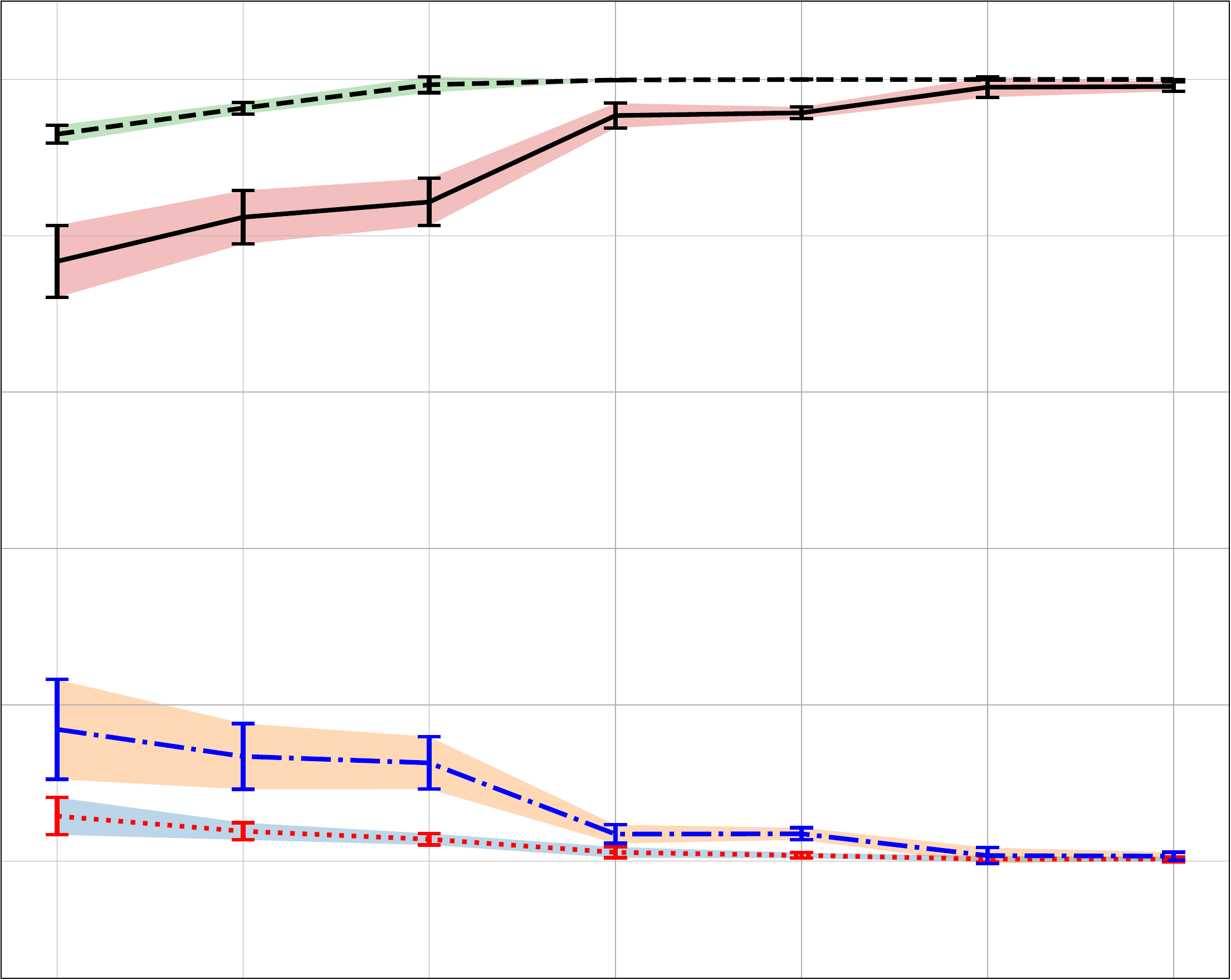}}
  &
  \raisebox{-.5\height}{\includegraphics[height=4cm]{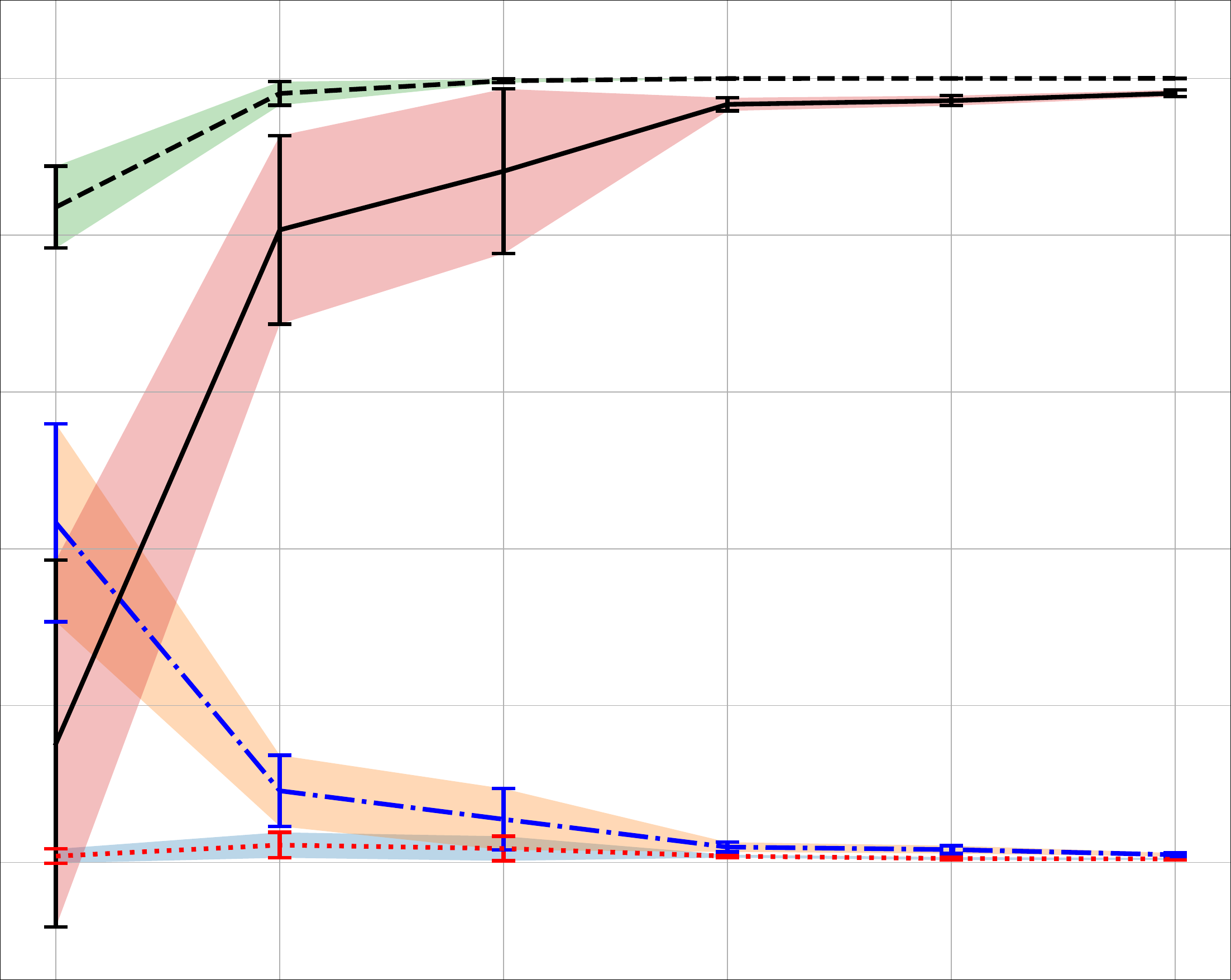}}
        \\
                 &
       \includegraphics[width=5cm]{img/bound_comparison/bound_x_scale.pdf}
                &
       \includegraphics[width=5cm]{img/bound_comparison/bound_x_scale.pdf}
       \\
       & $k$ & $k$ \\
&
Net
  &
  CN
  \\
            \raisebox{-.5\height}{\includegraphics[height=4cm]{img/bound_comparison/bound_y_scale.pdf}}
            &
\raisebox{-.5\height}{\includegraphics[height=4cm]{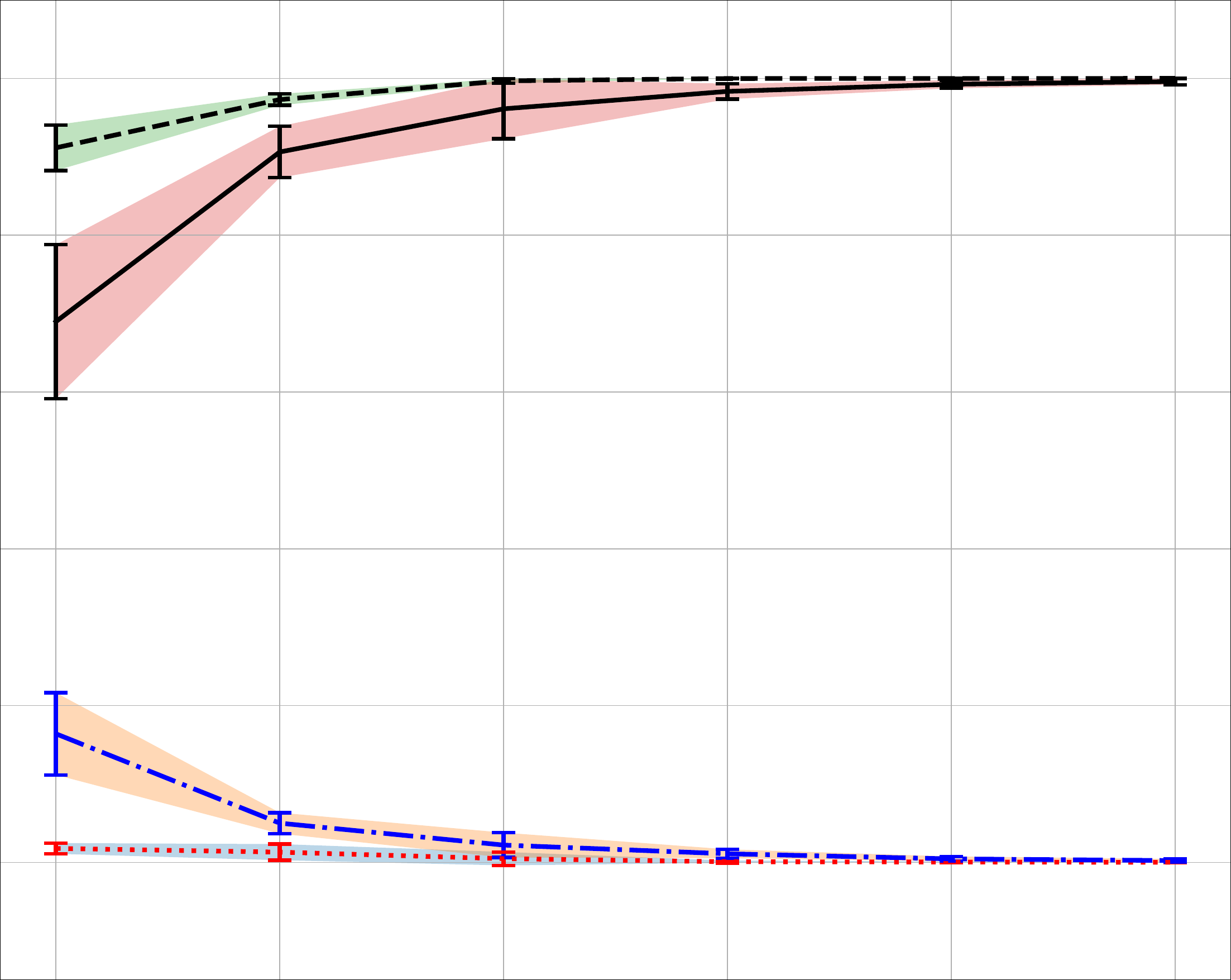}}
        &
\raisebox{-.5\height}{\includegraphics[height=4cm]{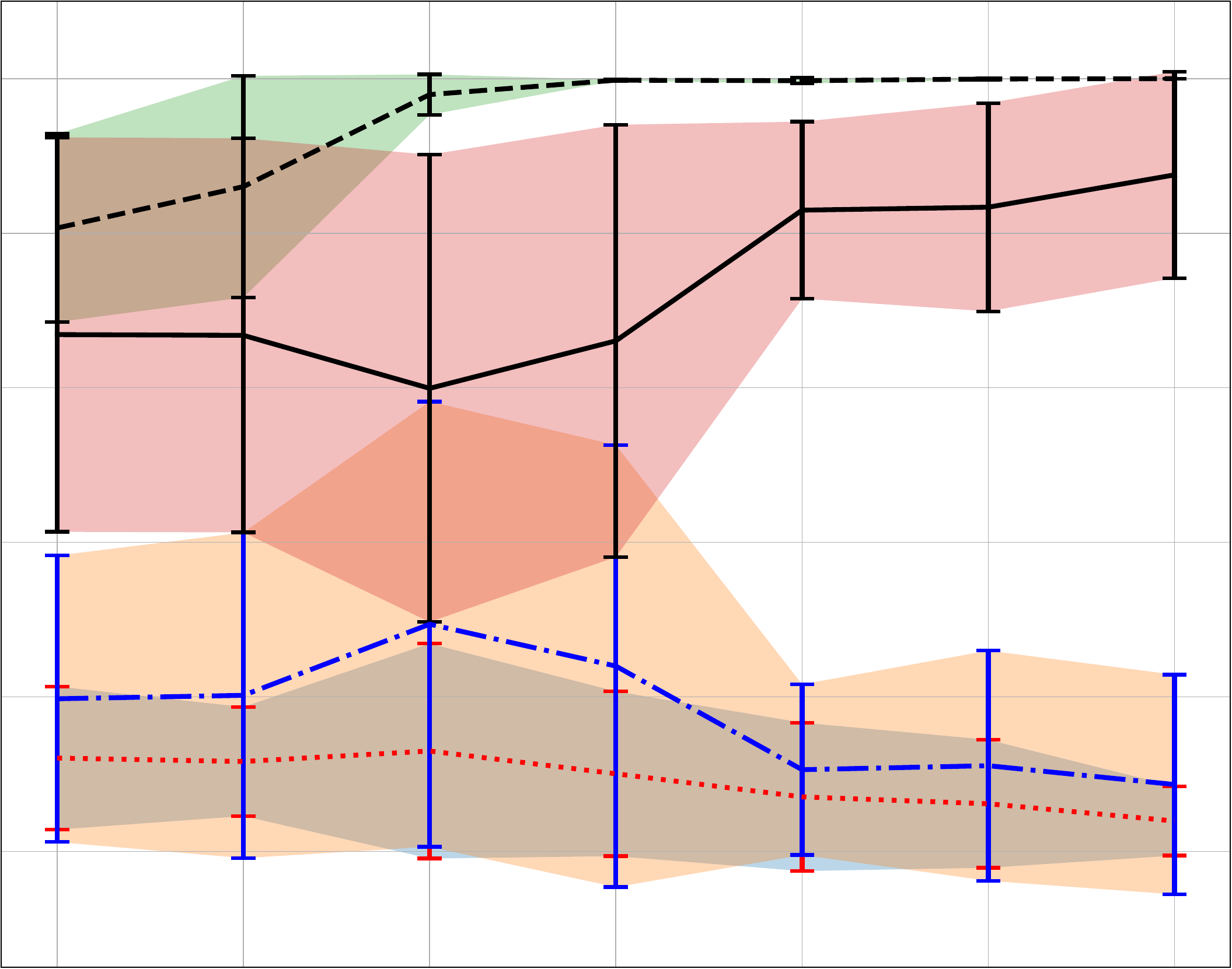}}
         \\
                &
       \includegraphics[width=5cm]{img/bound_comparison/bound_x_scale.pdf}
                &
       \includegraphics[width=5cm]{img/bound_comparison/bound_x_scale.pdf}
             \\
         &
       $k$
       &
       $k$
       \\
       &
       \multicolumn{3}{c}{\includegraphics[width=10cm]{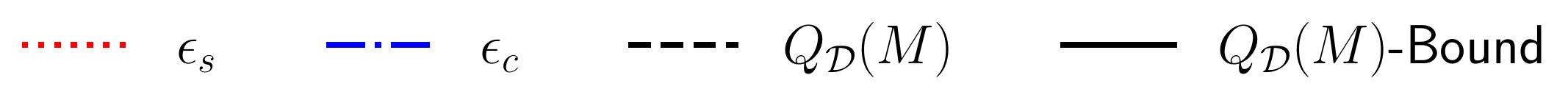}}
    \end{tabular}
}